\documentclass[final,3p,times,twocolumn]{elsarticle}
\usepackage{amssymb}
\usepackage{amsmath,amssymb,amsfonts}
\usepackage{algorithmic}
\usepackage{algorithm}
\usepackage{array}
\usepackage[caption=false,font=small,labelfont=rm,textfont=rm]{subfig}
\usepackage{amsthm}

\usepackage{epsfig} % for postscript graphics files
\usepackage{hyperref}
\usepackage{multirow}
\usepackage{booktabs}
\usepackage{threeparttable}
\usepackage[table,xcdraw]{xcolor}
\usepackage{adjustbox}
\usepackage{wrapfig}
\usepackage{bm}

\usepackage{multicol}
\usepackage{tikz}
\journal{Robotics and Autonomous Systems}

\begin{document}
	
	\begin{frontmatter}
		
		
		\title{A Rapid Iterative Trajectory Planning Method for Automated Parking through Differential Flatness\tnoteref{label1}}
		% \tnotetext[label1]{123}
		\author[1,2]{Zhouheng Li}
		\author[1]{Lei Xie\corref{cor1}}
		\ead{leix@iipc.zju.edu.cn}
		\author[1]{Cheng Hu}
		\author[1]{Hongye Su}
		
		\tnotetext[label1]{This work was supported in part by the National Natural Science Foundation of P.R. China (NSFC: 62073286), in part by the Supported by Jianbing Lingyan Foundation of Zhejiang Province, P.R. China (Grant No. 2023C01022).}
		
		\cortext[cor1]{Corresponding author.}

		\affiliation[1]{organization={State Key Laboratory of Industrial Control Technology},
			addressline={Zhejiang University},
			city={Hangzhou},
			postcode={310027},
			country={China}}
		
		\affiliation[2]{organization={Ningbo Innovation Center},
			addressline={Zhejiang University},
			city={Ningbo},
			postcode={315100},
			country={China}}

		\begin{abstract}
			As autonomous driving continues to advance, automated parking is becoming increasingly essential. However, significant challenges arise when implementing path velocity decomposition (PVD) trajectory planning for automated parking. The primary challenge is ensuring rapid and precise collision-free trajectory planning, which is often in conflict. The secondary challenge involves maintaining sufficient control feasibility of the planned trajectory, particularly at gear shifting points (GSP). This paper proposes a PVD-based rapid iterative trajectory planning (RITP) method to solve the above challenges. The proposed method effectively balances the necessity for time efficiency and precise collision avoidance through a novel  collision avoidance framework. Moreover, it enhances the overall control feasibility of the planned trajectory by incorporating the vehicle kinematics model and including terminal smoothing constraints (TSC) at GSP during path planning. Specifically, the proposed method leverages differential flatness to ensure the planned path adheres to the vehicle kinematic model. Additionally, it utilizes TSC to maintain curvature continuity at GSP, thereby enhancing the control feasibility of the overall trajectory. The simulation results demonstrate superior time efficiency and tracking errors compared to model-integrated and other iteration-based trajectory planning methods. In the real-world experiment, the proposed method was implemented and validated on a ROS-based vehicle, demonstrating the applicability of the RITP method for real vehicles.
		\end{abstract}
		
		\begin{graphicalabstract}
			\newcommand{\myrect}[1][red]{%
				\tikz[baseline=0ex]\filldraw[#1, line width=0.5mm] (0,0) rectangle (1ex,1ex);
			}
			\begin{figure}[htbp]
				\centering
				\includegraphics[scale=0.62]{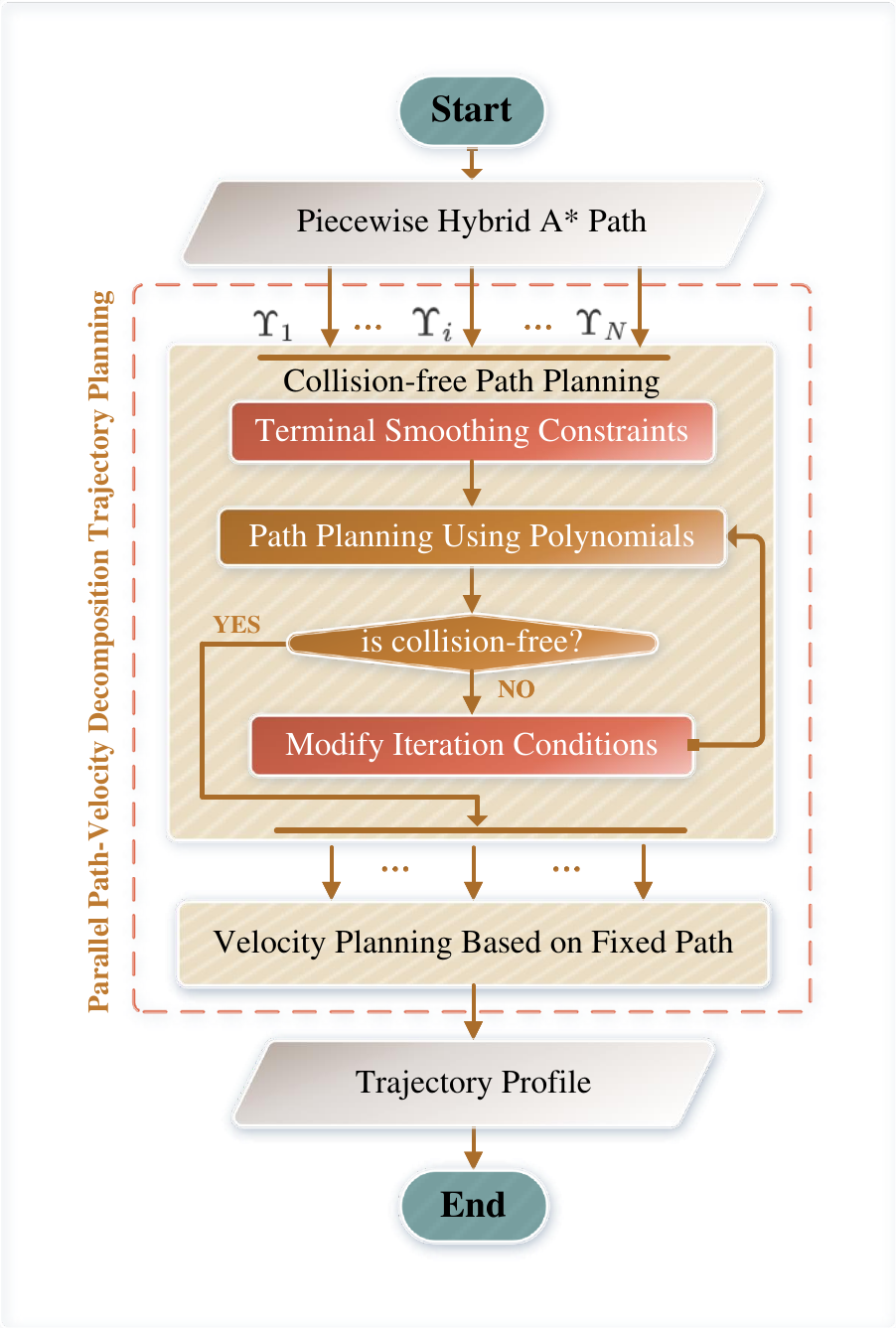}
				\caption{Flowchart of the Rapid Iterative Trajectory Planning (RITP) Method. The proposed method effectively balances the necessity for time efficiency and precise collision avoidance with a novel collision avoidance framework. Moreover, it enhances the overall control feasibility of the planned trajectory by incorporating the vehicle kinematics model and including terminal smoothing constraints (TSC) at gear shifting points (GSP) during path planning.}
			\end{figure}
			\vspace{1cm}
			\newcommand{\newimgs}{0.18}
			\newcommand{\imgs}{0.15}
			\begin{figure}[htbp]
				\centering
				\subfloat[The cyan path indicates the planning result at the first iteration. The different colored curves represent the planned paths generated during iterations.]{\includegraphics[scale=0.18]{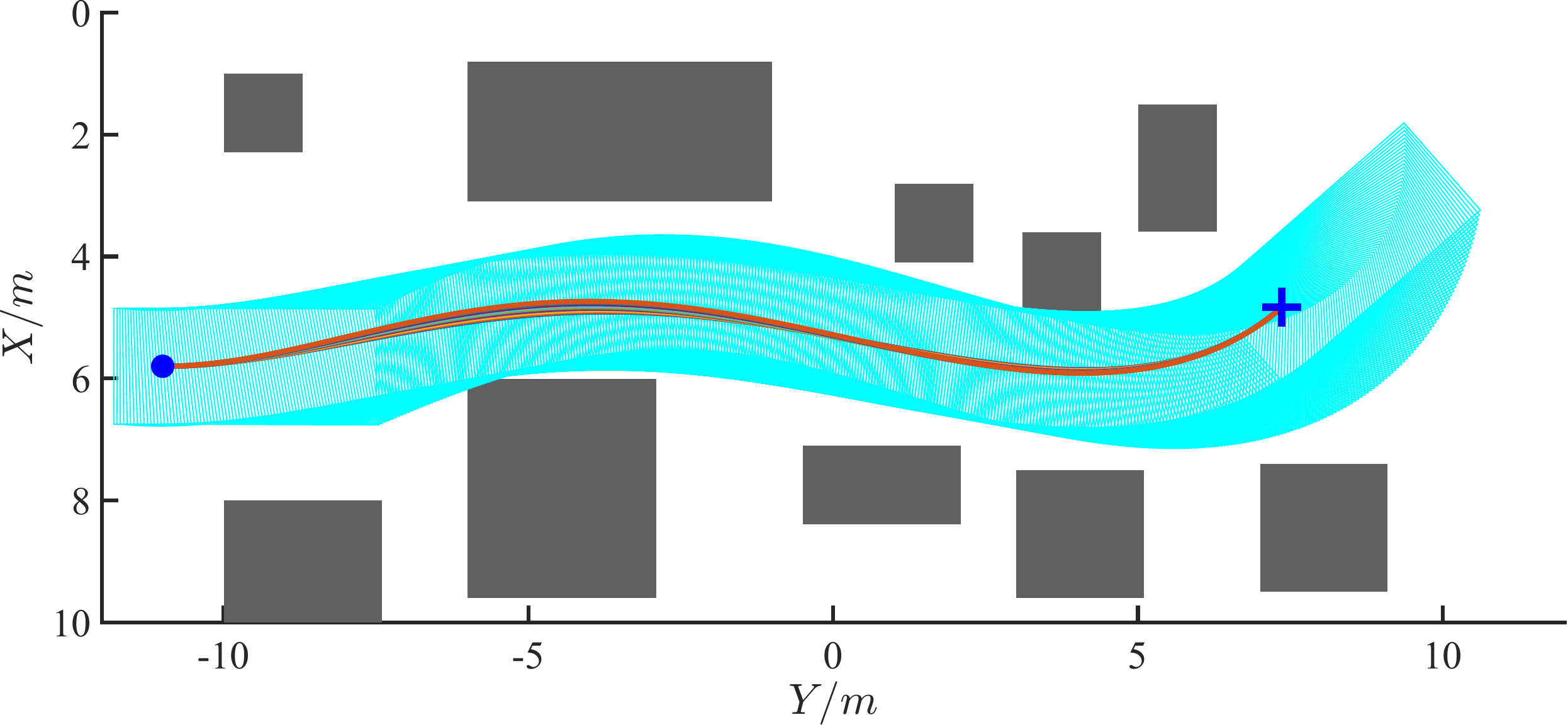}}
				\hfil
				\subfloat[Details of the collision between the planned path and obstacles in the first iteration.]{\includegraphics[scale=0.24]{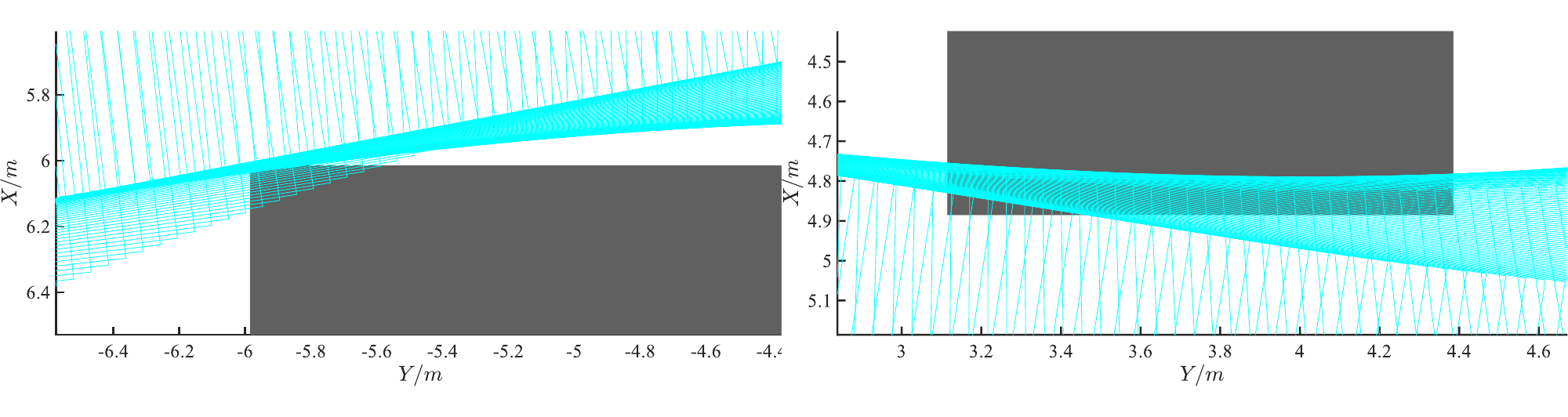}}
				\hfil
				\subfloat[Schematic diagram of the variation of the error vector $\mathbf{e}$ during Iterative Collision Avoidance (ItCA) process.]{\includegraphics[scale=0.18]{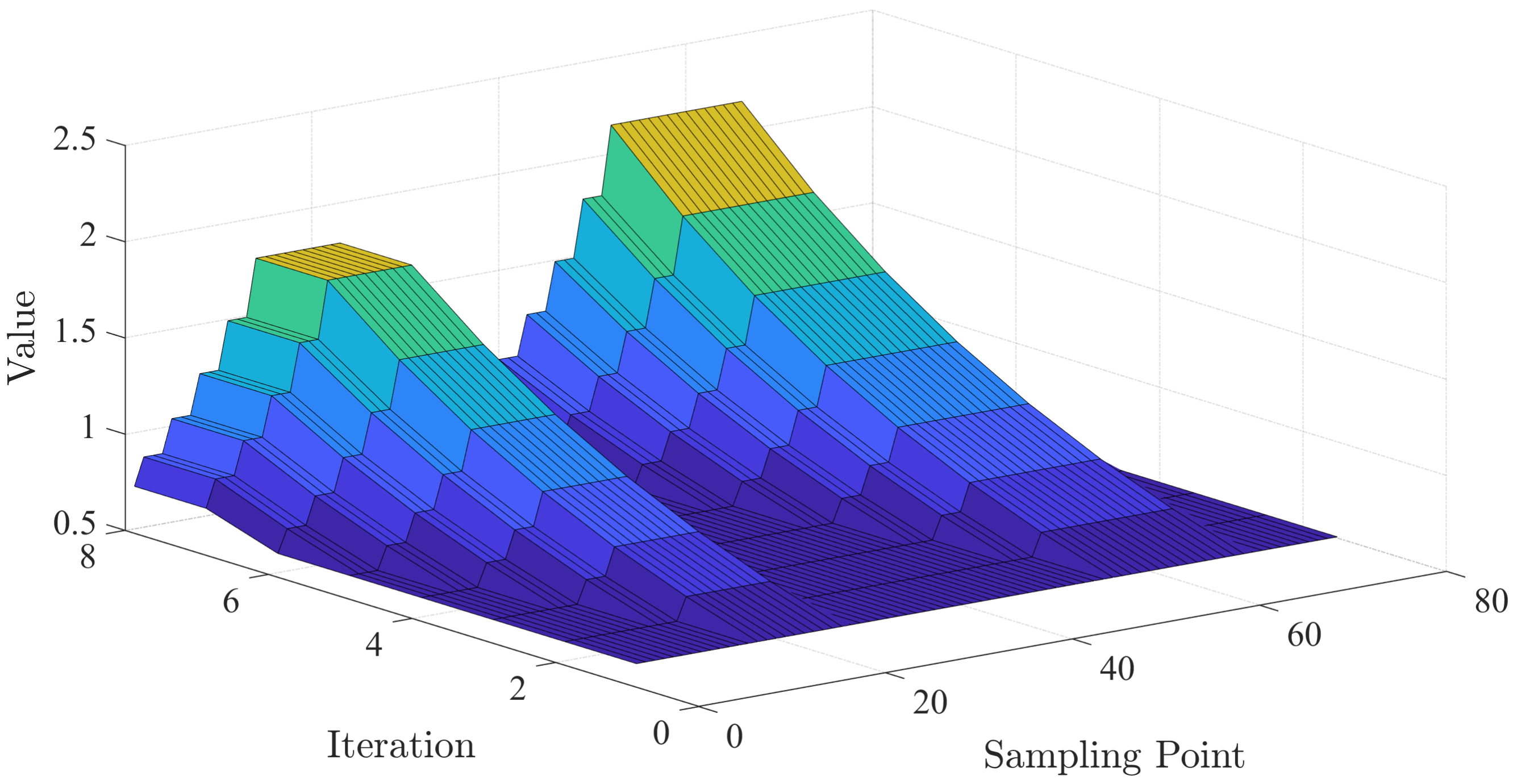}}
				\hfil
				\caption{Schematic illustration of the effectiveness of the ItCA method. In the first iteration, the planned path collides with obstacles, and after a finite number of iterations, ItCA path planning yields a collision-free path. The two peaks indicate the highest values of the error vector $\mathbf{e}$ corresponding to the two collision parts of the iterative path.}
			\end{figure}
		\end{graphicalabstract}

		\begin{highlights}
			
			\item \textcolor{black}{This paper presents a novel method for rapidly generating collision-free trajectories with high control feasibility in automated parking scenarios.}
			
			\item The precise collision avoidance method is in conflict with the time efficiency of trajectory planning. To address this issue, an efficient and reliable iterative collision avoidance framework is designed to ensure real-time collision-free trajectory planning by introducing parallel computing with explicit collision detection and avoidance. This method guarantees rapid trajectory planning under dense collision detection conditions, and the mean computation time of the proposed trajectory planning method ranges from \textbf{0.042} to \textbf{0.065}s in different scenarios.
			
			\item Planning paths that theoretically adhere to the vehicle model \textcolor{black}{presents} a challenge. To overcome this, the proposed method utilizes differential flatness to prove that the cubic polynomial is the minimum requirement to satisfy the vehicle kinematics model. The iterative collision avoidance (ItCA) path planning is then performed by manipulating polynomials satisfying the degree requirement so that the vehicle kinematic model can be fully integrated into the path planning framework to ensure the control feasibility of the final planned trajectory;
			
			\item The control feasibility of the planned trajectory at gear shifting points (GSP) is difficult to guarantee. To address this, the proposed method incorporates terminal smoothing constraints (TSC) into the path planning optimization and applies terminal kinematic constraints during velocity planning. This improves the control feasibility of the planned trajectories by \textbf{8.9\%}-\textbf{33.3\%} in different scenarios.
		\end{highlights}
		
		
	\end{frontmatter}

	\section{INTRODUCTION}\label{Introduction}
	\setcounter{figure}{0}
	Automated driving research has been of growing interest in recent years\cite{wang2014automatic,li2016time,li2015unified}. An effective autonomous driving system has the potential to not only alleviate the burden of driving for humans but also enhance the efficiency of traffic transportation. One of the most commonly used autonomous driving functions in daily life is automated parking, making research on trajectory planning for automated parking scenarios crucial and meaningful. Therefore, it is imperative to conduct further research in this field to develop efficient and optimized automated parking systems\cite{gonzalez2015review,paden2016survey}.
	
	Given the unstructured nature of automated parking scenarios, trajectory planning typically involves two essential steps: generating initial paths and planning refined  trajectories\cite{zhang2018autonomous}. Initial paths are typically generated through sampling or search methods without adequately incorporating considerations of control feasibility\cite{ajanovic2018search,hart1968formal,LaValle1998RapidlyexploringRT}. While these methods can generate initial paths quickly, they suffer from safety issues because it is challenging for the controller to track them accurately\cite{chaiDualLoopTubeBasedRobust2022,chaiAttitudeTrackingControl2022,zhu2015convex}. Therefore, precise trajectory planning is crucial for the success of the automated parking task\cite{zhang2020optimization,micheli2023nmpc,fan2018baidu}.  Model-free trajectory planning methods represented by path-velocity decomposition (PVD) \cite{kant1986toward} and model-integrated trajectory planning are two approaches to realize precise trajectory planning\cite{weng2024aggressive}.  A highly effective trajectory planning method should be capable of real-time planning\cite{fan2018baidu,zhou2020dl}.\textcolor{black}{Moreover, the planned trajectory is expected to demonstrate excellent control feasibility and collision-free characteristics.}\cite{zhou2020dl}.

	Collision avoidance is an essential factor in trajectory planning for automated parking\cite{zhang2018autonomous,zhang2020optimization}. \textcolor{black}{However}, realizing time-efficient and high-precision collision avoidance simultaneously is challenging. The model-integrated trajectory planning methods commonly use collision avoidance constraints in the optimization problem thereby implementing collision-free trajectories. But this approach is usually time inefficient because the mathematical transformation of collision avoidance constraints is generally non-linear/non-convex\cite{zhu2015convex,chaiMultiphaseOvertakingManeuver2023}, requiring significant computational resources\cite{schulman2014motion}.  Therefore, to address this issue, the PVD-based trajectory planning methods are utilized to reduce the complexity of collision avoidance constraints by conducting it into path planning using collision detection. This approach iteratively updates the planned path to ensure the collision-free characteristic. While this approach reduces the complexity of the optimization problem, it also introduces a further challenge. The collision detection within each iteration still consumes significant computation time. To address these challenges, this paper proposes a fast iterative collision avoidance (ItCA) path planning method. Different from conventional iteration-based methods, the proposed ItCA path planning method achieves high efficiency by distributing the computational workload of collision detection across multiple computational units. The key advantage of our proposed ItCA method is that collision avoidance constraints are decoupled from the optimization problem, and the computational burden caused by collision detection is distributed through a parallel computing architecture, thereby enhancing the time efficiency.  Based on the rapid iterative, the proposed ItCA method minimizes computational time while ensuring precise collision avoidance.
	
	The trajectory planning method should also guarantee the strong control feasibility of the planned trajectory. Firstly, to ensure excellent control feasibility of the planned trajectory, the accurate vehicle model is necessary\cite{zhang2021unified}. \textcolor{black}{However, the PVD-based trajectory planning methods struggle to integrate the vehicle model into the path planning process.} Secondly, due to the unique characteristics of automated parking scenarios, the planned trajectory often involves gear shifting points (GSP). However, the current trajectory planning method fails to consider the control feasibility of the planned trajectory at GSP\cite{zhu2015convex,fan2018baidu,zhou2020dl}, thereby reducing the overall control feasibility of the planned trajectory. To address these two issues, the PVD-based trajectory planning method proposed in this paper employs differential flatness to incorporate the accurate vehicle model into path planning, guaranteeing the control feasibility and ensuring the efficiency of solving the optimization problem. Additionally, this method improves the control feasibility of the planned trajectory at GSP by adding terminal smoothing constraints (TSC) in path planning and kinematic constraints at terminals in velocity planning. This method has the advantage of incorporating an accurate vehicle kinematics model into the path planning and improves the control feasibility of the planned trajectory at GSP through the TSC.

	In general, automated parking scenarios require the PVD-based trajectory planning method that ensures collision-free characteristics, time efficiency, and high control feasibility. However, integrating the vehicle model into real-time, collision-free path planning and ensuring the control feasibility of the planned trajectory at GSP pose significant challenges. The proposed method utilizes rapid and iteration-based path planning to minimize computational resources while ensuring precise collision avoidance. It also employs differential flatness to accurately satisfy the vehicle kinematics model, ensuring high control feasibility of the planned trajectory. Additionally, the proposed method imposes TSC on the planned path and terminal kinematic constraints on the velocity planning to improve control feasibility at GSP.

	\subsection{Motivations}\label{Motivations}
	
	The following three problems exist in the application of the PVD trajectory planning method to automated parking scenarios: 
	\begin{enumerate}

		\item The precise collision avoidance method is in conflict with the computational time of trajectory planning.
		\item Planning paths that theoretically adhere to the vehicle model \textcolor{black}{presents} a challenge.
		
		\item The control feasibility of the planned trajectory at GSP is difficult to guarantee.

	\end{enumerate}
	These three issues will directly affect the safety of vehicles when the PVD trajectory planning method is applied to real-world scenarios, so it is necessary and urgent to address them. 
	
	\begin{figure}[!t]
		\centering
		\includegraphics[scale=0.50]{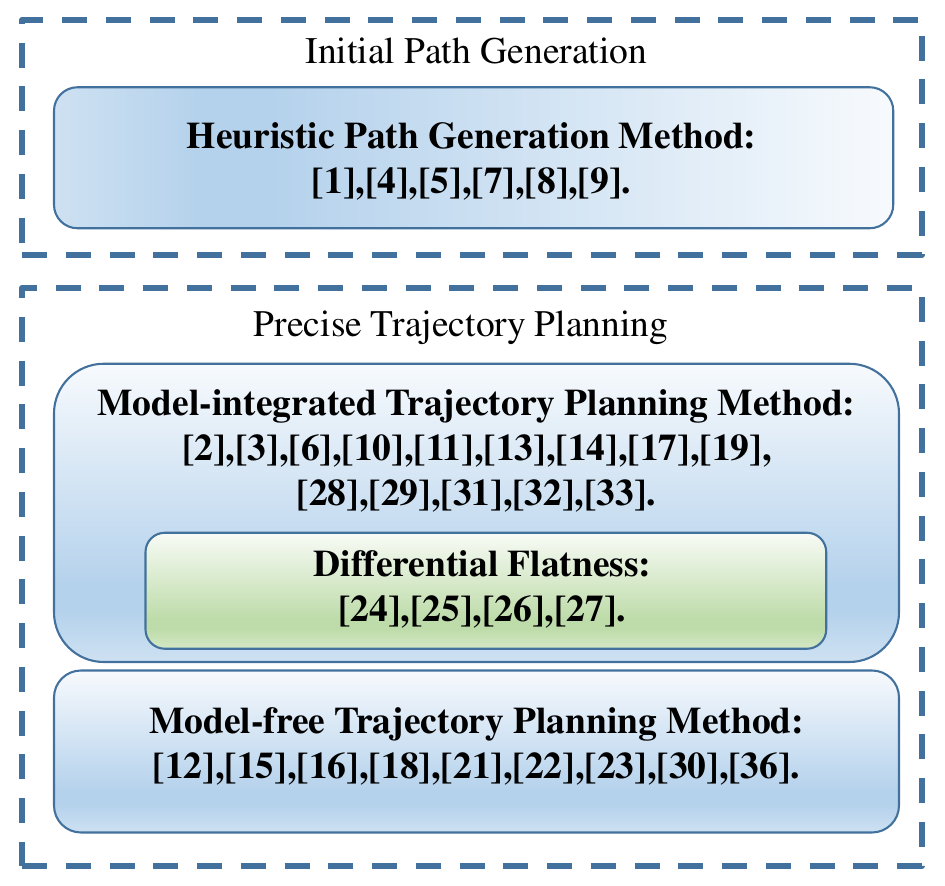}
		\caption{The general process and key technologies of trajectory planning.}
		\label{fig_Related_Works}
	\end{figure}
	
	\subsection{Contributions}\label{Motivations_and_Contributions}
	To tackle the problems mentioned in section \ref{Motivations}, this paper proposes a novel method to quickly generate strong control feasibility and collision-free trajectories for automated parking scenarios. This method is based on the pre-generated hybrid A* global collision-free path, and it offers the following contributions:
	
	\begin{enumerate}
		
		\item An efficient and reliable collision avoidance framework is designed to ensure real-time and collision-free trajectory planning with explicit collision detection and avoidance by introducing parallel computing. This method guarantees rapid trajectory planning under dense collision detection conditions, and the mean computation time of the proposed trajectory planning method ranges from \textbf{0.042} to \textbf{0.065}s in different scenarios.
		
		\item Differential flatness is used to prove that the cubic polynomial is the minimum requirement to satisfy the vehicle kinematics model. The ItCA path planning is then performed by manipulating polynomials satisfying the degree requirement so that the vehicle kinematic model can be fully integrated into the path planning framework to ensure the control feasibility of the final planned trajectory.
		
		\item The control feasibility of the planned trajectories at GSP is ensured by incorporating TSC into the path planning optimization and applying terminal kinematic constraints during velocity planning. This improves the control feasibility of the planned trajectories by \textbf{8.9\%}-\textbf{33.3\%}  in different scenarios.

	\end{enumerate}
	
	\subsection{Related Works}\label{Related_Works}
	
	Path planning is a crucial aspect of PVD trajectory planning. As mentioned in the previous section, integrating vehicle models to maintain control feasibility into path planning is challenging. \cite{fan2018baidu,zhou2020dl,werling2010optimal,liu2017convex} that rely on the point-mass model use curvature to estimate the control feasibility of the planned path, but it is unreliable when the planned path points are unevenly spaced\cite{zhu2015convex}. Moreover, integrating the vehicle kinematic model into the path-planning process presents challenges when using the point-mass model. Differential flatness is an effective method for simplifying path planning\cite{han2023efficient}.  Flat outputs for common systems are provided in \cite{murray1995differential}. The vehicle kinematic model can be incorporated into path planning by finding a reasonable flat output\cite{xu2021autonomous}. Planning flat outputs is a beneficial approach when dealing with flat systems. This method allows for the simultaneous planning of states and control inputs, ensuring control feasibility and simplifying the planning problem. The method proposed in \cite{rouchon1993flatness} uses the Cartesian coordinates of the last trailer as the flat output for motion planning in the standard n-trailer system due to the nature of differential flatness. In conclusion, differential flatness is an effective way to simplify path planning while maintaining control feasibility.
	
	The time-efficient and precise collision avoidance method is also essential for trajectory planning. The optimization-based collision avoidance (OBCA) methods\cite{zhang2018autonomous,zhang2020optimization} introduce dual variables and thus accurately represent the collision avoidance constraints without approximation. However, it is computationally expensive and neglects collision issues within sampling time. The geometric-based method\cite{zhang2021guaranteed} broadens the vehicle width to form a protection frame, ensuring collision-free trajectories throughout the planning horizon. Although collision avoidance can be effectively achieved using the virtual protection frame (VPF) method, it still incurs high computational costs due to the large number of collision avoidance constraints introduced in the optimization problem. \cite{li2015unified,chaiMultiphaseOvertakingManeuver2023} employ an effective method to realize collision avoidance by determining the relationship between the positions of vehicle vertices and obstacle vertices. Although they utilize the model-integrated trajectory planning method, thus ensuring control feasibility, the collision-free property between sampled points cannot be guaranteed. Implementing collision-free trajectories based on iterations is also an effective approach. The convex feasible set (CFS) method\cite{liu2017convex} employs an iterative approach to realize collision avoidance. While demonstrating convergence and time efficiency, it introduces approximations of control feasibility, leading to sharp trajectories at GSP, thus impacting the safety of the planned trajectory.  Overall, precise collision avoidance methods tend to be time-consuming, whereas faster approaches often sacrifice the control feasibility of the planned trajectory, which can negatively impact vehicle safety.
	
	Highly efficient and accurate collision detection methods are essential, whether integrated into the optimization problem or used in an iterative approach. Li and Shao\cite{li2015unified} proposed a precise method based on the geometric containment relationship between obstacle and vehicle vertexes. Projection distance\cite{yamaguchi2021model} is another practical collision detection method that maps the vehicle as a circle and does the same for obstacles. Collision avoidance can be achieved by constraining obstacles from entering the interior of the vehicle circle. Using finite circles\cite{micheli2023nmpc} to enclose the vehicle to maintain collision-free characteristics is also useful. Constructing a collision-free convex hull within the feasible region and planning trajectories within it ensures collision-free paths, which is also reasonable\cite{ding2019safe}. However, it is difficult to generate enough collision-free feasible regions in narrow scenarios such as parking. These collision detection methods are highly effective. However, integrating them into path planning while ensuring time efficiency and a high success rate is challenging.

	Learning-based methods are also an effective research area. An effective method for real-time optimal parking control using deep neural network (DNN) is presented in \cite{chaiDesignImplementationDeep2022}. However, the vehicle sizes in the actual vehicle experiments are small compared to the parking space, and the performance in narrow scenarios is not explored. The approach proposed in \cite{chaiDeepLearningBasedTrajectory2023} uses a similar method, but its validity is further enhanced by the use of a vehicle of comparable size to the parking space in the experiment. The framework proposed in \cite{chaiDesignExperimentalValidation2024} trains DNNs on pre-optimized state-action datasets, enabling online motion planning to iteratively predict optimal maneuvers. Deep reinforcement learning is then employed to solve motion control problems in uncertain environments. Learning-based methods usually perform well in real-world scenarios if the state distribution is similar to the training dataset. However, the effectiveness of the method is difficult to guarantee when the scenario changes significantly, such as the presence of a large number of obstacles.

	\subsection{Organization}\label{Organization}
	
	This paper consists of several sections. Section \ref{PRELIMINARIES} delineates the preprocessing method for the rough reference path generated by the hybrid A* algorithm and the explicit collision detection representation method. Section \ref{RITP} provides a detailed explanation of the RITP method, encompassing the processes of differential flatness and the ItCA path planning method, along with the velocity planning method incorporating vehicle kinematics constraints. Section \ref{EXPERIMENTAL} includes numerical simulations and experiments, while Section \ref{Conclusion} serves as the conclusion, summarizing the essential findings and outlining directions for future research. 
	
	\section{PRELIMINARIES}\label{PRELIMINARIES}
	
	This section introduces the preprocessing method for the reference path generated by the hybrid A* algorithm. Furthermore, we present an efficient representation of collision detection. The main notation used in the paper is summarized in Table~\ref{tab:Notation}.
	
	\subsection{Preprocessing Method for the Reference Path}\label{Preprocessing}
	
	The hybrid A*\cite{dolgov2010path} algorithm fully integrates the kinematic properties of the vehicle to obtain kinematically feasible paths, which includes the GSP. Consequently, the hybrid A* algorithm is employed to generate the reference path. After dividing the whole reference path into pieces according to GSP,  we use $\Upsilon$ to denote the entire reference path, where $\Upsilon_i$ is the i-th piecewise reference path. 
	
\begin{figure}[tbp]
		\centering
		\includegraphics[scale=0.26]{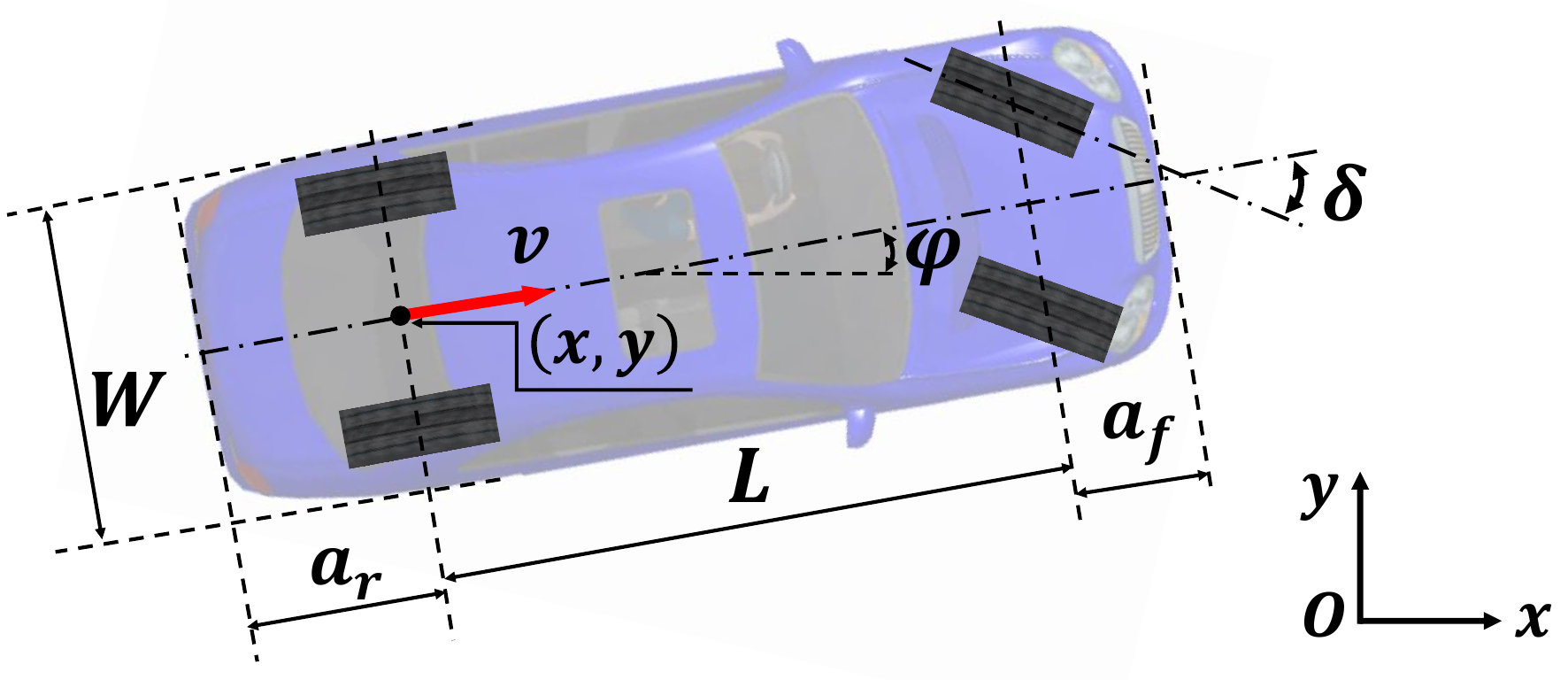}
		\caption{$L$ represents the wheelbase, while $W$ denotes the width of the vehicle. The distance from the front axle to the front of the vehicle is given by $a_f$, and $a_r$ refers to the distance from the rear axle to the rear of the vehicle.}
		\label{fig_model}
	\end{figure}

	As shown in Fig.~\ref{fig_model}, $x$, and $y$ are the horizontal and vertical coordinates of the mid of the rear axle, and $\varphi$ is the vehicle's yaw angle. The coordinate of the k-th point in the i-th piecewise reference path is denoted as $\mathbf{p}_{i,k}={\begin{bmatrix} {x}_{i,k} & {y}_{i,k} \end{bmatrix}}$, and the corresponding yaw angle is denoted as ${\varphi}_{i,k}$. The number of path points in $\Upsilon_i$ is represented as $N_i$.
	
	For $\Upsilon_i$, the calculation of the travel distance from the first point to the k-th$(2 \leq k \leq N_i)$ path point is: 
	\begin{equation}\label{eq:S}
		S_{i,k}=\sum_{k=2}^{N_i}{\Vert \mathbf{p}_{i,k} - \mathbf{p}_{i,k-1} \Vert}.
	\end{equation}
	
	After defining $S_{i,1}=0$, we approximate the vector $S_i$ representing the vehicle travel distance to the path length at each path point. 
	
	\begin{table}[!b]
		\centering
		\caption{Notations used in this paper.}
		\begin{adjustbox}{max width=8cm, max height=9cm}
			\begin{tabular}{l|l}
				\toprule
				\textbf{Indices} & \textbf{Meaning} \\
				\midrule
				\midrule
				i   & piecewise path index, $i \in [1,N]$\\
				\midrule
				j   & sampling point index on the polynomial \\
				\midrule
				k   & path point index on the reference path \\
				\midrule
				\textbf{Parameters} & \textbf{Meaning} \\
				\midrule
				\midrule
				$N$ & number of piecewise reference paths \\
				\midrule
				$\mathbf{p}_i$ & path points on the i-th reference path \\
				\midrule
				$N_i$ & number of path points on the i-th reference path \\
				\midrule
				\multirow{3}{*}{$S_i$}   & distance vector corresponding to the path point  \\
				\,    &   on i-th the reference path (from the initial point), $S_{i,\textrm{end}}$   \\
				\, &  is the  path length of the reference path\\
				\midrule
				\multirow{2}{*}{$\varphi_i$} & vehicle yaw angle vector corresponding to the path \\
				\, &  point on the i-th reference path\\
				\midrule
				$\bm{\xi}$ & flat output of the vehicle kinematics model \ref{eq:differential_equations} \\
				\midrule
				\multirow{2}{*}{$\mathbf{s}_i$} & distance vector corresponding to the sampling points on \\
				\,    &   the polynomial used for i-th piecewise path planning \\
				\midrule
				\multirow{2}{*}{$M_{i,\mathbf{s}}$} & number of sampling points on the polynomial used  \\
				\, &  for  i-th piecewise path planning\\
				\midrule
				\multirow{2}{*}{${\theta}_i$} & vehicle yaw angle vector calculated from sampling \\
				\,    &  points on the i-th planned path and \ref{eq:varphi_dot} \\
				\midrule
				\multirow{2}{*}{$\mathbf{t}_i$} & time vector corresponding to the sampling point on the \\
				\,    &  polynomial  used for i-th piecewise velocity planning \\
				\midrule
				\multirow{2}{*}{$M_{i,\mathbf{t}}$} & number of sampling points on the polynomial used  \\
				\,    & for i-th piecewise velocity planning\\
				\bottomrule
			\end{tabular}%
		\end{adjustbox}
		\label{tab:Notation}%
	\end{table}%

	\subsection{Collision Detection Method}\label{Collision Detection Method}
	
	For collision detection, we assume that both obstacles and vehicles are convex rectangles and the obstacles are static. 
	
	At the sampling point $j$, $\mathbb{E}(j)$ denotes the space occupied by the vehicle, $\mathbb{V}_{\textrm{vehicle}}(j)$ is the set of vertexes representing the vehicle. We assume that there are $n$ obstacles in the environment. Then $\mathbb{O}(c)$ refers to the space occupied by the c-th obstacle, $\mathbb{V}_{\textrm{obs}}(c)$ denotes the set of vertexes representing the c-th obstacle.
	
	\begin{figure}[t]
		\centering
		\includegraphics[scale=0.3]{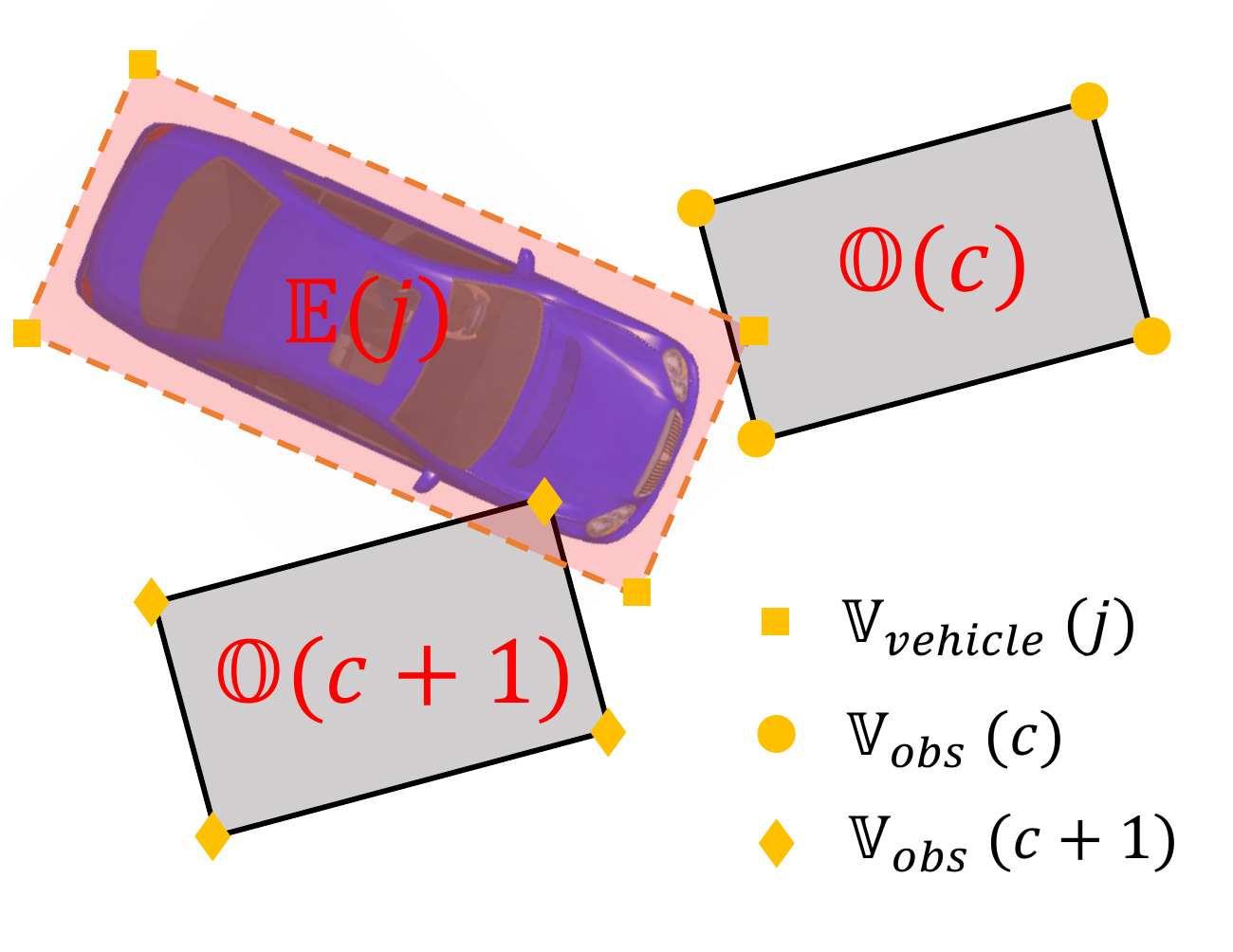}
		\caption{Schematic diagram of the collision detection method.}
		\label{fig:CA_diag}
	\end{figure}

	Then the collision avoidance constrains\cite{li2015unified} at the sampling point $k$ can be expressed as follows:
	\begin{equation}
		\begin{aligned}\label{collision_detection}
			&\mathbb{V}_{\textrm{vehicle}}(j) \cap \mathbb{O}(c) = \emptyset ,\\
			&\mathbb{E}(j) \cap  \mathbb{V}_{\textrm{obs}}(c) = \emptyset,\,\forall \, c=1,\cdots,n.
		\end{aligned}
	\end{equation}
	
	When the position relationship between the vehicle and all obstacles in the environment meets the above equation, it can be considered that the vehicle has not collided with obstacles (see Fig.~\ref{fig:CA_diag}).

	\section{The Rapid Iterative Trajectory Planning Method}\label{RITP}
	
	\begin{figure}[t]
		\centering
		\includegraphics[scale=0.6]{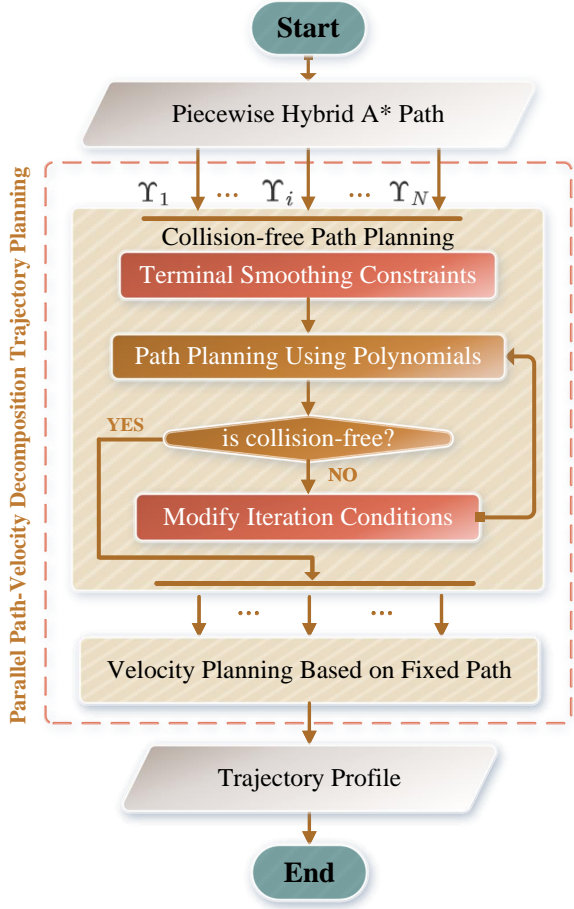}
		\caption{Flowchart of the rapid iterative trajectory planning method.}
		\label{fig_algo}
	\end{figure}

	In this section, the leading spirit and implementation of the rapid iterative trajectory planning (RITP) method will be introduced. The overall algorithm process is shown in Fig.~\ref{fig_algo}.
	
	\subsection{Overall Framework}\label{Overall_Framework}
	The RITP method generates piecewise reference paths using the hybrid A* algorithm. Moreover, parallel computing is introduced to enhance the efficiency of trajectory planning. The main components of the RITP method are as follows. Firstly, it is proved that the cubic polynomial is the minimum requirement to satisfy the vehicle kinematics model using differential flatness. Secondly, ItCA-based path planning using polynomials that satisfy the degree requirement is performed within each parallel computing unit based on piecewise reference paths. TSC is incorporated to enhance control feasibility, while collision avoidance is achieved by dynamically reconstructing the optimization problem throughout the iterations. Thirdly, in the velocity planning phase, vehicle kinematics constraints are considered in the optimization problem to improve the control feasibility of the planned trajectory at GSP. Finally, the piecewise planned trajectories are joined to create the complete parking trajectory.
	\subsection{ItCA Path Planning}\label{Path_Planning}

	This subsection introduces the ItCA path planning method based on the piecewise reference path within each parallel computing unit. For clarity, the piecewise reference path utilized in \textcolor{black}{the} parallel computing unit is denoted as $\Upsilon_i$ in the following content.
	
	\subsubsection{\bf Differential Flatness}\label{Differential_Flatness}

	Differential flatness is a valuable property widely utilized in motion planning \cite{rouchon1993flatness}. It has been employed by \cite{han2023efficient} for simplified vehicle trajectory planning tasks. 
	
	For the commonly used vehicle kinematic model:
	\begin{subequations}
		\label{eq:differential_equations}
		\begin{align}
			\dot{x}  &= \cos (\varphi) \cdot v\\ 
			\dot{y}&=  \sin (\varphi) \cdot v\\ 
			\dot{\varphi}&=  \tan (\delta)\cdot {L^{-1}} \cdot v \label{eq:varphi_dot}\\
			\dot{\delta}&=  \omega \label{eq:omega_dot}
		\end{align}
	\end{subequations}
	where $\omega$ is the steering angle rate. Let $\chi={\begin{bmatrix} x & y & \varphi  &\delta \end{bmatrix}}^{\intercal}$ represents the state of the vehicle, $u ={\begin{bmatrix} v & \omega \end{bmatrix}}^{\intercal}$ represents the control input.
	
	For a flat dynamic system, a flat output $\bm{{\xi}}$ can be found that satisfies the following conditions\cite{murray1995differential}:
	\begin{equation}
		\bm{{\xi}}=\gamma(\chi,u,\dot{u},\cdots,u^{(r)})
	\end{equation}
	such that,
	\begin{equation}
		\begin{split}
			\chi&=\Phi(\bm{{\xi}},\dot{\bm{{\xi}}},\cdots,\bm{{\xi}}^{(r)})\\
			u&=\phi(\bm{{\xi}},\dot{\bm{{\xi}}},\cdots,\bm{{\xi}}^{(r)})
		\end{split}
	\end{equation}
	where $r$ is the highest derivative involved. $\gamma,\Phi$ and $\phi$ are the mapping rules.
	
	For flat systems, planning flat outputs $\bm{{\xi}}$ allows for simultaneous planning of states and control inputs. The reasons are as follows. Firstly, the flat output can be directly mapped to control inputs by mapping $\phi$. Secondly, the planned flat output automatically satisfies the dynamical system due to the existence of $\Phi$ mapping.  Therefore, planning using the flat output $\bm{{\xi}}$ not only ensures control feasibility, but also simplifies the trajectory planning problem. 
	
	Next, it will be shown that the global coordinate of the rear axle center of the vehicle is the flat output of the vehicle kinematic model~\eqref{eq:differential_equations}, which is denoted as $\bm{{\xi}}:={\begin{bmatrix} \bm{{\xi}}_x & \bm{{\xi}}_y \end{bmatrix}}^{\intercal}$.
	
	The following equations can be obtained from the geometry:
	\begin{subequations}
		\begin{align}
			&\Phi_1(\bm{{\xi}})= {\begin{bmatrix} 1 & 0 \end{bmatrix}} \cdot \bm{{\xi}} = \bm{{\xi}}_x  = x \\
			&\Phi_2(\bm{{\xi}})= {\begin{bmatrix} 0 & 1 \end{bmatrix}} \cdot \bm{{\xi}} = \bm{{\xi}}_y = y \\
			&\Phi_3(\dot{\bm{{\xi}}})=  \arctan\left(\frac{\dot{\bm{{\xi}}_y}}{\dot{\bm{{\xi}}_x}}\right) =\varphi  \label{eq:varphi}
		\end{align}\label{eq:1}
	\end{subequations}
	
	In order to simplify the form of the expression, the forward movement of the vehicle is assumed here. Therefore, from the definition of velocity $v$, the following equation can be deduced:
	\begin{equation}
		\begin{aligned}
			\phi_1(\dot{\bm{{\xi}}}) = (\dot{\bm{{\xi}}_x}^2 + \dot{\bm{{\xi}}_y}^2) ^{\frac{1}{2}} =v\label{v_defi}
		\end{aligned}
	\end{equation}
	where the vehicle's backward movement is proved using the same approach.
	
	According to the~\eqref{eq:varphi_dot}, ~\eqref{eq:varphi}and~\eqref{v_defi}, it can be obtained that:
	\begin{equation}
		\begin{aligned}
			\Phi_4(\dot{\bm{{\xi}}},\ddot{\bm{{\xi}}}) = \arctan\left(\frac{\left(\ddot{\bm{{\xi}}_y}\dot{\bm{{\xi}}_x}-\ddot{\bm{{\xi}}_x}\dot{\bm{{\xi}}_y}\right)}{\left(\dot{\bm{{\xi}}_x}^2+\dot{\bm{{\xi}}_y}^2\right)^{\frac{3}{2}}} \cdot L \right) = \delta \label{delta}
		\end{aligned}
	\end{equation}
	
	Differentiating the~\eqref{delta}  according to~\eqref{eq:omega_dot} gives $\omega$:
	\begin{equation}
		\begin{split}
			&\phi_2(\dot{\bm{{\xi}}},\ddot{\bm{{\xi}}},\dddot{\bm{{\xi}}}) \\
			&= \frac{\left(\dddot{\bm{{\xi}}}_y\,\dot{\bm{{\xi}}}_x - \dddot{\bm{{\xi}}}_x\,\dot{\bm{{\xi}}}_y\right)\phi_1^{3}L
			}{{L^2(\ddot{\bm{{\xi}}}_y\,\dot{\bm{{\xi}}}_x - \ddot{\bm{{\xi}}}_x\,\dot{\bm{{\xi}}}_y)^2+\phi_1^6}}- \\
			&\frac{\left(\ddot{\bm{{\xi}}}_y\,\dot{\bm{{\xi}}}_x - \ddot{\bm{{\xi}}}_x\,\dot{\bm{{\xi}}}_y\right)\left(\ddot{\bm{{\xi}}}_x\,\dot{\bm{{\xi}}}_x + \ddot{\bm{{\xi}}}_y\,\dot{\bm{{\xi}}}_y\right)3\phi_1L}{L^2(\ddot{\bm{{\xi}}}_y\,\dot{\bm{{\xi}}}_x - \ddot{\bm{{\xi}}}_x\,\dot{\bm{{\xi}}}_y)^2+\phi_1^6} = \omega \label{eq:omega}
		\end{split}
	\end{equation}
	
	Following the derivation outlined above, it can be concluded that the vehicle kinematic model~\eqref{eq:differential_equations} is a flat system and $\bm{{\xi}}$ is the flat output when $\bm{{\xi}}$ is differentiable of order three. 
	
	Thus, the control input $u ={\begin{bmatrix} v & \omega \end{bmatrix}}^{\intercal}$ can be derived from the mapping $\phi=\{\phi_1(\dot{\bm{{\xi}}}),\phi_2(\dot{\bm{{\xi}}},\ddot{\bm{{\xi}}},\dddot{\bm{{\xi}}})\}$ and the flat output $\bm{\xi}$ and its derivatives.
	
	\subsubsection{\bf Iterative Path Planning Method with TSC}\label{Iterative Path Planning Method}
	
	Based on the aforementioned proof, the cubic polynomial meets the minimum requirements of vehicle kinematics. Therefore, polynomials whose order exceeds cubic can be utilized for path planning to ensure that the planned path adheres to the vehicle kinematics model. To avoid the Runge phenomenon caused by the excessive order of polynomials\cite{fornberg2007runge}, the highest order of polynomials is limited to quintic\cite{fan2018baidu,zhou2020dl,werling2010optimal}.

	To enhance the generalization ability of the ItCA method, polynomials meeting the following equations can be employed for path planning:
	\begin{equation}
		\begin{split}\label{eq:fxy}
			f_x(\mathbf{s}) &= a_{\alpha} \mathbf{s}^{\alpha} + a_{\alpha-1} \mathbf{s}^{\alpha-1} + \cdots + a_{1} \mathbf{s} + a_0 \\
			f_y(\mathbf{s}) &= b_{\alpha} \mathbf{s}^{\alpha} + b_{\alpha-1} \mathbf{s}^{\alpha-1} +\cdots + b_1 \mathbf{s} + b_0  \\
			& \alpha \in [3,5],\,\alpha \in \mathbb{Z} . \\
		\end{split}
	\end{equation}
	where $\mathbf{s}$ is the path length and $\alpha$ is the highest order of the polynomial. Then flat output $\bm{\xi}$ can be represented as ${\begin{bmatrix} f_x(\mathbf{s}) & f_y(\mathbf{s}) \end{bmatrix}}$.
	
	\begin{figure}[t]
		\centering
		\includegraphics[scale=0.23]{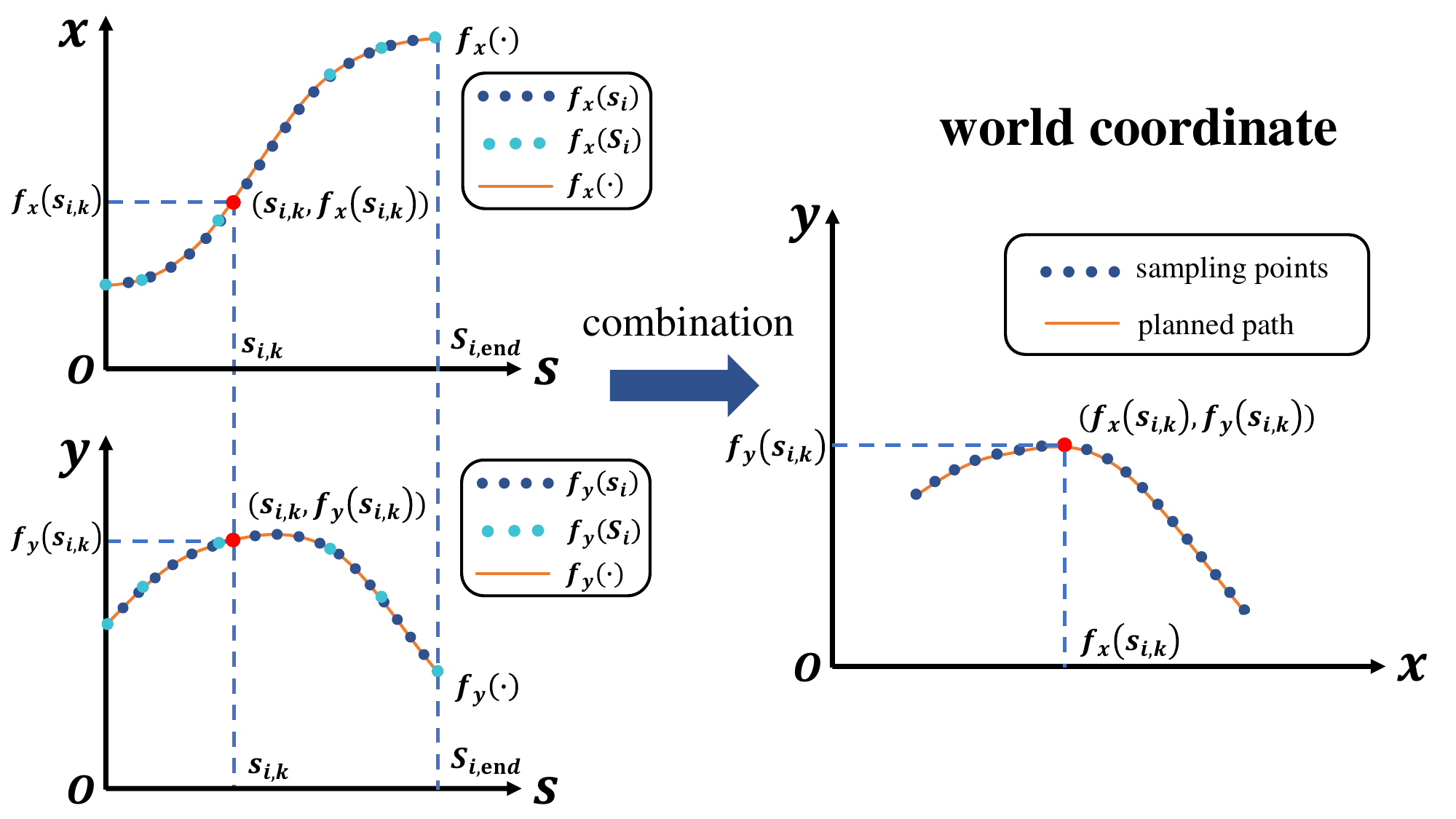}
		\caption{Illustration of the planned path. As shown in the figure, transforming $y(x)$ planning into $f_x(\mathbf{s})$ and $f_y(\mathbf{s})$ planning improves the maneuverability of the planned path.}
		\label{fig:JP_path}
	\end{figure}
	
	The collision iteration problem is considered next. The ItCA method performs collision detection by dense discrete sampling on the polynomials of the iterative path. Since a large number of sampling points $M_{i,\mathbf{s}}$ are used for collision detection, the distance between each sampling point on the planned path to equal length $\mathbf{d}_i$\cite{zhu2015convex}, which is represented as $\mathbf{d}_i:=S_{i,\textrm{end}} / (M_{i,\mathbf{s}}-1)$. 
	
	So the continuous path length $\mathbf{s}$ can be discretized into the distance vector $\mathbf{s}_i$ based on sampling points, which is expressed as: 
	\begin{equation}
		\begin{split}
			&\mathbf{s}_i={\begin{bmatrix} 0 & \cdots & g(j) & \cdots & g(M_{i,\mathbf{s}}) \end{bmatrix}}^{\intercal},\\
			&g(j)=(j-1)\cdot \mathbf{d}_i,1\leq j \leq M_{i,\mathbf{s}}
		\end{split}\label{eq:even_dis}
	\end{equation}
	where $\mathbf{s}_{i,j}$ denotes the j-th element in $\mathbf{s}_i$.
	
	\begin{figure}[!t]
		\centering
		\includegraphics[scale=0.25]{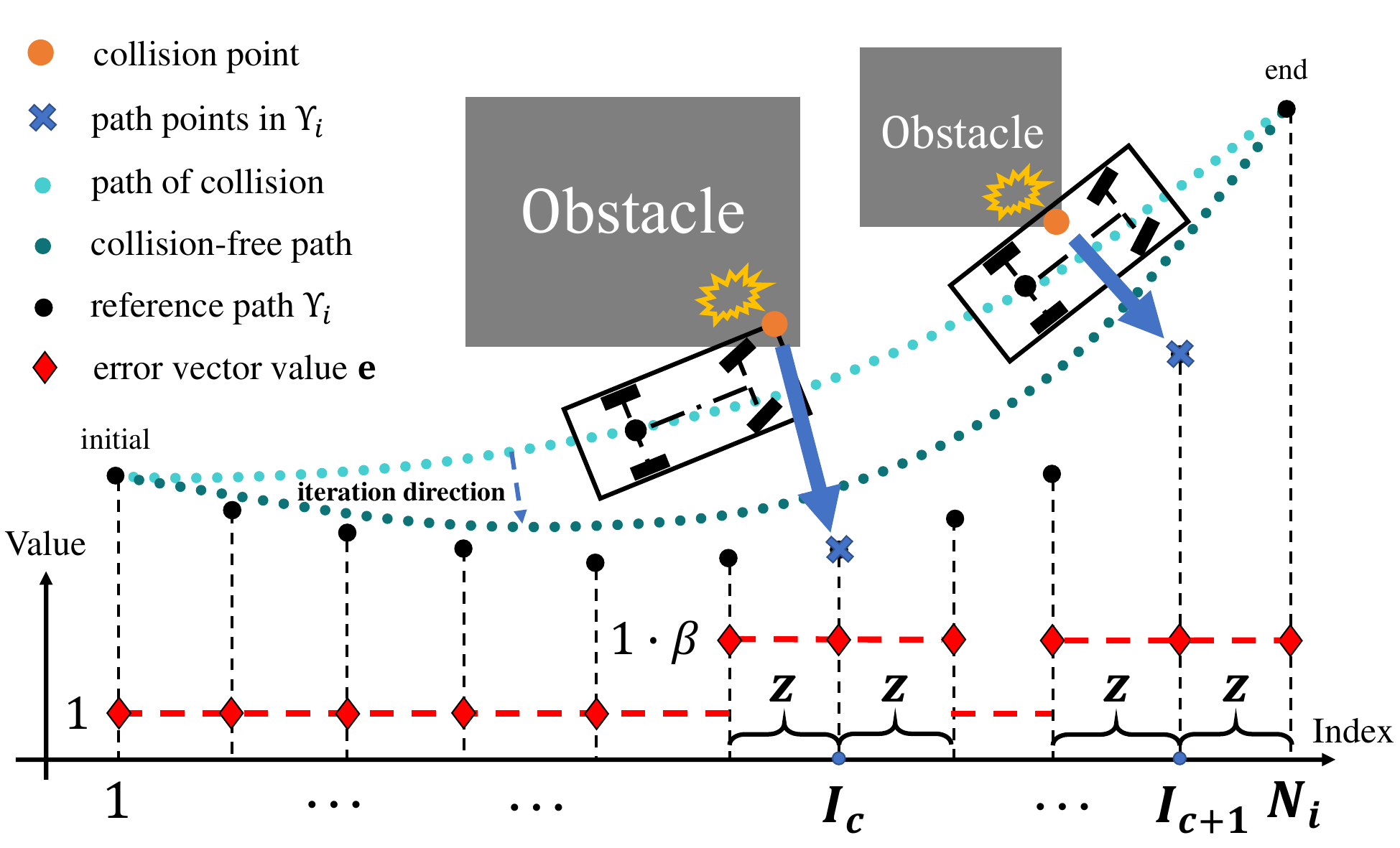}
		\caption{
			The schematic diagram illustrates the ItCA process. In the current iteration, the cyan path collides with the obstacle. Subsequently, a collision-free path represented in dark green is achieved in the next iteration by regenerating the optimization problem.}
		\label{fig_collision_free}
	\end{figure}
	
	The smoothness of the planned path and its similarity with the piecewise reference path $\Upsilon_i$ are usually mutually exclusive factors. Therefore, we take the balance of the two as the objective function of the optimization problem. We describe the similarity between the planned path and the $\Upsilon_i$ as the following expression:
	\begin{equation}
		\begin{split}
			J_1(k)=\lVert f_x(S_{i,k})- x_{i,k} \rVert ^2+ \lVert  f_y(S_{i,k}) - y_{i,k} \rVert  ^2, \\
			1 \leq k \leq N_i. \label{eq:j1k}
		\end{split}
	\end{equation}
	where $S_{i,k}$ is the path length corresponding to the k-th path point $\mathbf{p}_{i,k}={\begin{bmatrix} {x}_{i,k} & {y}_{i,k} \end{bmatrix}}$ in the i-th piecewise reference path according to~\eqref{eq:S}.
	
	The smoothness of the planned path is described as the following expression:
	\begin{equation}
		\begin{split}
			J_2(j)&=\lVert  \dot {f_x}(\mathbf{s}_k) \rVert ^2+ \lVert  \dot {f_y}(\mathbf{s}_k) \rVert ^2, \\ 
			J_3(j)&=\lVert  \ddot {f_x}(\mathbf{s}_k) \rVert ^2+ \lVert  \ddot {f_y}(\mathbf{s}_k) \rVert ^2 ,
			1 \leq j \leq M_{i,\mathbf{s}}. \label{eq:j2k}
		\end{split}
	\end{equation}
	
	Since the unknown quantities to be solved for are the coefficients of the polynomial, the~\eqref{eq:fxy}  is converted to the form of the matrix listed below:
	\begin{subequations}
		\label{eq:getx_y}
		\begin{align}
			\mathbf{F}_{i,x}= \mathbf{K}_{\mathbf{s}} \cdot \Xi_{i,x}, \label{eq:Fx}\\
			\mathbf{F}_{i,y}= \mathbf{K}_{\mathbf{s}} \cdot \Xi_{i,y}. \label{eq:Fy}
		\end{align}
	\end{subequations}
	where,
	\begin{equation}
		\begin{split}
			\mathbf{K}_{\mathbf{s}} &= {\begin{bmatrix} \mathbf{s}^{\alpha} & \mathbf{s}^{\alpha-1} & \cdots & \mathbf{s} & 1 \end{bmatrix}}, \\
			\Xi_{i,x}&= {\begin{bmatrix} a_{\alpha} & a_{\alpha-1} & \cdots & a_1 & a_0 \end{bmatrix}}^{\intercal}, \\
			\Xi_{i,y}&= {\begin{bmatrix} b_{\alpha} & b_{\alpha-1} & \cdots & b_1 & b_0 \end{bmatrix}}^{\intercal}.
		\end{split}
	\end{equation}

	The matrix of the derivative of the polynomial~\eqref{eq:Fx} with respect to the variable $\mathbf{s}$ can be expressed in the following form:
	\begin{equation}
		\begin{split}
			&{\mathbf{F}}_{i,x}^{'}=\frac {\partial \mathbf{F}_{x}} {\partial \mathbf{s}}=\frac {\partial \mathbf{K}_{\mathbf{s}}} {\partial \mathbf{s}} \cdot \Xi_{i,x} = {\mathbf{K}}_{\mathbf{s}}^{'}\cdot \Xi_{i,x}, \\
			&{\mathbf{F}}_{i,x}^{''}=\frac {\partial^2 \mathbf{F}_{x}} {\partial \mathbf{s}^2}=\frac {\partial^2 \mathbf{K}_{\mathbf{s}}} {\partial \mathbf{s}^2} \cdot \Xi_{i,x}= {\mathbf{K}}_{\mathbf{s}}^{''}\cdot \Xi_{i,x}.
		\end{split}
	\end{equation}

	For~\eqref{eq:Fy} can also be expressed in the same form:~${\mathbf{F}}_{i,y}^{'}={\mathbf{K}}^{'}_{\mathbf{s}}\cdot \Xi_{i,y},{\mathbf{F}}_{i,y}^{''}={\mathbf{K}}_{\mathbf{s}}^{''}\cdot \Xi_{i,y}$.

	Let $\Xi_i = {\begin{bmatrix} \Xi_{i,x} & \Xi_{i,y} \end{bmatrix}}$, then the summation of formulas~\eqref{eq:j1k} and~\eqref{eq:j2k} can then be written in the following form:
	\begin{equation}
		\begin{aligned}
			\sum_{k=1}^{N_i}
			\Vert \mathbf{K}_{S_{i,k}}\Xi_i-
			\mathbf{p}_{i,k}
			\Vert^2_{Q_1}
			+
			\sum_{j=1}^{M_{i,\mathbf{s}}}
			\Vert {\mathbf{K}}_{\mathbf{s}_{i,j}}^{'}\Xi_i
			\Vert^2_{Q_2}
			+
			\Vert {\mathbf{K}}_{\mathbf{s}_{i,j}}^{''}\Xi_i
			\Vert^2_{Q_3}
			\label{eq:sum_without_e}
		\end{aligned}
	\end{equation}
	where $Q_1$, $Q_2$ and $Q_3$ are the coefficient matrices.
	
	Equation~\eqref{eq:sum_without_e} represents the proximity of the polynomials to the reference path and its smoothness. Collision avoidance is achieved by increasing the proximity of the collision point on the planned path to the reference path. Therefore, the iterative path converges to the reference path when the collision is detected (see Fig.~\ref{fig_collision_free}). 
	
	First of all, the error vector $\mathbf{e} \in \mathbb{R}^{N_i \times 1}$ is defined to denote the weight of each path point corresponding to the reference path at the current iteration.

	Secondly, construct the diagonal matrix $\mathbf{D} \in \mathbb{R}^{N_i \times N_i}$:
	\begin{equation}\label{eq:D}
		\begin{split}
			\mathbf{D} =
			\begin{bmatrix}
				1 & 0 & \cdots & 0 & \cdots & 0 \\
				0 & {\beta} & \ddots & \ddots & \ddots & 0 \\
				%	\hline
				0 & \phantom{1}\ddots & {\beta} & \ddots & \ddots & 0 \\
				%	\hline
				0 & 0 & \ddots & {\beta} & \ddots & 0 \\
				%	\hline
				0 & 0 & \ddots & \ddots & \ddots & 0 \\
				0 & \cdots & 0 & \cdots & \cdots & 1 \\
				%	\rule{5em}{0.7pt} & \rule{2.5em}{0.7pt} & \rule{2.5em}{0.7pt} & \rule{2.5em}{0.7pt} & \rule{2.5em}{0.7pt} & \rule{2.5em}{0.7pt} \\
			\end{bmatrix}
			\begin{array}{c}
				\vphantom{\dfrac{1}{2}}\textrm{row } I_c-\mathbf{z} \\
				\vdots \\
				\vphantom{\dfrac{1}{2}}\textrm{row } I_c+\mathbf{z} \\
				\\
			\end{array}
		\end{split}
	\end{equation}
	where the value from row $I_c-\mathbf{z}$ to row $I_c+\mathbf{z}$ on the main diagonal is $\beta$, which is the expansion factor. $I_c$ denotes the index of the nearest point on the reference path to the collision point of the planned path with the c-th obstacle.
	
	Then $\mathbf{e}$ updates when collision occurs using the following expression:
	\begin{equation}\label{eq:update_e}
		\begin{split}
			\mathbf{e} =  \mathbf{D}  \cdot \mathbf{e}_{\textrm{last}} 
		\end{split}
	\end{equation}
	where $\mathbf{e}_{\textrm{last}}$ is the value of the error vector at the last iteration. In the first iteration, the error vector $\mathbf{e}$ is initialized with all values set to 1.
	
	\renewcommand{\algorithmicrequire}{\textbf{Input:}}
	\renewcommand{\algorithmicensure}{\textbf{Output:}}
	\begin{algorithm}[H]
		\caption{ItCA Path Planning Method with TSC}\label{alg:ICA_process}
		\begin{algorithmic}[1]
			\REQUIRE{piecewise hybrid A* path:\,$\Upsilon_i$,\; initial diagonal matrix $\mathbf{D}$,\; initial error vector $\mathbf{e}$,\; smoothing order $m$ of TSC.}
			\WHILE {True}
			\STATE  $\mathbf{\xi}_i \gets$  QP problem$\eqref{eq:pathplanning}$($\Upsilon_i,\mathbf{D},\mathbf{e},m$)
			\FORALL {$\mathbf{s}_{i,j} \in \mathbf{s}_{i}$}
			\STATE  Calculate vehicle pose $[\bm{\xi}_{i,x},\, \bm{\xi}_{i,y},\, {\theta}_{i}]$ using $\mathbf{d}_{i}$, $\mathbf{s}_{i,j}$, and $\eqref{eq:varphi_dot}$
			\IF  {Collision occurs during collision detection using~\eqref{collision_detection}}
			\STATE {Update diagonal matrix $\mathbf{D}$ using $\eqref{eq:D}$}
			\STATE {Calculate the error vector $\mathbf{e}$ using~\eqref{eq:update_e}}
			\STATE \textbf{break} \quad \quad \% The iteration path collides, 
			\STATE \quad \quad \quad \% reconstructing the QP problem
			\ENDIF
			\ENDFOR
			\IF  {The planned path $\bm{\xi}_i$ is collision-free}
			\STATE \textbf{return} \% The iteration path is collision-free
			\ENDIF
			\ENDWHILE
			\ENSURE{$\textrm{Collision-free path pose sequences}\, [\bm{\xi}_{i,x},\, \bm{\xi}_{i,y},\, {\theta}_{i}]$}
		\end{algorithmic}
	\end{algorithm}
	
	%Therefore, it is necessary to consider reflecting this part in Formula 1.
	The error vector $\mathbf{e}$ needs to be taken into account in the optimization objective~\eqref{eq:sum_without_e}. The following expression can be obtained:
	\begin{equation}
		\begin{aligned}
			\sum_{k=1}^{N_i}
			\mathbf{e}\Vert \mathbf{K}_{S_{i,k}}\Xi_i
			-
			\mathbf{p}_{i,k}
			\Vert^2_{Q_1}
			+
			\sum_{j=1}^{M_{i,\mathbf{s}}}
			\Vert {\mathbf{K}}_{\mathbf{s}_{i,j}}^{'}\Xi_i
			\Vert^2_{Q_2}
			+
			\Vert {\mathbf{K}}_{\mathbf{s}_{i,j}}^{''}\Xi_i
			\Vert^2_{Q_3}
		\end{aligned}
	\end{equation}
	
	According to the previously described content, the ItCA method ensures that the planned path converges to a collision-free path by dynamically adjusting the value of the error vector $\mathbf{e}$ during the iteration process by reconstructing the optimization problem. Next, the smoothness at the terminals of the planned paths is considered.
	
	The high-order curvature continuity maintained at GSP ensures the control feasibility of the planned path. The TSC is used to realize this objective.

	Ideally, the curvature of the planned path should satisfy~\eqref{eq:varphi}, but this constraint introduces strong nonlinearities in the optimization problem, thus reducing the computational efficiency.
	
	To solve this problem, ~\eqref{eq:varphi} can be simplified through approximation as:
	\begin{equation}
		\begin{split}\label{eq:delta_discre}
			\theta_{i,j} = 
			\arccos & \left(  {\left(\mathbf{K}_{\mathbf{s}_{i,j+1}} - \mathbf{K}_{\mathbf{s}_{i,j}} \right)  \Xi_{i,x}} \cdot \Vert ({\mathbf{K}}_{\mathbf{s}_{i,j+1}} - {\mathbf{K}}_{\mathbf{s}_{i,j}})\Xi_i  \Vert^{-1}  \right) \\
			\approx  \arccos & \left(  {\left(\mathbf{K}_{\mathbf{s}_{i,j+1}} - \mathbf{K}_{\mathbf{s}_{i,j}} \right)  \Xi_{i,x}}\cdot{ \mathbf{d}_i}^{-1}   \right), 
			1 \leq j \leq M_{i,\mathbf{s}}-1.
		\end{split}
	\end{equation}
	where $\theta_{i,j}$ is the yaw angle of the j-th path point on the planned path sampled according to $\mathbf{s}_{i,j}$ and $\theta_{i,\text{end}}=\theta_{i,\text{end-1}}$. This approximation is reasonable, as it assumes that the distances between adjacent sampling points along the planned path are approximately equal, as described in~\eqref{eq:even_dis}.

	In order to achieve the continuity of yaw angle $\theta$  at GSP, the following constraints are required to be satisfied:
	\begin{equation}
		\begin{split}\label{eq:delta_discre_0}
			&\Delta \theta_{i,1} = \Delta \theta_{i,{\text{end}}} = 0 \\
		\end{split}
	\end{equation}
	where $\Delta \theta_{i,1} = \theta_{i,1} - {{\varphi}}_{i,1}$ and $\Delta \theta_{i,\text{end}} = \theta_{i,\text{end}} - {{\varphi}}_{i,\text{end}}$. Since this constraint is only used to control the curvature continuity of the planned path at GSP without affecting the order of the polynomials, the planned path still satisfies the differential flatness property.

	To transform~\eqref{eq:delta_discre_0} into linear constraints, it is sufficient to ensure that the two terminal points of the planned path coincide with the two terminal points of the reference path. Then, the angle formed by the alignment points with terminals can be approximated as the yaw angle.

	Therefore,~\eqref{eq:delta_discre} and~\eqref{eq:delta_discre_0} can then be approximated by the following equation:
	\begin{equation}
		\begin{split}\label{eq:noextend_yaw}
			&\cos(\varphi_{i,1}) = \zeta{\left(\mathbf{K}_{\mathbf{s}_{i,2}}
				\Xi_i
				-
				\mathbf{p}_{i,1}
				\right)
				\cdot{\begin{bmatrix}  
						1 \\
						0
				\end{bmatrix}} \cdot (\mathbf{d}_i )^{-1}}  \\
			&\cos(\varphi_{i,\text{end}}) = -\zeta{\left(\mathbf{K}_{\mathbf{s}_{i,\text{end}-1}}
				\Xi_i
				-
				\mathbf{p}_{i,\text{end}}
				\right)
				\cdot{\begin{bmatrix}  
						1 \\
						0
				\end{bmatrix}} \cdot (\mathbf{d}_i )^{-1}}  
		\end{split}
	\end{equation}
	where $\zeta$ is to indicate the direction of operation of the vehicle. When moving forward $\zeta=1$, and when moving backward $\zeta=-1$. 
	
	Achieving higher-order smoothing at GSP can be accomplished by increasing the distance between the terminal and alignment path points.
	
	Therefore, equation~\eqref{eq:noextend_yaw} can be modified as shown below to ensure a higher-order degree of smoothness:
	\begin{equation}
		\begin{split}\label{eq:TSC}
			&\cos(\varphi_{i,1}) = \zeta{\left(\mathbf{K}_{\mathbf{s}_{i,1+m}}
				\Xi_i
				-
				\mathbf{p}_{i,1}
				\right)
				\cdot{\mathbf{b}} \cdot (m*  \mathbf{d}_i )^{-1}}  \\
			&\cos(\varphi_{i,\text{end}}) = -\zeta{\left(\mathbf{K}_{\mathbf{s}_{i,\text{end}-m}}
				\Xi_i - \mathbf{p}_{i,\text{end}} \right) \cdot 
				\mathbf{b}} \cdot  {(m \cdot \mathbf{d}_i)^{-1}}
		\end{split}
	\end{equation}
	where $1 \leq m$ represents the smoothing order of TSC and $\mathbf{b} = {\begin{bmatrix} 1 & 0 \end{bmatrix}}^{\intercal}$. Equation~\eqref{eq:TSC} is the TSC formulation of the ItCA path planning phase.
	
	Smoothing constraints at GSP are also used at the terminals of the planned paths in \cite{zhu2015convex} and \cite{zhou2020dl}, but are limited to first-order smoothing without considering the higher-order smoothing. The proposed TSC with high-order smoothing can effectively enhance the control feasibility of the planned trajectory.
	
	At this point, the optimization problem for generating the iterative path is fully constructed. The structure of optimization problems is expressed in the following form:
	\begin{equation}
		\begin{aligned}\label{eq:pathplanning}
			\min_{\Xi_i} \quad & \sum_{k=1}^{N_i} \mathbf{e} \Vert \mathbf{K}_{S_{i,k}}\Xi_i - \mathbf{p}_{i,k} \Vert^2_{Q_1}  \\
			& + \sum_{j=1}^{M_{i,\mathbf{s}}} \| {\mathbf{K}}_{\mathbf{s}_{i,j}}^{'}\Xi_i\Vert^2_{Q_2} + \| {\mathbf{K}}_{\mathbf{s}_{i,j}}^{''}\Xi_i\Vert^2_{Q_3}  \\
			\text{s.t.} \quad 
			& \cos(\varphi_{i,1}) =  \zeta{\left(\mathbf{K}_{\mathbf{s}_{i,1+m}} \Xi_i - \mathbf{p}_{i,1}\right) \cdot \mathbf{b}} \cdot  {(m \cdot \mathbf{d}_i)^{-1}}   \\ 
			& \cos(\varphi_{i,\text{end}}) = -\zeta{\left(\mathbf{K}_{\mathbf{s}_{i,\text{end}-m}} \Xi_i - \mathbf{p}_{i,\text{end}}\right) \cdot \mathbf{b}} \cdot  {(m \cdot \mathbf{d}_i)^{-1}}   \\
			& \mathbf{p}_{i,1} = \mathbf{K}_{\mathbf{s}_{i,1}} \Xi_i, \;  \mathbf{p}_{i,\text{end}} = \mathbf{K}_{\mathbf{s}_{i,\text{end}}} \Xi_i,1 \leq m 
		\end{aligned}
	\end{equation}

	Equation~\eqref{eq:pathplanning} is a classical quadratic programming (QP) problem that can be solved effectively, enabling fast iterative path planning.

	The planned path is generated during each iteration, and the vehicle's pose is computed using sampling points $\mathbf{s}_i$. Collision detection is then performed, and if the collision is detected, the error vector $\mathbf{e}$ is calculated using~\eqref{eq:D} and~\eqref{eq:update_e} for collision avoidance. The optimization problem~\eqref{eq:pathplanning} is then reconstructed to generate a replacement planned path. This iterative process continues until the collision-free path is ultimately generated. The overall process is illustrated in Algorithm~\ref{alg:ICA_process}.

	After the iterative collision-free path generation in this section, the discrete vehicle poses $[\bm{\xi}_{i,x},\, \bm{\xi}_{i,y},\, {\theta}_{i}]$ are obtained. The planned collision-free path adheres to the vehicle's non-holonomic kinematic constraints, which are guaranteed by differential flatness. In the following section, velocity planning will be conducted based on the planned path.

	\subsection{Velocity Planning Based on Fixed Path}\label{speed_Planning}

	In this section, we construct the optimization problem for velocity planning based on the path obtained from the path planning phase.

	Inspired by \cite{qi2022hierarchical}, the quintic polynomial is used to represent the relationship between the path length and maneuvering time $\mathbf{t}$ as follows:
	\begin{equation}
		\begin{split}\label{eq:ft}
			f_t(\mathbf{t}) &= c_5 \mathbf{t}^5 + c_4 \mathbf{t}^4 + c_3 \mathbf{t}^3 + c_2 \mathbf{t}^2  + c_1 \mathbf{t} + c_0 \\
			%	,s\in S_{i}
		\end{split}
	\end{equation}
	
	The maximum maneuver time of the vehicle is a hyperparameter, which is denoted as $\mathbf{t}_{i,{\textrm{max}}}$. $M_{i,\mathbf{t}}$ is the number of sampling points and satisfies $ M_{i,\mathbf{t}} = \lfloor  M_{i,\mathbf{s}} \cdot \lambda \rfloor,\lambda<1$. 
	
	So the interval of time is defined as $\Delta \mathbf{t}_i:=\mathbf{t}_{i,\textrm{max}}/(M_{i,\mathbf{t}} - 1)$.  Then the continuous maneuvering time $\mathbf{t}$ can be discretized into the time vector $\mathbf{t}_i$ based on sampling points, which is expressed as:
	\begin{equation*}
		\begin{split}
			&\mathbf{t}_i={\begin{bmatrix} 0 & \cdots & g_\mathbf{t}(j) & \cdots & g_\mathbf{t}(M_{i,\mathbf{t}}) \end{bmatrix}}^{\intercal},\\
			&g_\mathbf{t}(j)=(j-1)\cdot \Delta \mathbf{t}_i,1\leq j \leq M_{i,\mathbf{t}}
		\end{split}
	\end{equation*}
	where $\mathbf{t}_{i,j}$ denotes the j-th element in $\mathbf{t}_i$.
	
	The~\eqref{eq:ft} can also be expressed in matrix form:
	\begin{equation}
		\begin{split}\label{eq:Ft}
			\mathbf{F}_{i,\mathbf{t}}= \mathbf{K}_{\mathbf{t}} \cdot \Xi_{i,\mathbf{t}}
		\end{split}
	\end{equation}
	where:
	\begin{equation}
		\begin{split}
			\mathbf{K}_{\mathbf{t}} &= {\begin{bmatrix} \mathbf{t}^5 & \mathbf{t}^4 & \mathbf{t}^3 & \mathbf{t}^2 & \mathbf{t} & 1 \end{bmatrix}}, \\
			\Xi_{i,\mathbf{t}}&= {\begin{bmatrix} c_5 & c_4 & c_3 & c_2 & c_1 & c_0 \end{bmatrix}}^{\intercal}.
		\end{split}
	\end{equation}
	
	The matrix of the derivative of the quintic polynomial~\eqref{eq:Ft} with respect to the variable $\mathbf{t}$ can be expressed in the following form:
	\begin{equation}
		\begin{split}
			&\dot{\mathbf{F}}_{i,\mathbf{t}}=\frac {\partial \mathbf{F}_{i,\mathbf{t}}} {\partial \mathbf{t}}=\frac {\partial \mathbf{K}_{\mathbf{t}}} {\partial \mathbf{t}} \cdot \Xi_{i,\mathbf{t}} = \dot{\mathbf{K}}_{\mathbf{t}}\cdot \Xi_{i,\mathbf{t}} \\
			&\ddot{\mathbf{F}}_{i,\mathbf{t}}=\frac {\partial^2 \mathbf{F}_{i,\mathbf{t}}} {\partial \mathbf{t}^2}=\frac {\partial^2 \mathbf{K}_{\mathbf{t}}} {\partial \mathbf{t}^2} \cdot \Xi_{i,\mathbf{t}} = \ddot{\mathbf{K}}_{\mathbf{t}}\cdot \Xi_{i,\mathbf{t}} \\
			&\dddot{\mathbf{F}}_{i,\mathbf{t}}=\frac {\partial^3 \mathbf{F}_{i,\mathbf{t}}} {\partial \mathbf{t}^3}=\frac {\partial^3 \mathbf{K}_{\mathbf{t}}} {\partial \mathbf{t}^3} \cdot \Xi_{i,\mathbf{t}} = \dddot{\mathbf{K}}_{\mathbf{t}}\cdot \Xi_{i,\mathbf{t}} \\
		\end{split}
	\end{equation}
	
	To ensure the smoothness of the quintic polynomial, the smoothness of $f_t(\mathbf{t})$  is used as part of the optimization objective:
	\begin{equation}
		\begin{split}
			\sum_{j=1}^{M_{i,\mathbf{t}}}
			\Vert \dot{\mathbf{K}}_{\mathbf{t}_{i,j}}\Xi_{i,\mathbf{t}} \Vert^2_{R_1} +\Vert \ddot{\mathbf{K}}_{\mathbf{t}_{i,j}}\Xi_{i,\mathbf{t}} \Vert^2_{R_2}+\Vert \dddot{\mathbf{K}}_{\mathbf{t}_{i,j}}\Xi_{i,\mathbf{t}} \Vert^2_{R_3}
		\end{split}
		\label{eq:v_cost}
	\end{equation}
	where $R_1$, $R_2$ and $R_3$ are the coefficient matrices.
	
	Now, consider the constraints in velocity planning. First, reaching the terminal from the initial pose within the specified time is necessary. So, terminal constraints of velocity planning can be expressed as:
	\begin{equation}
		\begin{split}\label{eq:cons1}
			&{\mathbf{K}}_{\mathbf{t}_{i,1}}\Xi_{i,\mathbf{t}} = 0, \;{\mathbf{K}}_{\mathbf{t}_{i,\text{end}}}\Xi_{i,\mathbf{t}} = S_{i,\textrm{end}}
		\end{split}
	\end{equation}
	
	\newcommand{\myplus}[1][blue]{%
		\tikz[baseline=-0.5ex]\draw[#1, line width=0.5mm] (0,0) -- (1ex,0) (0.5ex,-0.5ex) -- (0.5ex,0.5ex);
	}
	\newcommand{\bluedot}{\textcolor{blue}{\textbullet}\;}
	\newcommand{\myrect}[1][red]{%
		\tikz[baseline=0ex]\filldraw[#1, line width=0.5mm] (0,0) rectangle (1ex,1ex);
	}

	Similar to~\cite{han2023efficient}, in order to improve the control feasibility of the planned trajectory at GSP, the velocity and acceleration are set to be 0, utilizing the following kinematic constraints:
	\begin{equation}
		\begin{split}\label{eq:cons2}
			&\dot{\mathbf{K}}_{\mathbf{t}_{i,1}}\Xi_{i,\mathbf{t}} = \ddot{\mathbf{K}}_{\mathbf{t}_{i,1}}\Xi_{i,\mathbf{t}} =\dot{\mathbf{K}}_{\mathbf{t}_{i,\text{end}}}\Xi_{i,\mathbf{t}} =\ddot{\mathbf{K}}_{\mathbf{t}_{i,\text{end}}} \Xi_{i,\mathbf{t}} = 0
		\end{split}
	\end{equation}
	
	It is also necessary to make the planned velocity satisfy the constraints of maximum velocity and maximum acceleration by the following constraints:
	\begin{equation}
		\begin{split}\label{eq:cons3}
			\lvert \dot{\mathbf{K}}_{\mathbf{t}_{i,j}}\Xi_{i,\mathbf{t}} \rvert \leq v_{\text{max}} , \lvert \ddot{\mathbf{K}}_{\mathbf{t}_{i,j}}\Xi_{i,\mathbf{t}} \rvert \leq a_{\text{max}},\\ \text{for}\,\,j=2,\cdots,M_{i,\mathbf{t}}-1 
		\end{split}
	\end{equation}

	Combining~\eqref{eq:v_cost},~\eqref{eq:cons1} ,~\eqref{eq:cons2} and~\eqref{eq:cons3} means that the optimization problem of velocity planning can be constituted as follows:
	\begin{align}\label{eq:speedplanning}
		\min_{\Xi^\mathbf{\mathbf{t}}_i} \quad & \sum_{j=2}^{M_{i,\mathbf{t}}-1}
		\Vert \dot{\mathbf{K}}_{\mathbf{t}_{i,j}}\Xi_{i,\mathbf{t}} \Vert^2_{R_1} + \Vert \ddot{\mathbf{K}}_{\mathbf{t}_{i,j}}\Xi_{i,\mathbf{t}} \Vert^2_{R_2} + \Vert \dddot{\mathbf{K}}_{\mathbf{t}_{i,j}}\Xi_{i,\mathbf{t}} \Vert^2_{R_3}  \nonumber \\
		%	&+ \Vert \dddot{\mathbf{K}}_{\mathbf{t}_{i,j}}\Xi_{i,\mathbf{t}} \Vert^2_{R_3} \nonumber  \\
		\text{s.t.} \quad & {\mathbf{K}}_{\mathbf{t}_{i,1}}\Xi_{i,\mathbf{t}} = 0, {\mathbf{K}}_{\mathbf{t}_{i,\text{end}}}\Xi_{i,\mathbf{t}} = S_{i,\textrm{end}}  \\
		& \dot{\mathbf{K}}_{\mathbf{t}_{i,1}}\Xi_{i,\mathbf{t}} = \ddot{\mathbf{K}}_{\mathbf{t}_{i,1}}\Xi_{i,\mathbf{t}} =\dot{\mathbf{K}}_{\mathbf{t}_{i,\text{end}}}\Xi_{i,\mathbf{t}} =\ddot{\mathbf{K}}_{\mathbf{t}_{i,\text{end}}} \Xi_{i,\mathbf{t}} = 0  \nonumber \\
		& \lVert \dot{\mathbf{K}}_{\mathbf{t}_{i,j}}\Xi_{i,\mathbf{t}} \rVert \leq v_{\text{max}} , \lVert \ddot{\mathbf{K}}_{\mathbf{t}_{i,j}}\Xi_{i,\mathbf{t}} \rVert \leq a_{\text{max}} \nonumber \\
		& \text{for}\,\,j=2,\cdots,M_{i,\mathbf{t}}-1 \nonumber
	\end{align}
	where~\eqref{eq:speedplanning} is a classical QP problem that can be solved quickly.
	
	The final planned velocity $v_i \in \mathbb{R}^{M_{i,\mathbf{t}} \times 1}$ is obtained by the following equation:
	\begin{equation}
		\begin{split}
			v_i = \dot{\mathbf{K}}_{\mathbf{t}_{i}}\Xi_{i,\mathbf{t}} \cdot \zeta
		\end{split}
	\end{equation}

	By utilizing the RITP method described earlier, the parallel computing unit can generate piecewise planned trajectories based on the piecewise reference paths $\Upsilon$. These planned trajectories are then joined to form the final parking trajectory.

	\section{EXPERIMENTAL RESULTS AND DISCUSSION}\label{EXPERIMENTAL}

	\newcommand{\imgscale}{0.128}
	\begin{figure*}[!t]
		\centering
		\subfloat[]{\includegraphics[scale=\imgscale]{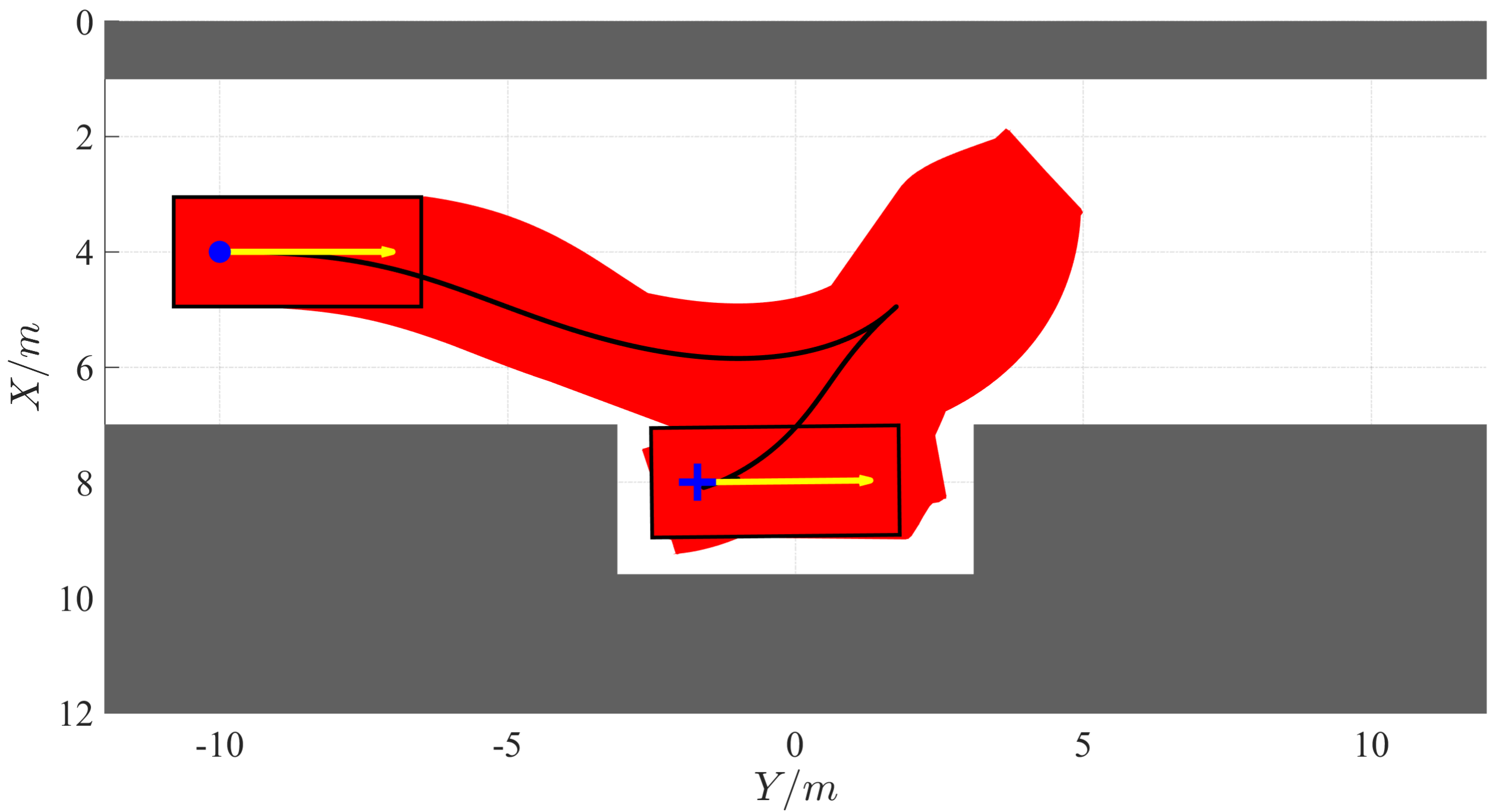}\label{fig:parallel_no_obs}}
		\hfil
		\subfloat[]{\includegraphics[scale=\imgscale]{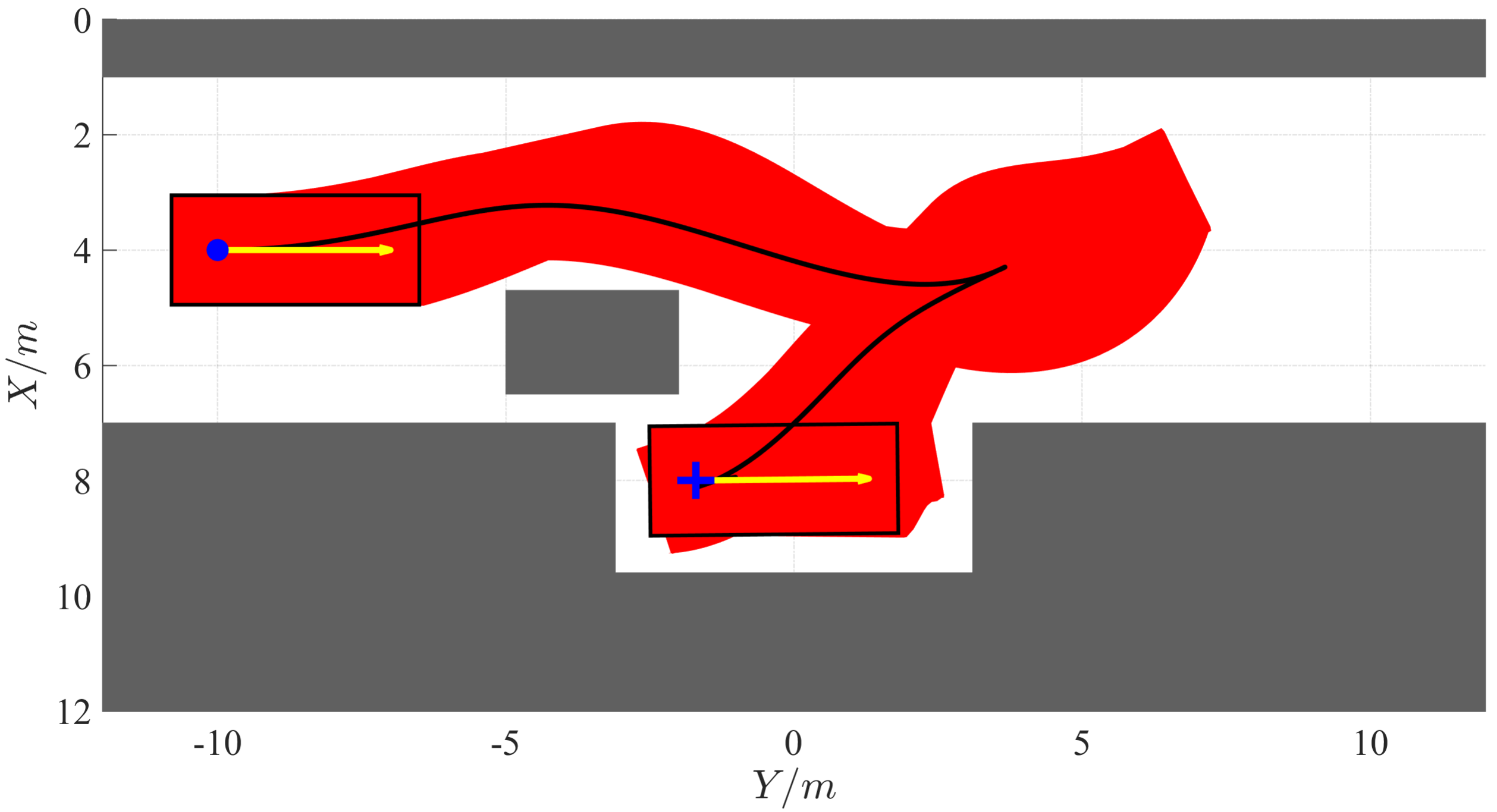}\label{fig:parallel_addobs_extra}}
		\hfil
		\subfloat[]{\includegraphics[scale=\imgscale]{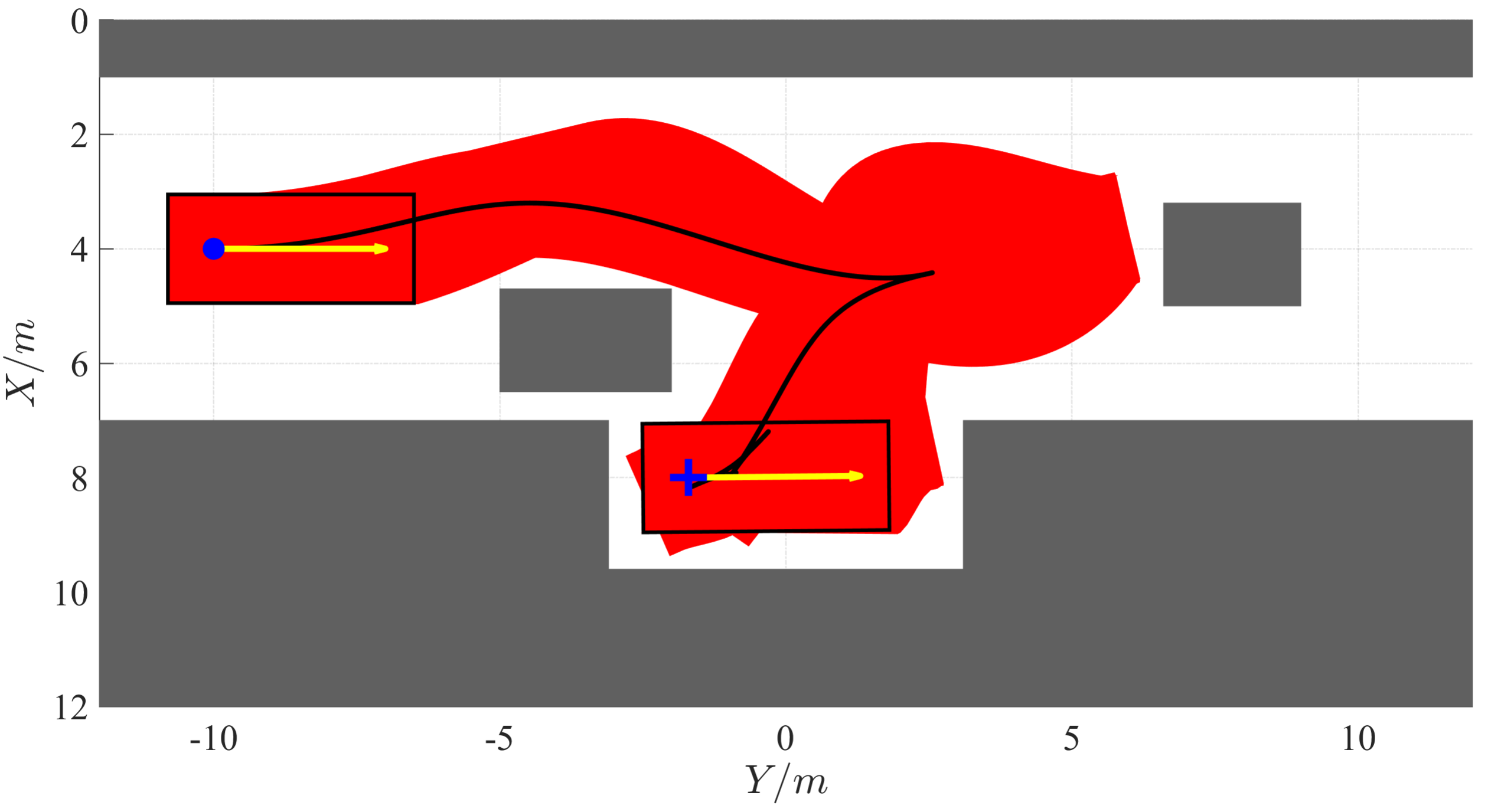}\label{fig:parallel_addobs}}
		\hfil
		\subfloat[]{\includegraphics[scale=\imgscale]{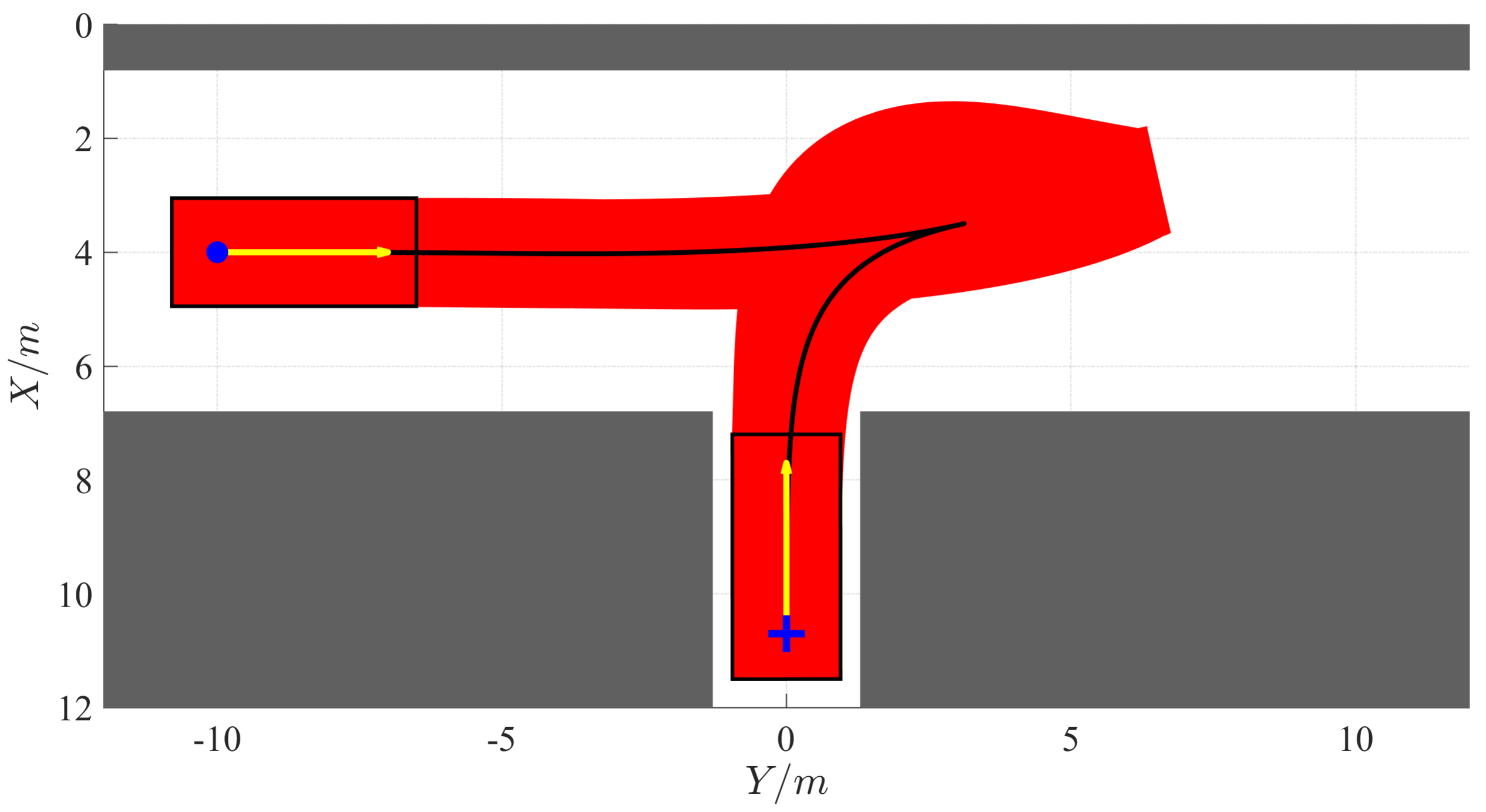}\label{fig:reverse_no_obs}}
		\hfil
		\subfloat[]{\includegraphics[scale=\imgscale]{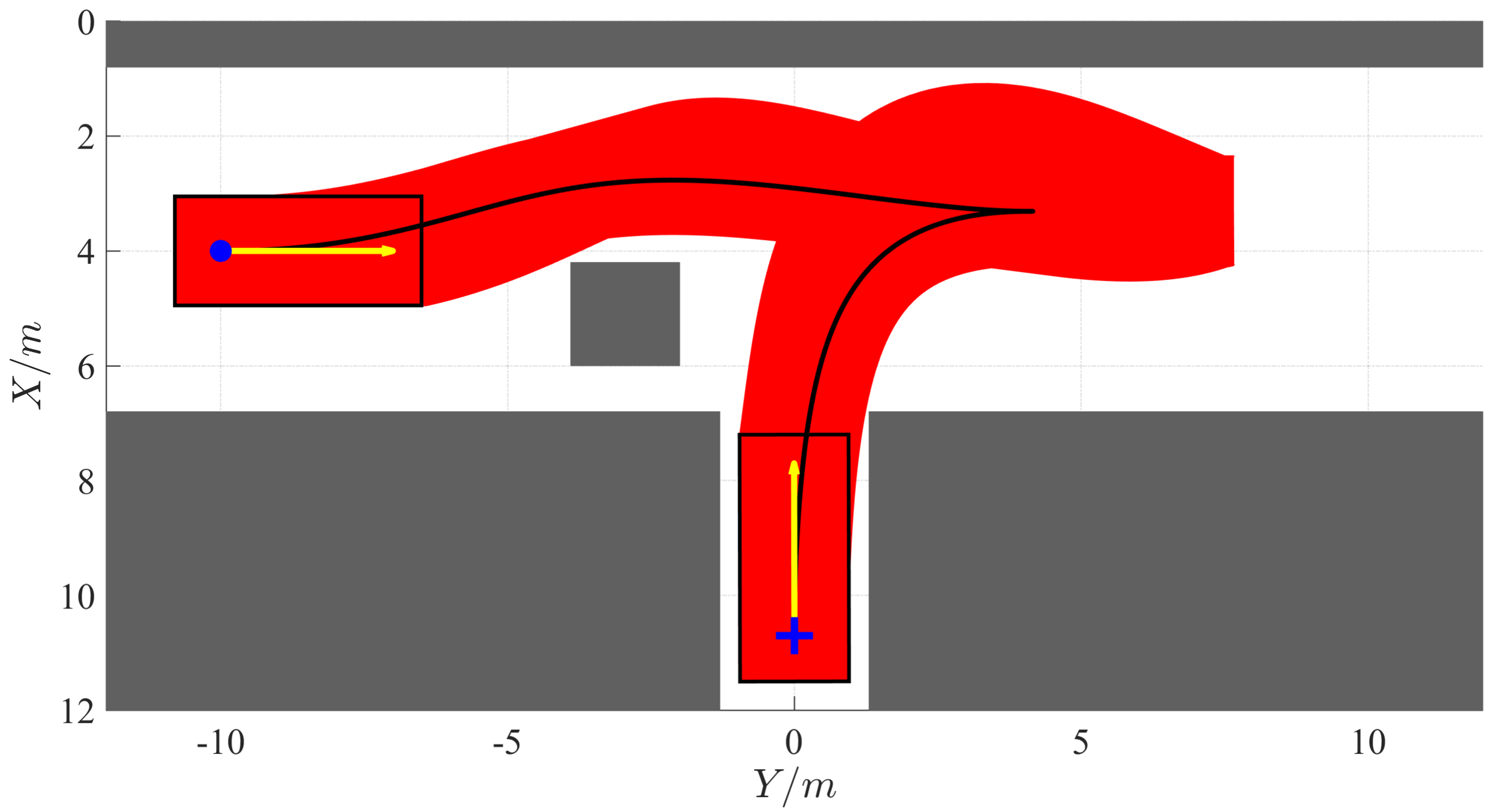}\label{fig:reverse_addobs_extra}}
		\hfil
		\subfloat[]{\includegraphics[scale=\imgscale]{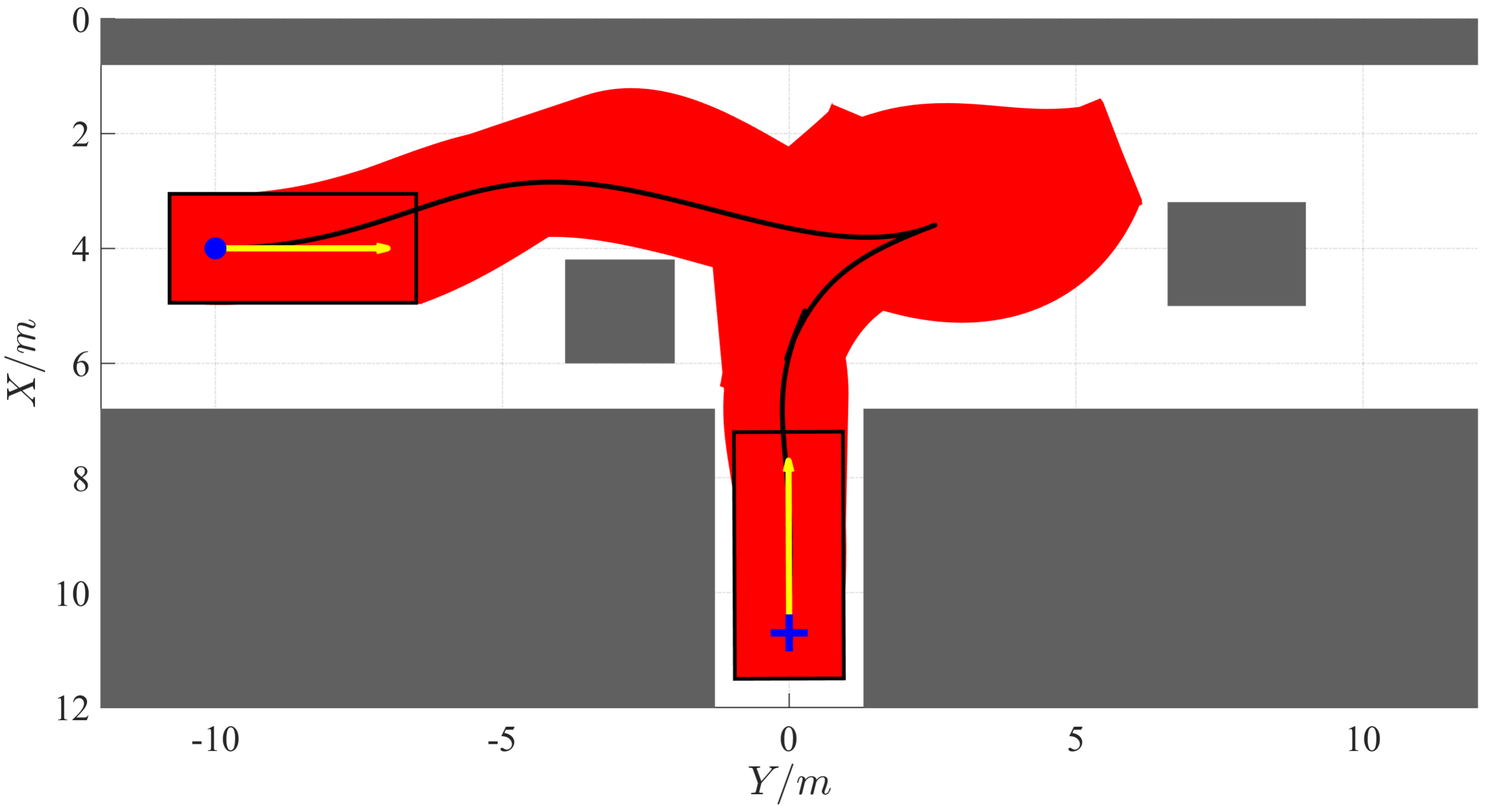}\label{fig:reverse_addobs}}
		\hfil
		\subfloat[]{\includegraphics[scale=\imgscale]{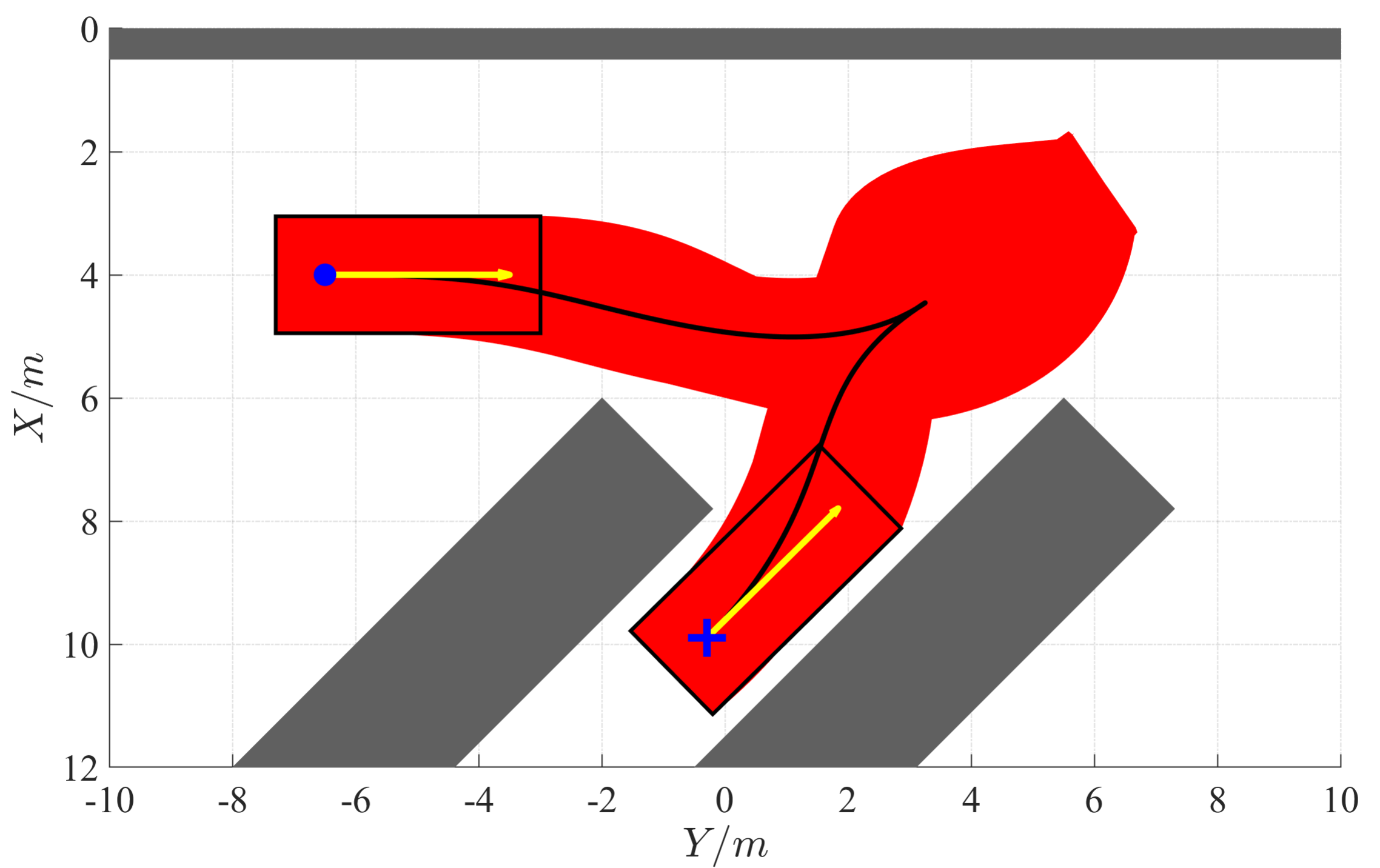}\label{fig:diagonal_no_obs}}
		\hfil
		\subfloat[]{\includegraphics[scale=\imgscale]{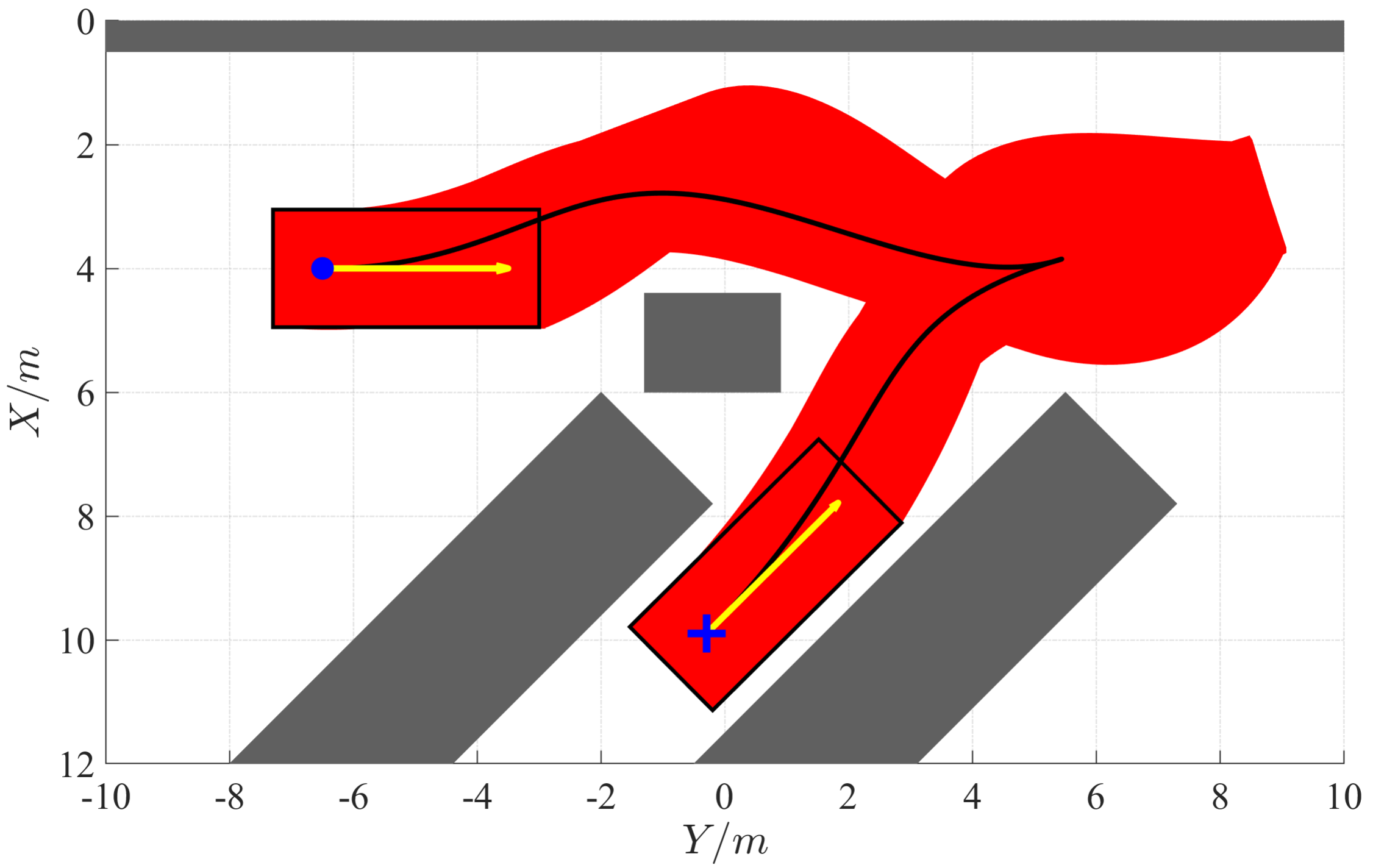}\label{fig:diagonal_addobs_extra}}
		\hfil
		\subfloat[]{\includegraphics[scale=\imgscale]{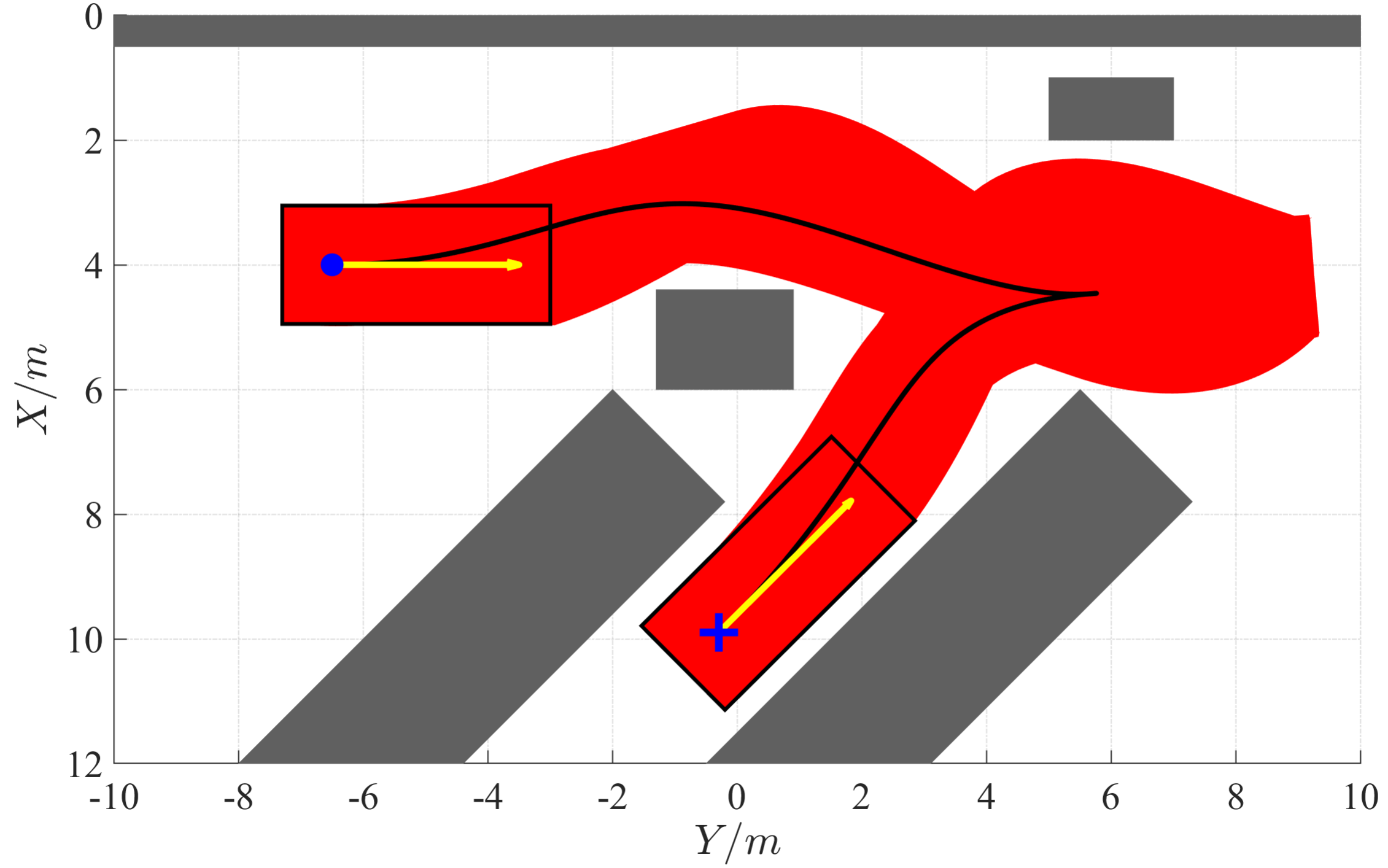}\label{fig:diagonal_addobs}}
		\caption{The first column of figures demonstrates the application of the RITP method to solve the trajectory planning problem in common parking scenarios. The figures in the second and third columns show the results of the RITP method after adding obstacles to the space traversed by the planned trajectories in the first and second columns of the scenario, respectively. \bluedot is the initial pose, and \myplus is the end pose. \myrect represents the space through which the planned trajectory travels.}
		\label{fig:ritp}
	\end{figure*}

	In this section, we first perform numerical simulations to demonstrate the effectiveness of the RITP method in various parking scenarios. We compare the RITP method with the state-of-the-art methods in terms of computation time and tracking error. Moreover, the effect of controller uncertainty on the control feasibility of the planned trajectory is analyzed. Secondly, we conduct real-world tests on parallel and reverse parking using the NanoCar platform based on the Ackerman steering architecture with ROS (Robot Operating System).
	
	\subsection{Simulation Setup}\label{Simulation_Setup}
	The numerical simulation software is MatlabR2023a. The CPU is 12th Gen Intel(R) Core(TM) i7-12700   2.10 GHz. The solver for the QP problem is quadprog. Parallel computing is performed using the Parallel Computing Toolbox in Matlab. 
	
	\begin{table}[!b]
		\centering
		\caption{Initial/End Poses and Vehicle Parameters of Trajectory Planning in Simulation Scenarios. }
		\begin{adjustbox}{max width=8cm, max height=4cm}
			\begin{tabular}{cl|cc}
				\toprule
				& \multicolumn{1}{c}{\textcolor[rgb]{ .137,  .122,  .125}{\textbf{Initial/End Pose}}} & \textcolor[rgb]{ .137,  .122,  .125}{\textbf{Parameters}} & \textcolor[rgb]{ .137,  .122,  .125}{\textbf{Setting}} \\
				\midrule
				\midrule
				\multirow{2}[1]{*}{\textcolor[rgb]{ .137,  .122,  .125}{\textbf{Parallel}}} & $[4.00, -10.00, \pi/2\,\textrm{rad}]$ & \textcolor[rgb]{ .137,  .122,  .125}{$v_\textrm{max}$}  & \textcolor[rgb]{ .137,  .122,  .125}{$2 \, $m/s} \\
				& $[8.00, -1.70,  \; \; \pi/2\,\textrm{rad}]$ & \textcolor[rgb]{ .137,  .122,  .125}{$a_\textrm{max}$}  & $\,1 \,$m/s$^2$ \\
				\multirow{2}[0]{*}{\textcolor[rgb]{ .137,  .122,  .125}{\textbf{Reverse}}} & $[4.00, -10.00, \pi/2\,\textrm{rad}]$ & $L$  & $2.8 \, $m \\
				& $[10.7, 0.00, \pi\,\textrm{rad}]$ & $W$  & $1.9 \, $m \\
				\multirow{2}[1]{*}{\textcolor[rgb]{ .137,  .122,  .125}{\textbf{Diagonal}}} & $[4.00, -6.50, \pi/2\,\textrm{rad}]$ & $a_f$  & $0.7 \, $m \\
				& $[9.90, -0.30, 0.75 \pi\,\textrm{rad}]$ & $a_r$  & $0.8 \, $m \\
				\bottomrule
			\end{tabular}%
		\end{adjustbox} 
		\label{tab:single_traj}%
	\end{table}%

	Referring to \cite{zhang2020optimization,zhang2021guaranteed}, we model the numerical simulation scenarios to the following dimensions:
	\begin{itemize}
		\item{Parallel parking: the length of the parking space is $6.2 \, $m, and the width is $2.6 \, $m.}
		\item{Reverse parking: the length of the parking space is $5.2 \, $m, and the width is $2.6 \, $m.}
		\item{Diagonal parking: the inclination angle of the parking space is $0.25 \, $rad, and the width of the parking space is about $2.7 \, $m.}
	\end{itemize}
	
	\begin{table}[!t]
		\centering
		\caption{The computation time of the ItCA method for different obstacles (in [s]), the corresponding length of the longest piecewise reference path (in [m]), and the number of sampling points associated with the path.}
		\begin{adjustbox}{max width=6cm, max height=10cm}
			\begin{tabular}{c|c|ccc}
				\toprule
				\multicolumn{1}{c}{\multirow{2}[4]{*}{\textbf{Scenario}}} & \multicolumn{1}{c}{\multirow{2}[4]{*}{\textbf{Metrics}}} & \multicolumn{3}{c}{\textbf{Extra Obstacle}} \\
				\cmidrule{3-5}    \multicolumn{1}{c}{} & \multicolumn{1}{c}{} & \boldmath{}\textbf{\textbf{$\textbf{0}$}}\unboldmath{} & \boldmath{}\textbf{\textbf{$\textbf{1}$}}\unboldmath{} & \boldmath{}\textbf{\textbf{$\textbf{2}$}}\unboldmath{} \\
				\midrule
				\midrule
				\multirow{3}[2]{*}{\textbf{$\textrm{Parallel}$}} & Time  & 0.0219  & 0.0232  & 0.0243  \\
				& $S_{i,\textrm{end}}$ & 12.299  & 14.099  & 12.899  \\
				& $M_{i,\mathbf{s}}$ & 246   & 282   & 258  \\
				\midrule
				\multirow{3}[2]{*}{\textbf{$\textrm{Reverse}$}} & Time  & 0.0198  & 0.0225  & 0.0232  \\
				& $S_{i,\textrm{end}}$ & 13.192  & 14.442  & 12.907  \\
				& $M_{i,\mathbf{s}}$ & 264   & 289   & 258  \\
				\midrule
				\multirow{3}[2]{*}{\textbf{$\textrm{Diagonal}$}} & Time  & 0.0215  & 0.0245  & 0.0247  \\
				& $S_{i,\textrm{end}}$ & 9.912  & 12.310  & 12.605  \\
				& $M_{i,\mathbf{s}}$ & 198   & 246   & 252  \\
				\bottomrule
			\end{tabular}%
		\end{adjustbox}
		\label{tab:iter_efficiency}%
		\begin{tablenotes}
			\footnotesize
			\item $M_{i,\mathbf{s}} = \lfloor  S_{i,\textrm{end}} \cdot 20 + 0.5 \rfloor$.
		\end{tablenotes}
	\end{table}%

	The width of the maneuvering area is $6 \, $m in all scenarios. We model the vehicle as a $ 4.3 \, \textrm{m}  \times 1.9 \, \textrm{m}$ rectangular. The vehicle kinematic parameters and vehicle geometric parameters are shown in Table~\ref{tab:single_traj}. The parameters of hybrid A* are set up reasonably to generate a global collision-free reference path.

	\subsection{On the Efficiency of Our RITP Method}\label{Efficiency}

	The superiority of the RITP method for ItCA and TSC will be verified in this subsection.
	
	\subsubsection{Effectiveness of the ItCA method}\label{collision-free_planning-based}

	We first demonstrate the effectiveness of the ItCA method in common parallel, reverse, and diagonal parking scenarios. Then, we investigate the influence of additional obstacles in the aforementioned common scenarios on collision-free performance and computation time. Furthermore, we verify the outstanding collision-free performance of the paths generated by ItCA in cluttered and complex environments. In addition, we intuitively show the effectiveness of the ItCA method by visualizing the values of the error vector $\mathbf{e}$ during the iterations.  The quadratic polynomial is chosen for path planning. The initial and end poses are shown in Table~\ref{tab:single_traj}.
	% Additionally, we confirm the effectiveness of the ItCA path planning. 
	
	To begin with, we validate the collision-free performance of the ItCA method in common parking scenarios. We validate parallel (Fig.~\ref{fig:parallel_no_obs}), reverse (Fig.~\ref{fig:reverse_no_obs}), and diagonal (Fig.~\ref{fig:diagonal_no_obs}) parking under open maneuvering space conditions. The ItCA method has been observed to effectively plan collision-free paths. Next, we add obstacles in the maneuvering space to simulate the possible occurrence of pedestrians, and the results of the path planning are shown in Fig.~\ref{fig:ritp}. The maneuvering space is fairly narrow, but the paths planned by the ItCA method effectively achieve collision avoidance. From this simulation, we conclude that the ItCA method can generate collision-free paths in common and rather narrow environments, given a suitable initialization path.
	
	\textcolor{black}{Next, we analyze the impact of adding additional obstacles on computation time.} The computation time of the ItCA process of Fig.\ref{fig:ritp} is shown in Table~\ref{tab:iter_efficiency}. The computational efficiency of the ItCA method is excellent, with or without the addition of obstacles. The path planning time can be maintained near \textbf{0.02}s. In all three scenarios, the difference in computation time between cases involving the addition of two obstacles and cases without obstacles is insignificant, indicating that obstacles do not significantly affect computation time. This is because the polynomial shape can be modified to achieve collision avoidance for multiple obstacles in a single iteration by adjusting the error vector $\mathbf{e}$. Additionally, it is worth noting that there exists a correlation between the computation time and the number of sampling points on the longest piecewise planned path. When there is sufficient maneuvering space in the parking lot, i.e., when there is no or only one obstacle, an increase in the number of sampling points results in a higher computation time, as the computational unit that requires the longest time plays a dominant role in determining the overall computational duration in parallel computing. However, its sensitivity to sampling points depends on the needs of the scenario, and more efficient path planning can be achieved in relatively open scenarios using fewer sampling points.
	
\begin{figure}[!t]
		\centering
		
		\subfloat[The cyan path indicates the planning result at the first iteration. The different colored curves represent the planned paths generated during iterations.]{\includegraphics[scale=0.18]{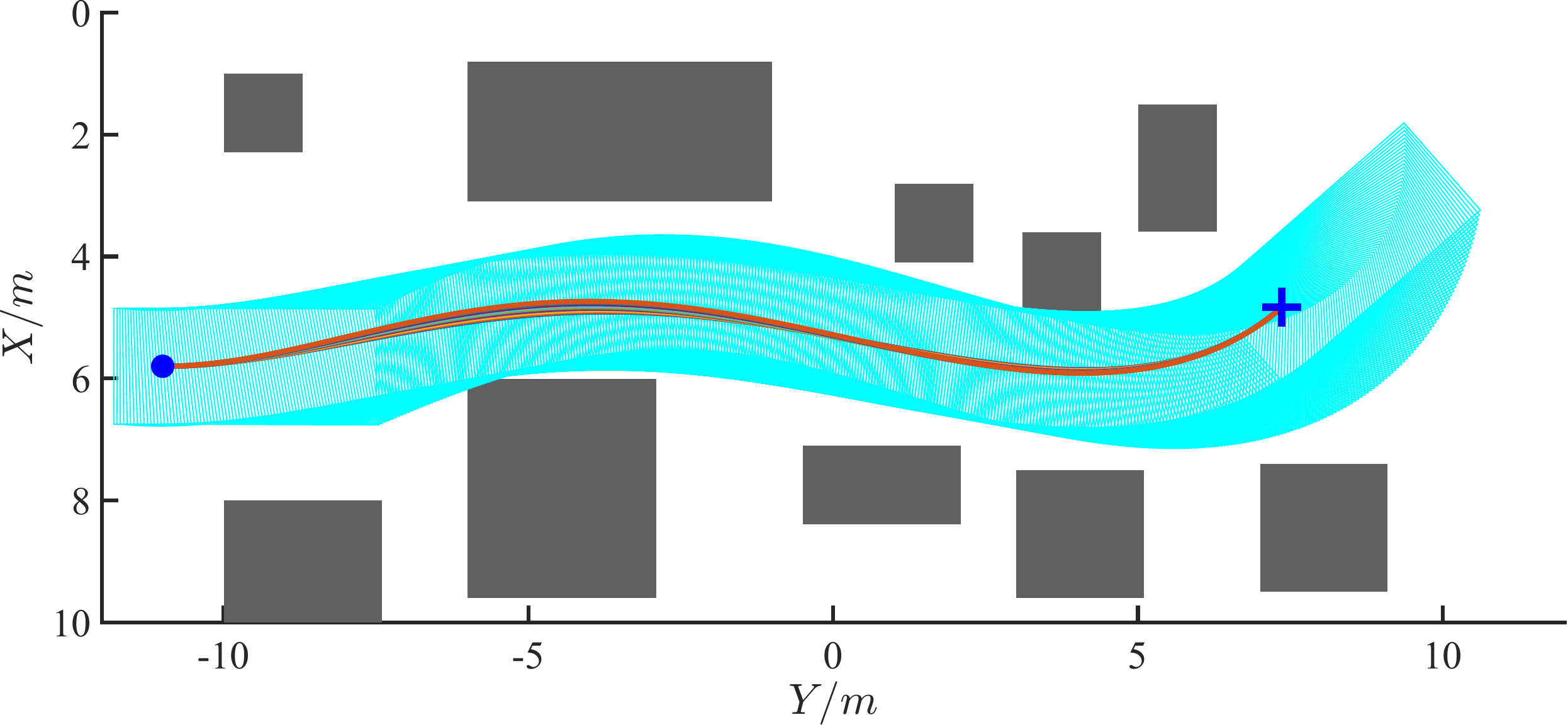}\label{fig:show_iter}}
		\hfil
		\subfloat[Details of the collision between the planned path and obstacles in the first iteration.]{\includegraphics[scale=0.24]{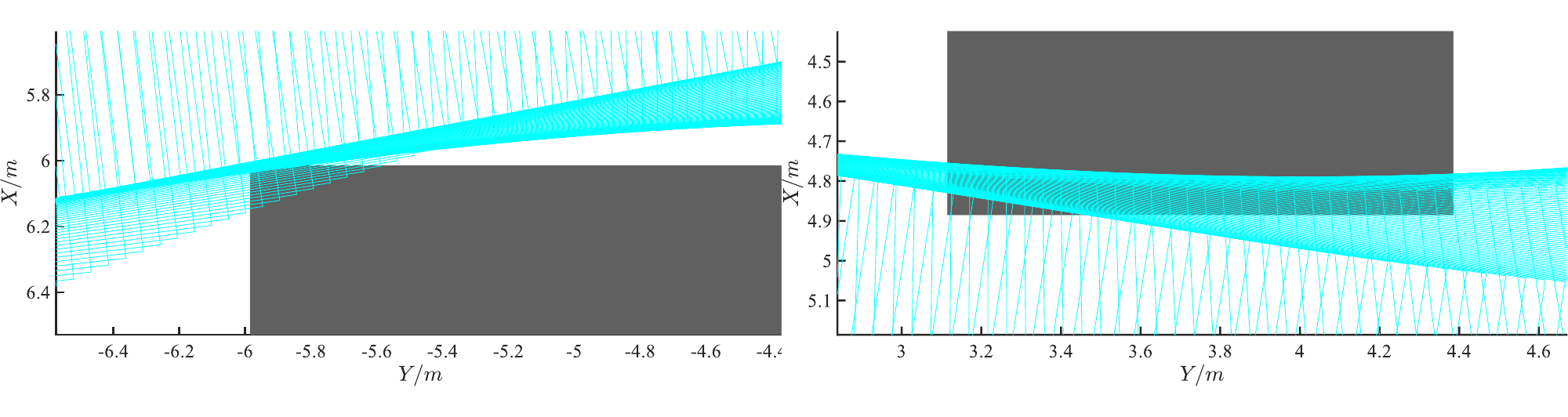}\label{fig:demon_iter}}
		\hfil
		\subfloat[Schematic diagram of the variation of the error vector $\mathbf{e}$ during ItCA process.]{\includegraphics[scale=0.18]{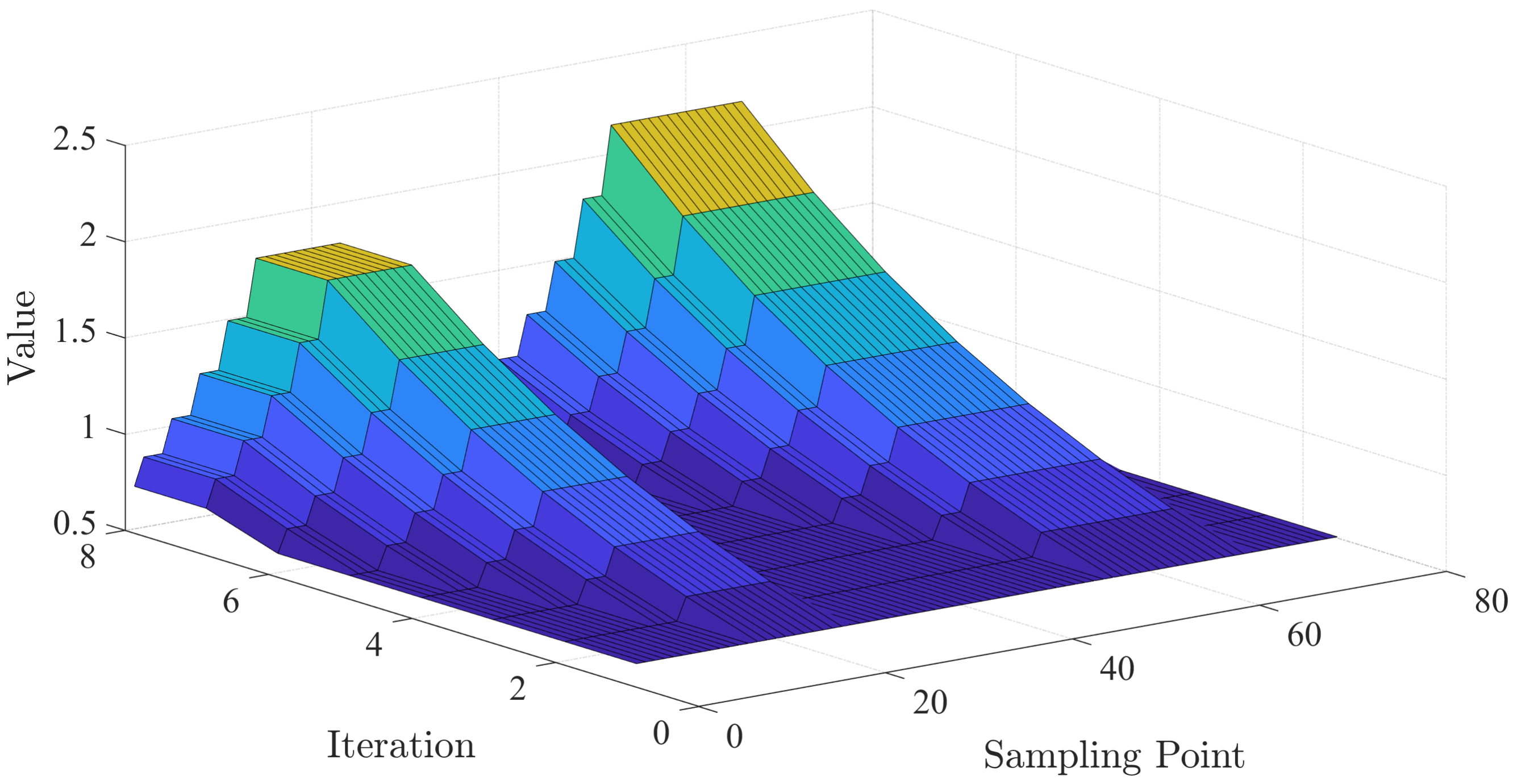}\label{fig:path_iter}}
		\hfil
		\caption{Schematic illustration of the effect of the ItCA method. In the first iteration, the planned path collides with obstacles (Fig.~\ref{fig:show_iter}), and after a finite number of iterations, ItCA path planning yields a collision-free path (Fig.~\ref{fig_clustter} and Fig.~\ref{fig:demon_iter_final}). The two peaks in Fig.~\ref{fig:path_iter} indicate the highest values of the error vector $\mathbf{e}$ corresponding to the two collision parts (Fig.~\ref{fig:demon_iter}) of the iterative path. The iteration coefficient $\beta$ is 1.2, and the number of iterations is 8. The initial values of error vectors $\mathbf{e}$ are all 0.6 to intuitively visualize the iterative process.}
		\label{fig:iter_total}
	\end{figure}

	Furthermore, we validate the effectiveness of the ItCA method through visualization, as depicted in Fig.~\ref{fig:iter_total} and Fig.~\ref{fig_narrow}. The results show that the first planned path collides with obstacles (see in Fig.~\ref{fig:iter_total}), and after ItCA, the planned path successfully achieves collision-free status (see in Fig.~\ref{fig_narrow}). This proves the effectiveness of the ItCA process. Meanwhile, we carry out the ItCA method in complex  (Fig.~\ref{fig_narrow1}) and cluttered (Fig.~\ref{fig_clustter}) scenarios, and the results are shown in Fig.~\ref{fig_narrow}. Even in cluttered environments, the ItCA method has demonstrated its capability to generate collision-free paths, fully showing its efficacy. In the complex scenario, the optimization-based collision avoidance (OBCA)\cite{zhang2020optimization} method also successfully plans a parking trajectory, but it ends up colliding with obstacles (see in  Fig.~\ref{fig_narrow1}). This is because the OBCA method does not consider collisions between sampling periods. On the other hand, the ItCA method utilizes dense sampling points to avoid collisions, which enables it to prevent potential collisions effectively. It is important to note that, as with other convex optimization-based trajectory planning methods, a collision-free reference path is necessary. There is a possibility that the RITP method fails to generate optimal solutions when the reference path is too close to obstacles. This case can be addressed by inflating obstacles when generating the reference path. This solution allows the reference path to maintain a safe distance from obstacles.

	\newcommand{\newimgs}{0.17}
	\newcommand{\imgs}{0.15}
	\begin{figure}[!t]
		\centering
		\subfloat[Comparison of collision-free performance of RITP (left) and OBCA (right) methods in the complex scenario.]{\includegraphics[scale=\imgs]{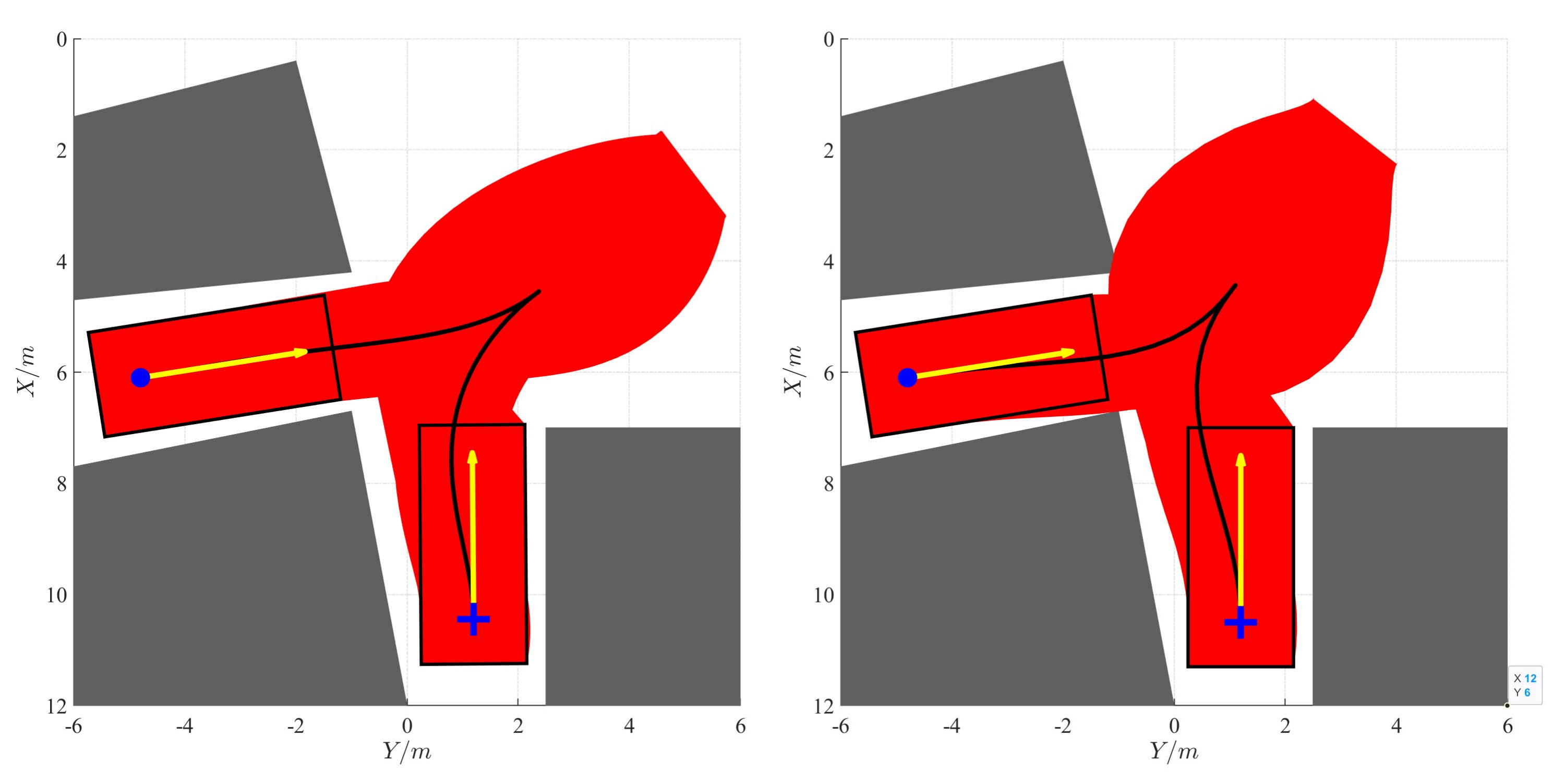}\label{fig_narrow1}}
		\hfil
		\subfloat[Path planned by the RITP method in the cluttered environment.]{\includegraphics[scale=\newimgs]{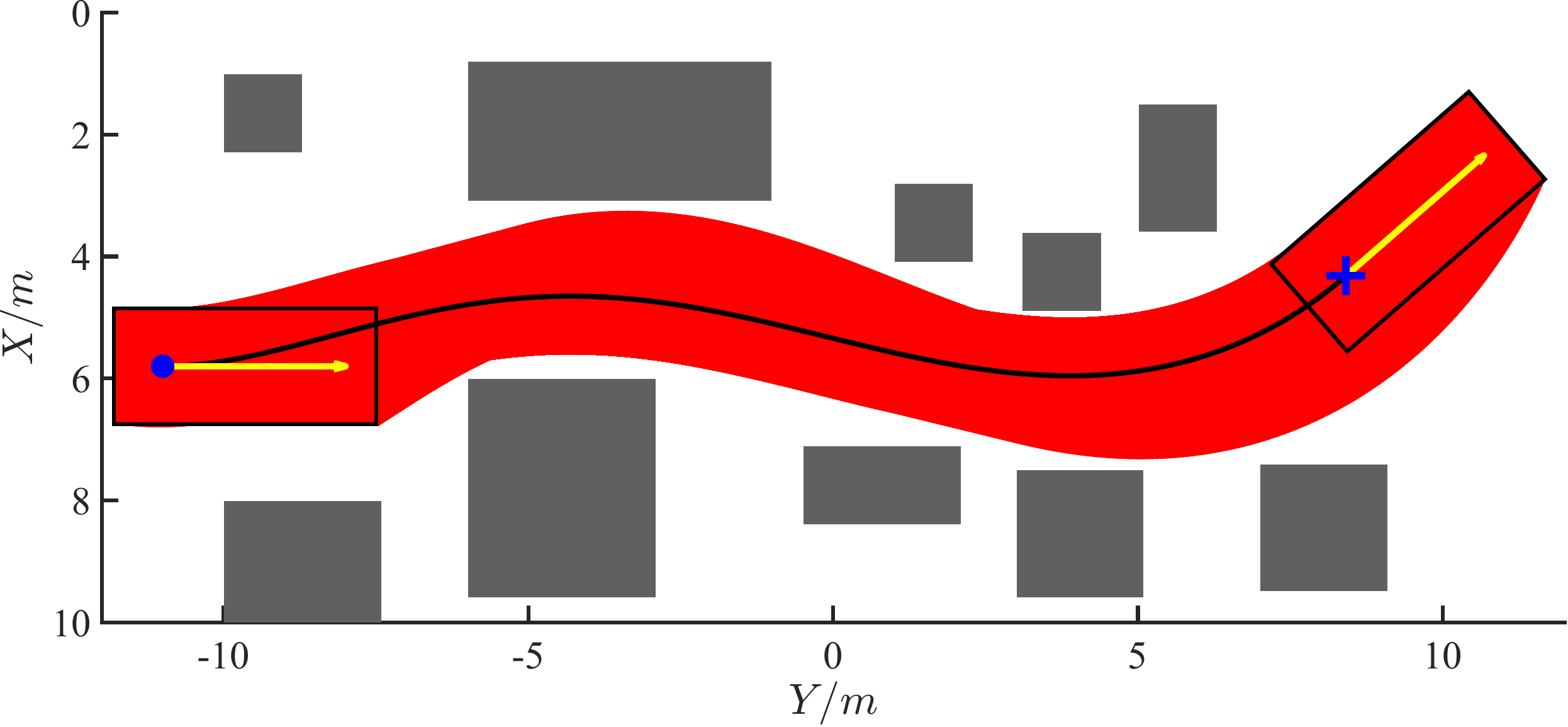}\label{fig_clustter}}
		\hfil
		\subfloat[A detailed diagram illustrates the collision-free path planned by the RITP method, depicting the position of the planned path in relation to obstacles.]{\includegraphics[scale=0.23]{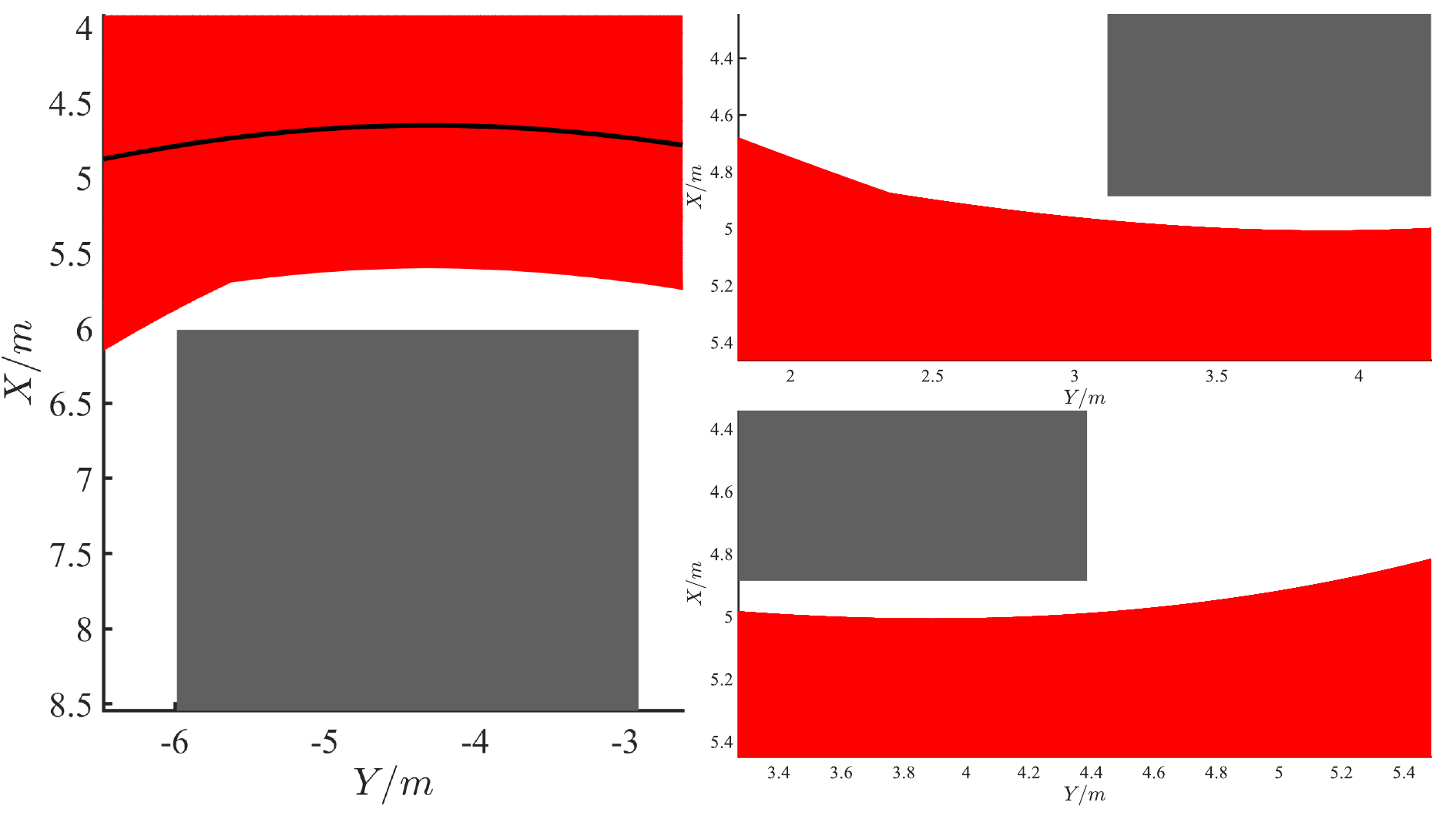}\label{fig:demon_iter_final}}
		\hfil
		\caption{The planning initial pose is $[6.10, -4.80, 0.55 \pi\,\textrm{rad}]$ and the end pose is $[10.50, 1.20, \pi\,\textrm{rad}]$ for the complex scenario Fig.~\ref{fig_narrow1}. \bluedot is the initial pose, and \myplus is the end pose. Cluttered environment Fig.~\ref{fig_clustter} is planned with the initial pose $[5.80, -11.00, 0.5\,\pi\,\textrm{rad}]$ and the end point pose $[4.60, 8.20, 0.75\,\pi\,\textrm{rad}]$.}
		\label{fig_narrow}
	\end{figure}

	To summarize, we present validation results on the collision-free performance and computational efficiency of the ItCA method in common scenarios and scenarios with additional obstacles. The analysis shows that the ItCA method guarantees excellent collision-free performance of the planned paths due to its advanced iterative process based on explicit collision detection, and the presence of additional obstacles in the scenarios does not significantly affect the computation time. Therefore, the ItCA method ensures excellent collision-free properties while maintaining real-time performance. 
	
	\begin{figure}[!t]
		\centering
		\includegraphics[scale=0.17]{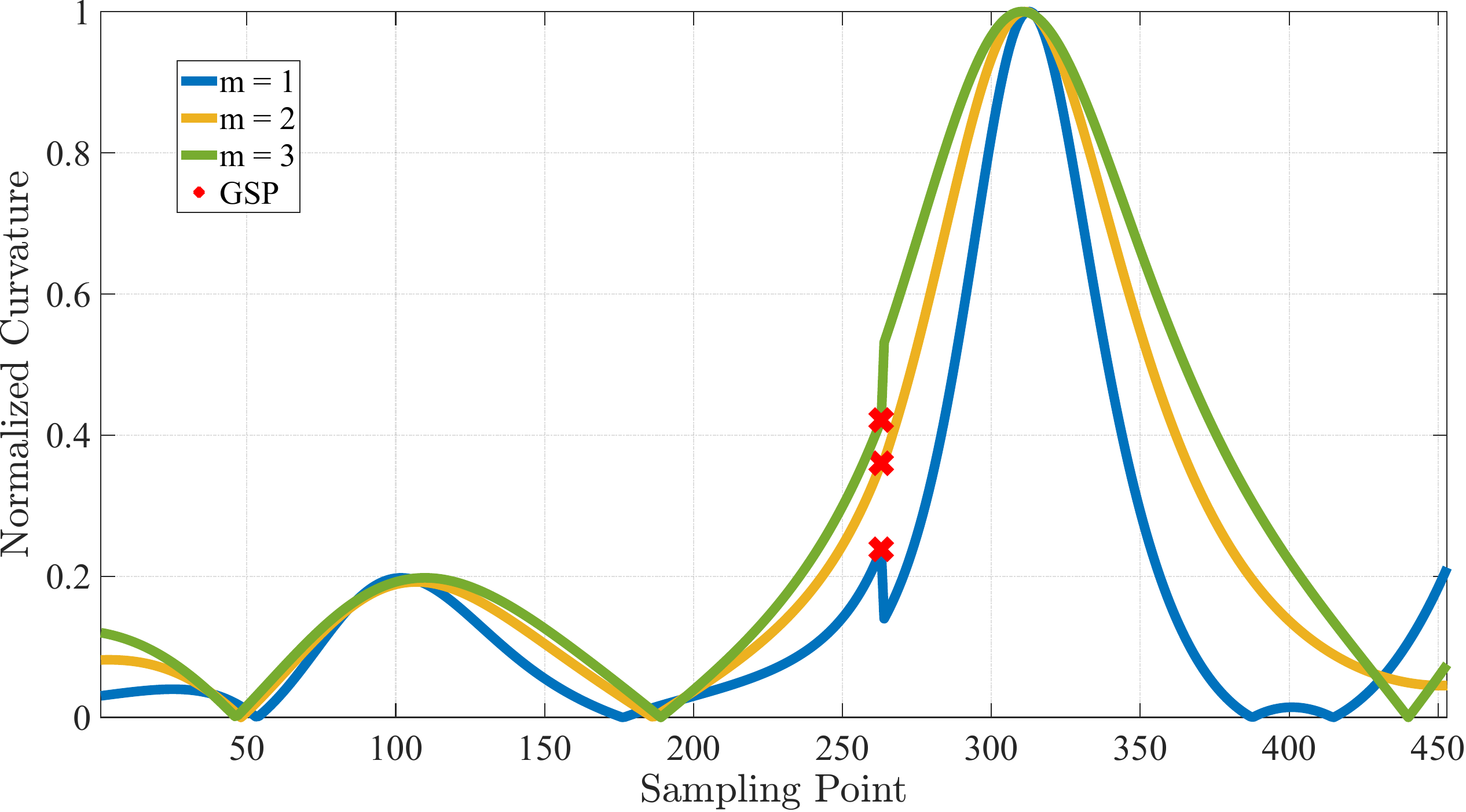}
		\caption{Normalized curvature of paths planned under different TSC at GSP. The initial pose is $[2.50, -9.10, 0.5\,\pi\,\textrm{rad}]$ and the end pose is $[10.70, 0, \pi\,\textrm{rad}]$. The task of automated parking is reverse parking. The quartic polynomial is chosen as the path carrier in the path-planning phase.}
		\label{fig:m_total}
	\end{figure}
	
	\begin{figure}[!t]
		\centering
		\subfloat[RITP trajectories with different polynomials and TSC at GSP.]{\includegraphics[scale=0.17]{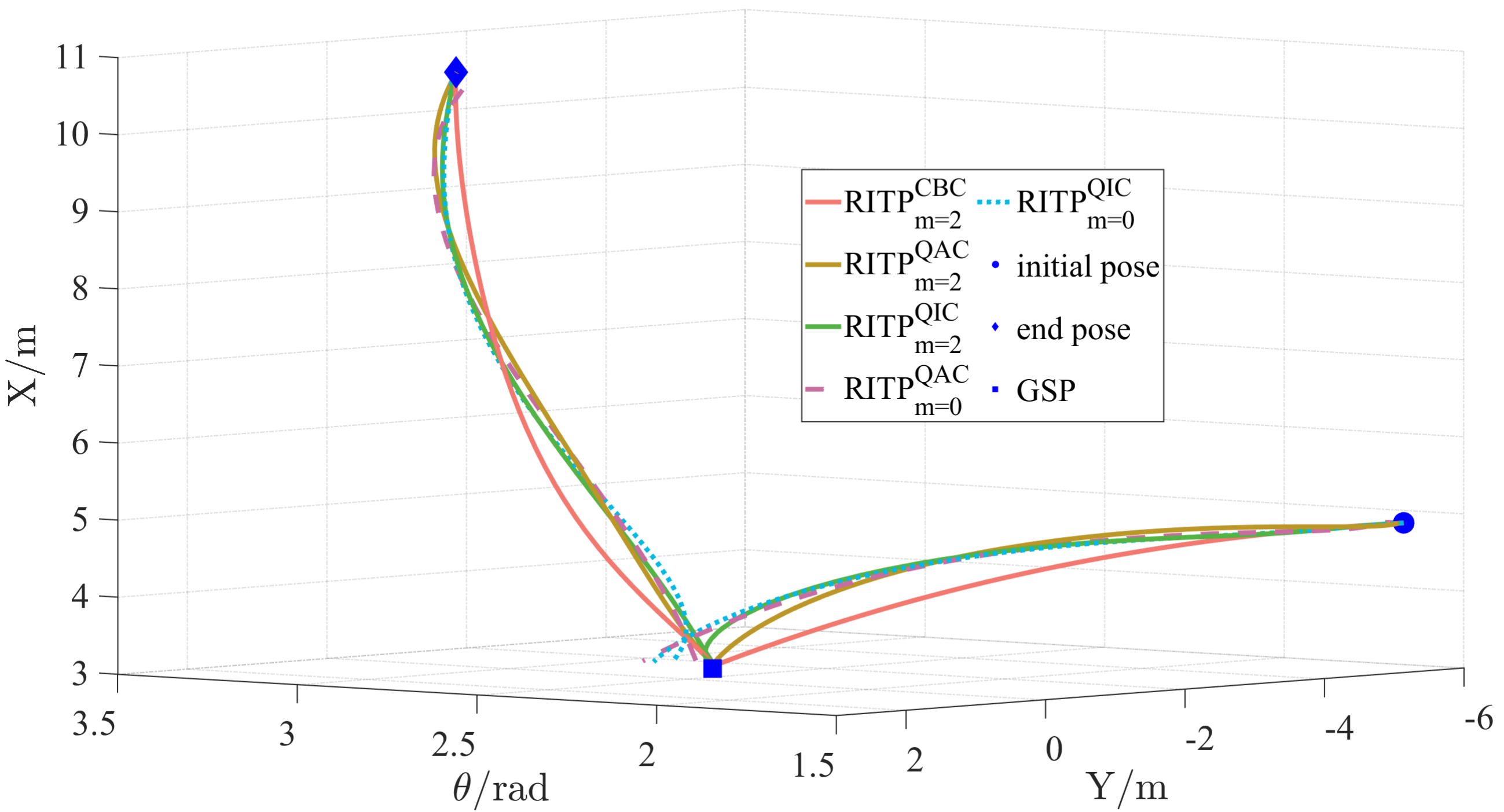}\label{fig:traj_tracked}}
		\hfil
		\subfloat[Boxplot of the tracking error for different planned trajectories.]{\includegraphics[scale=0.17]{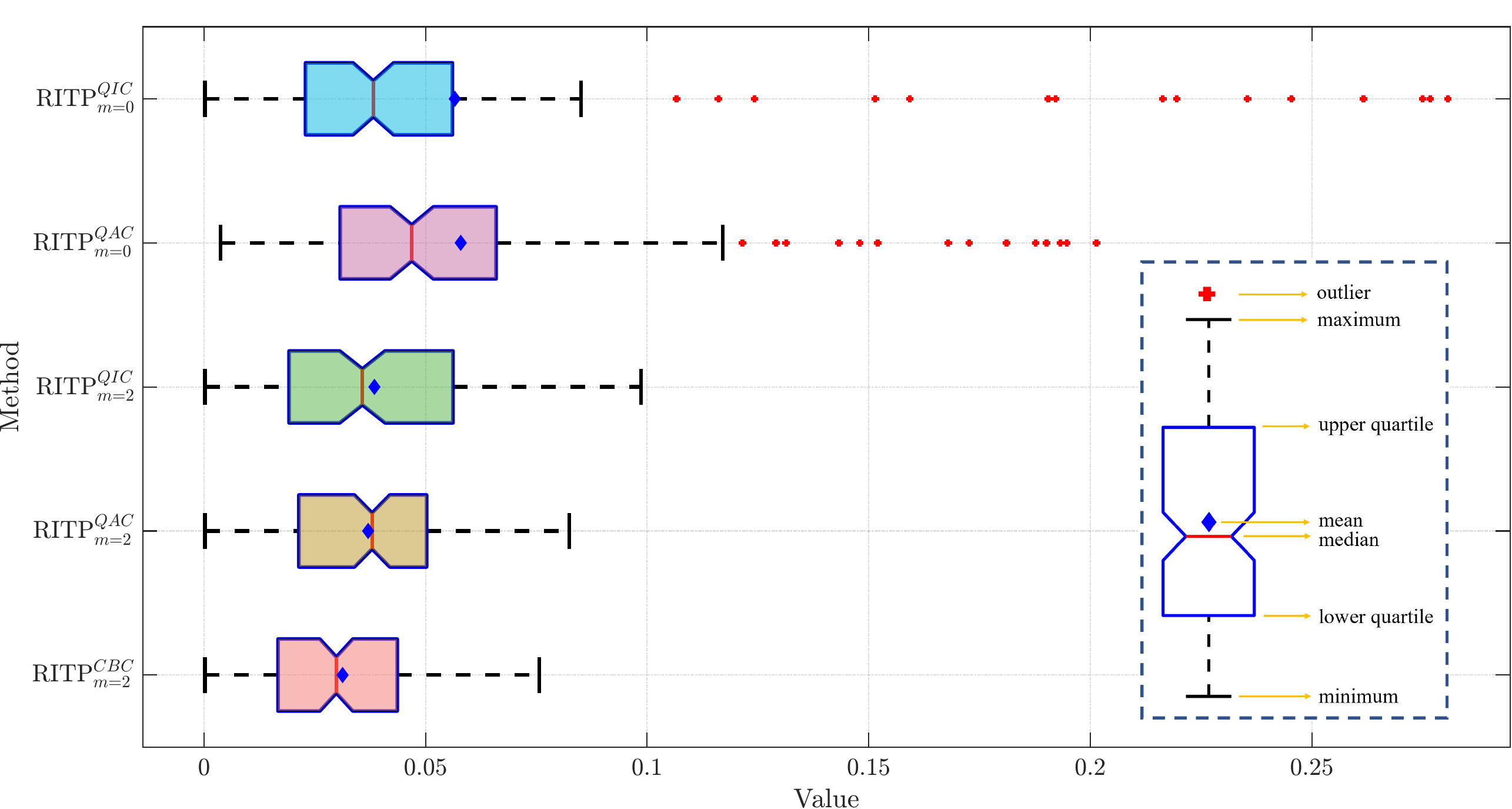}\label{fig:boxplot}}
		\hfil
		\caption{The initial pose is $[4.9, -5.50, 0.5\,\pi\,\textrm{rad}]$ and the end pose is $[10.70, 0, \pi\,\textrm{rad}]$. The task of automated parking is reverse parking. The yaw angle for the trajectories planned based on quartic and quintic polynomials is discontinuous at GSP when the smoothing order of TSC is 0.}
		\label{fig:total_poly}
	\end{figure}

	\subsubsection{Effectiveness of the TSC}\label{Terminal_Smoothing}

	\newcommand{\ColoredBoxGray}{\textcolor{gray}{\rule{0.7em}{0.5em}}}
	\begin{table*}[!t]
		\centering
		\caption{Evaluation Metrics and Experimental Results for Trajectory Planning in Common Parking Scenarios Demonstrate the Time Efficiency of the RITP Method and the Strong Control Feasibility of the Planned Trajectories.}
		\begin{adjustbox}{max width=16cm, max height=30cm}
			\begin{tabular}{c|p{5.2em}|c|c|ccccc|ccccc}
				\toprule
				\multicolumn{1}{c|}{\multirow{2}[2]{*}{\textbf{$\textrm{Scenario}^1$}}} & \multirow{2}[2]{*}{\textbf{Method}} & \multicolumn{1}{p{2.535em}|}{\textbf{$\textrm{Failure}^2$ }} & \multicolumn{1}{p{2.535em}|}{\textbf{$\textrm{Collision}^3$}} &       & \multicolumn{2}{p{5.07em}}{\textbf{\;\;\;\;\;\;\;\;\;\;\;\;$\textrm{M.CT}$}} &       &       &       & \multicolumn{2}{p{5.07em}}{\textbf{\;\;\;\;\;\;\;\;\;\;\;\;M.TE}} &       &  \\
				& \multicolumn{1}{c|}{} & \multicolumn{1}{p{2.535em}|}{\textbf{Rate}} & \multicolumn{1}{p{2.535em}|}{\textbf{ Rate}} & \multicolumn{1}{p{2.535em}}{\textbf{Max}} & \multicolumn{1}{p{2.535em}}{\textbf{Min}} & \multicolumn{1}{p{2.535em}}{\textbf{Mean}} & \multicolumn{1}{p{2.935em}}{\textbf{M.RP}} & \multicolumn{1}{p{2.535em}|}{\boldmath{}\textbf{$\sigma$}\unboldmath{}} & \multicolumn{1}{p{2.535em}}{\textbf{Max}} & \multicolumn{1}{p{2.535em}}{\textbf{Min}} & \multicolumn{1}{p{2.535em}}{\textbf{Mean}} & \multicolumn{1}{p{3.235em}}{\textbf{M.IP}} & \multicolumn{1}{p{2.535em}}{\boldmath{}\textbf{$\sigma$}\unboldmath{}} \\
				\midrule
				\midrule
				\multicolumn{1}{c|}{\multirow{8}[2]{*}{\textbf{Parallel}}} & OBCA  & 0.00\% & 88.75\% & 5.429  & 1.936  & 3.458  & \multicolumn{1}{p{2.935em}}{\;\;\;\;$\uparrow$} & 0.767  & 0.334  & 0.052  & 0.145  & 154.3\% & 0.062  \\
				& VPF   & 0.00\% & 62.50\% & 51.650  & 6.628  & 14.497  & \multicolumn{1}{p{2.935em}}{\;\;\;\;$\uparrow$} & 8.631  & 0.135  & \textbf{0.023 } & 0.062  & 8.7\% & 0.020  \\
				& CFS   & 1.25\% & 89.87\% & 0.584  & 0.049  & 0.205  & \multicolumn{1}{p{2.935em}}{\;\;\;\;$\uparrow$} & 0.144  & 56.285  & 0.672  & 7.430  & $\uparrow$ & 9.058  \\
				& $\textrm{RITP}^{CBC}_{m=2}$ & 8.75\% & 0.00\% & 0.109  & \textbf{0.022 } & \textbf{0.046 } & 24.5\% & \textbf{0.020 } & \textbf{0.080 } & 0.035  & 0.056  & -1.7\% & \textbf{0.009 } \\
				& $\textrm{RITP}^{QAC}_{m=2}$ & \textcolor[rgb]{ 1,  0,  0}{\textbf{0.00\%}} & \textcolor[rgb]{ 1,  0,  0}{\textbf{0.00\%}} & \cellcolor[rgb]{ .906,  .902,  .902} 0.110  & \cellcolor[rgb]{ .906,  .902,  .902} \textbf{0.022 } & \cellcolor[rgb]{ .906,  .902,  .902} 0.048  & \cellcolor[rgb]{ .906,  .902,  .902} 21.3\% & \cellcolor[rgb]{ .906,  .902,  .902} 0.021  & \cellcolor[rgb]{ .906,  .902,  .902} 0.085  & \cellcolor[rgb]{ .906,  .902,  .902} 0.043  & \cellcolor[rgb]{ .906,  .902,  .902} 0.058  & \cellcolor[rgb]{ .906,  .902,  .902} 1.7\% & \cellcolor[rgb]{ .906,  .902,  .902} \textbf{0.009 } \\
				& $\textrm{RITP}^{QAC}_{m=0}$ & 0.00\% & 1.25\% & \textbf{0.105 } & 0.024  & 0.049  & 19.6\% & \textbf{0.020 } & 0.123  & 0.039  & 0.060  & 5.2\% & 0.014  \\
				& $\textrm{RITP}^{QIC}_{m=2}$ & \textcolor[rgb]{ 1,  0,  0}{\textbf{0.00\%}} & \textcolor[rgb]{ 1,  0,  0}{\textbf{0.00\%}} & \cellcolor[rgb]{ .906,  .902,  .902} 0.437  & \cellcolor[rgb]{ .906,  .902,  .902} 0.030  & \cellcolor[rgb]{ .906,  .902,  .902} 0.061  & \cellcolor[rgb]{ .906,  .902,  .902} - & \cellcolor[rgb]{ .906,  .902,  .902} 0.051  & \cellcolor[rgb]{ .906,  .902,  .902} 0.090  & \cellcolor[rgb]{ .906,  .902,  .902} 0.036  & \cellcolor[rgb]{ .906,  .902,  .902} \textbf{0.057 } & \cellcolor[rgb]{ .906,  .902,  .902} - & \cellcolor[rgb]{ .906,  .902,  .902} \textbf{0.009 } \\
				& $\textrm{RITP}^{QIC}_{m=0}$ & 0.00\% & 5.00\% & 0.116  & 0.030  & 0.056  & 8.1\% & 0.021  & 0.729  & 0.036  & 0.071  & 24.5\% & 0.083  \\
				\midrule
				\multicolumn{1}{c|}{\multirow{8}[2]{*}{\textbf{Reverse}}} & OBCA  & 0.00\% & 71.25\% & 3.538  & 1.129  & 2.086  & $\uparrow$ & 0.486  & 0.297  & \textbf{0.023 } & 0.123  & 192.8\% & 0.050  \\
				& VPF   & 0.00\% & 83.75\% & 41.861  & 4.088  & 13.747  & $\uparrow$ & 6.713  & 0.238  & 0.028  & 0.063  & 50.0\% & 0.024  \\
				& CFS   & 0.00\% & 100.00\% & 0.432  & 0.031  & 0.136  & $\uparrow$ & 0.098  & 34.459  & 0.925  & 5.099  & $\uparrow$ & 5.132  \\
				& $\textrm{RITP}^{CBC}_{m=2}$ & 62.50\% & 0.00\% & \textbf{0.088 } & \textbf{0.025 } & \textbf{0.042 } & 35.3\% & \textbf{0.016 } & \textbf{0.071 } & 0.031  & 0.047  & 11.9\% & 0.012  \\
				& $\textrm{RITP}^{QAC}_{m=2}$ & \textcolor[rgb]{ 1,  0,  0}{\textbf{0.00\%}} & \textcolor[rgb]{ 1,  0,  0}{\textbf{0.00\%}} & \cellcolor[rgb]{ .906,  .902,  .902} 0.133  & \cellcolor[rgb]{ .906,  .902,  .902} 0.026  & \cellcolor[rgb]{ .906,  .902,  .902} 0.058  & \cellcolor[rgb]{ .906,  .902,  .902} 10.7\% & \cellcolor[rgb]{ .906,  .902,  .902} 0.023  & \cellcolor[rgb]{ .906,  .902,  .902} \textbf{0.071 } & \cellcolor[rgb]{ .906,  .902,  .902} 0.025  & \cellcolor[rgb]{ .906,  .902,  .902} \textbf{0.042 } & \cellcolor[rgb]{ .906,  .902,  .902} 0.0\% & \cellcolor[rgb]{ .906,  .902,  .902} 0.010  \\
				& $\textrm{RITP}^{QAC}_{m=0}$ & 0.00\% & 12.50\% & 0.133  & 0.027  & 0.060  & 7.6\% & 0.024  & 0.637  & 0.028  & 0.056  & 33.3\% & 0.067  \\
				& $\textrm{RITP}^{QIC}_{m=2}$ & \textcolor[rgb]{ 1,  0,  0}{\textbf{0.00\%}} & \textcolor[rgb]{ 1,  0,  0}{\textbf{0.00\%}} & \cellcolor[rgb]{ .906,  .902,  .902} 0.135  & \cellcolor[rgb]{ .906,  .902,  .902} 0.032  & \cellcolor[rgb]{ .906,  .902,  .902} 0.065  & \cellcolor[rgb]{ .906,  .902,  .902} - & \cellcolor[rgb]{ .906,  .902,  .902} 0.023  & \cellcolor[rgb]{ .906,  .902,  .902} \textbf{0.071 } & \cellcolor[rgb]{ .906,  .902,  .902} 0.029  & \cellcolor[rgb]{ .906,  .902,  .902} \textbf{0.042 } & \cellcolor[rgb]{ .906,  .902,  .902} - & \cellcolor[rgb]{ .906,  .902,  .902} \textbf{0.009 } \\
				& $\textrm{RITP}^{QIC}_{m=0}$ & 0.00\% & 10.00\% & 0.135  & 0.033  & 0.065  & 0.0\% & 0.023  & 0.123  & 0.031  & 0.050  & 19.0\% & 0.015  \\
				\midrule
				\multicolumn{1}{c|}{\multirow{8}[2]{*}{\textbf{Diagonal}}} & OBCA  & 0.00\% & 41.25\% & 5.009  & 1.550  & 3.279  & $\uparrow$ & 0.890  & 0.192  & \textbf{0.028 } & 0.090  & 66.6\% & 0.034  \\
				& VPF   & 0.00\% & 27.50\% & 56.715  & 5.360  & 14.279  & $\uparrow$ & 8.197  & 0.095  & 0.034  & 0.057  & 5.5\% & 0.013  \\
				& CFS   & 0.00\% & 100.00\% & 0.265  & 0.046  & 0.123  & $\uparrow$ & 0.052  & 39.511  & 0.927  & 5.527  & $\uparrow$ & 7.091  \\
				& $\textrm{RITP}^{CBC}_{m=2}$ & 3.75\% & 0.00\% & \textbf{0.079 } & \textbf{0.025 } & \textbf{0.045 } & 16.6\% & \textbf{0.011 } & 0.084  & 0.036  & \textbf{0.051 } & -5.5\% & 0.011  \\
				& $\textrm{RITP}^{QAC}_{m=2}$ & \textcolor[rgb]{ 1,  0,  0}{\textbf{0.00\%}} & \textcolor[rgb]{ 1,  0,  0}{\textbf{0.00\%}} & \cellcolor[rgb]{ .906,  .902,  .902} 0.083  & \cellcolor[rgb]{ .906,  .902,  .902} \textbf{0.025 } & \cellcolor[rgb]{ .906,  .902,  .902} 0.046  & \cellcolor[rgb]{ .906,  .902,  .902} 14.8\% & \cellcolor[rgb]{ .906,  .902,  .902} 0.012  & \cellcolor[rgb]{ .906,  .902,  .902} 0.079  & \cellcolor[rgb]{ .906,  .902,  .902} 0.035  & \cellcolor[rgb]{ .906,  .902,  .902} 0.052  & \cellcolor[rgb]{ .906,  .902,  .902} -3.7\% & \cellcolor[rgb]{ .906,  .902,  .902} 0.011  \\
				& $\textrm{RITP}^{QAC}_{m=0}$ & 0.00\% & 35.00\% & 0.083  & 0.027  & 0.050  & 7.4\% & 0.014  & 0.120  & 0.038  & 0.058  & 7.4\% & 0.016  \\
				& $\textrm{RITP}^{QIC}_{m=2}$ & \textcolor[rgb]{ 1,  0,  0}{\textbf{0.00\%}} & \textcolor[rgb]{ 1,  0,  0}{\textbf{0.00\%}} & \cellcolor[rgb]{ .906,  .902,  .902} 0.191  & \cellcolor[rgb]{ .906,  .902,  .902} 0.032  & \cellcolor[rgb]{ .906,  .902,  .902} 0.054  & \cellcolor[rgb]{ .906,  .902,  .902} - & \cellcolor[rgb]{ .906,  .902,  .902} 0.020  & \cellcolor[rgb]{ .906,  .902,  .902} \textbf{0.078 } & \cellcolor[rgb]{ .906,  .902,  .902} 0.039  & \cellcolor[rgb]{ .906,  .902,  .902} 0.054  & \cellcolor[rgb]{ .906,  .902,  .902} - & \cellcolor[rgb]{ .906,  .902,  .902} \textbf{0.010 } \\
				& $\textrm{RITP}^{QIC}_{m=0}$ & 0.00\% & 18.75\% & 0.089  & 0.034  & 0.053  & 1.8\% & 0.012  & 0.123  & 0.040  & 0.063  & 16.6\% & 0.016  \\
				\bottomrule
			\end{tabular}%
		\end{adjustbox} 
		\begin{tablenotes}
			\footnotesize
			\item 1 Corresponds to Fig.~\ref{fig_starts}.
			\item 2 
			Corresponds to the number of iterations of the RITP and CFS methods over 10.
			\item 3 Corresponds to the planned trajectory colliding with obstacles after being tracked by the controller.
			\item \textbf{Bold} indicates the best results under the same experimental conditions. \ColoredBoxGray\;represents experimental data for $\textrm{RITP}^{QAC}_{m=2}$ and $\textrm{RITP}^{QIC}_{m=2}$. $\uparrow$ indicates a significant increase, and - is the baseline.
		\end{tablenotes}
		\label{tab:time_control}%
	\end{table*}%
	
	In this subsection, we initially verify the curvature continuity of the planned paths with TSC of different orders at GSP. Subsequently, we analyze the impact of curvature continuity at GSP on the control feasibility of the planned trajectory. Next, we compare the trajectories planned by the RITP method using different polynomials as path carriers to assess the performance of various polynomials in trajectory planning. Finally, we track the planned trajectory using an existing controller and analyze the control feasibility of the planned trajectory based on the tracking error.
	
	At first, we use the quartic polynomial as the path carrier for the RITP method and apply TSC to different orders ($m=1,2,3$). Next, we calculate the normalized curvature of the planned path to verify curvature continuity. The result is shown in Fig.~\ref{fig:m_total}. It can be observed that at the TSC of order 2, the planned path can maintain optimal curvature continuity at GSP, thus ensuring excellent control feasibility. However, both relatively higher ($m=2$)  and relatively lower ($m=1$) smoothing orders affect the smoothness of the planned path. This is because the relatively lower smoothing order causes the planned path at GSP \textcolor{black}{to} be incompatible with the vehicle kinematics model. The relatively higher smoothing order causes the alignment point to be too distant from GSP, resulting in inaccurate yaw angle approximation. Therefore, it is crucial to choose the appropriate TSC to ensure the curvature continuity of the planned path.
	
	We further investigate the RITP method with and without TSC ($m=0$ and $m=2$) utilizing cubic (CBC), quartic (QAC), and quintic (QIC) polynomials. The planned trajectory is illustrated in Fig.\ref{fig:total_poly}. $\textrm{RITP}^{\textrm{CBC}}_{\textrm{m=2}}$ denotes the use of the CBC polynomial as the path carrier with the TSC order of 2. Similar meanings apply to other notations. In the case of $m=0$, the planned trajectories with QAC and QIC polynomials as path carriers exhibit different yaw angles at GSP (refer to Fig.\ref{fig:traj_tracked}), potentially compromising the control feasibility. However, when $m = 2$, the planned trajectories maintain excellent yaw angle continuity at GSP. This indicates that TSC ensures the continuity of the yaw angle in the state $\chi$ by guaranteeing the curvature continuity at GSP. Furthermore, it is noteworthy that there is an absence of planned trajectories based on CBC polynomials when $m=0$. This is because the CBC polynomial-based RITP method \textcolor{black}{lacks} the capability to iteratively generate collision-free paths within a limited number of iterations. This limitation arises from the restricted degrees of freedom of the CBC polynomial. In general, this comparison validates the significance of TSC in ensuring the kinematic feasibility of the planned trajectory at GSP.
	
	Next, we compare the effect of the RITP method for trajectory planning based on different polynomials. Fig.~\ref{fig:traj_tracked} illustrates that in the case of $m=2$, the planned trajectories generated with the QAC and QIC polynomials exhibit similarity. In contrast, the trajectory generated with the CBC polynomial significantly differs from the former two due to its lower degree of freedom. Also, although the CBC polynomial can meet the requirements of the vehicle kinematics model through differential flatness, their degrees of freedom are insufficient to generate collision-free paths in narrow parking spaces. In contrast, the QAC and QIC polynomials not only fulfill the requirements of the vehicle kinematics model but also can generate collision-free paths due to their high degrees of freedom. Therefore, utilizing the QAC and QIC polynomials for path planning when employing the RITP method for parking trajectory planning is advisable.

	To directly assess the control feasibility of the planned trajectories, we employ a highly efficient MPC-based automated parking trajectory controller, as introduced by Zhang et al.\cite{zhang2020trajectory}, to track the planned parking trajectories. The calculated tracking error results are depicted in Fig.~\ref{fig:boxplot}. The planned trajectories with CBC, QAC, and QIC polynomials as path carriers exhibit similar and excellent control feasibility at TSC of $m=2$. However, at $m=0$, the mean of the statistical values of the tracking errors for trajectories planned using the QAC and QIC polynomials closely aligns with the upper quartile (see $\textrm{RITP}^{\textrm{QAC}}_{\textrm{m=0}}$ and $\textrm{RITP}^{\textrm{QIC}}_{\textrm{m=0}}$ in Fig.~\ref{fig:boxplot}). Nevertheless, the minimum tracking error remains low. This suggests that the controller effectively tracks most planned trajectories, yet significant errors occur at GSP. After comparing the simulation results, it becomes evident that the TSC serves as an excellent approximation of the control feasibility at GSP.

	After conducting the simulation in this subsection, the following conclusions can be drawn: Firstly, the lack of TSC leads to significant deviations in the yaw angle of the planned trajectory at GSP, which affects the control feasibility and poses a potential risk to vehicle motion. However, implementing an appropriate TSC can significantly enhance the control feasibility of the planned trajectory, thereby improving overall vehicle safety. Secondly, trajectories planned with CBC, QAC, and QIC polynomials as path carriers exhibit similar control feasibility. However, CBC polynomial-based trajectory planning is unable to generate delicate parking trajectories due to its limited degrees of freedom.
	
	\subsection{On the Time Efficiency of Our RITP Method}\label{Time_Efficiency}
	
	\begin{figure}[!t]
		\centering
		\subfloat[Computation time of parallel parking.]{\includegraphics[scale=0.18]{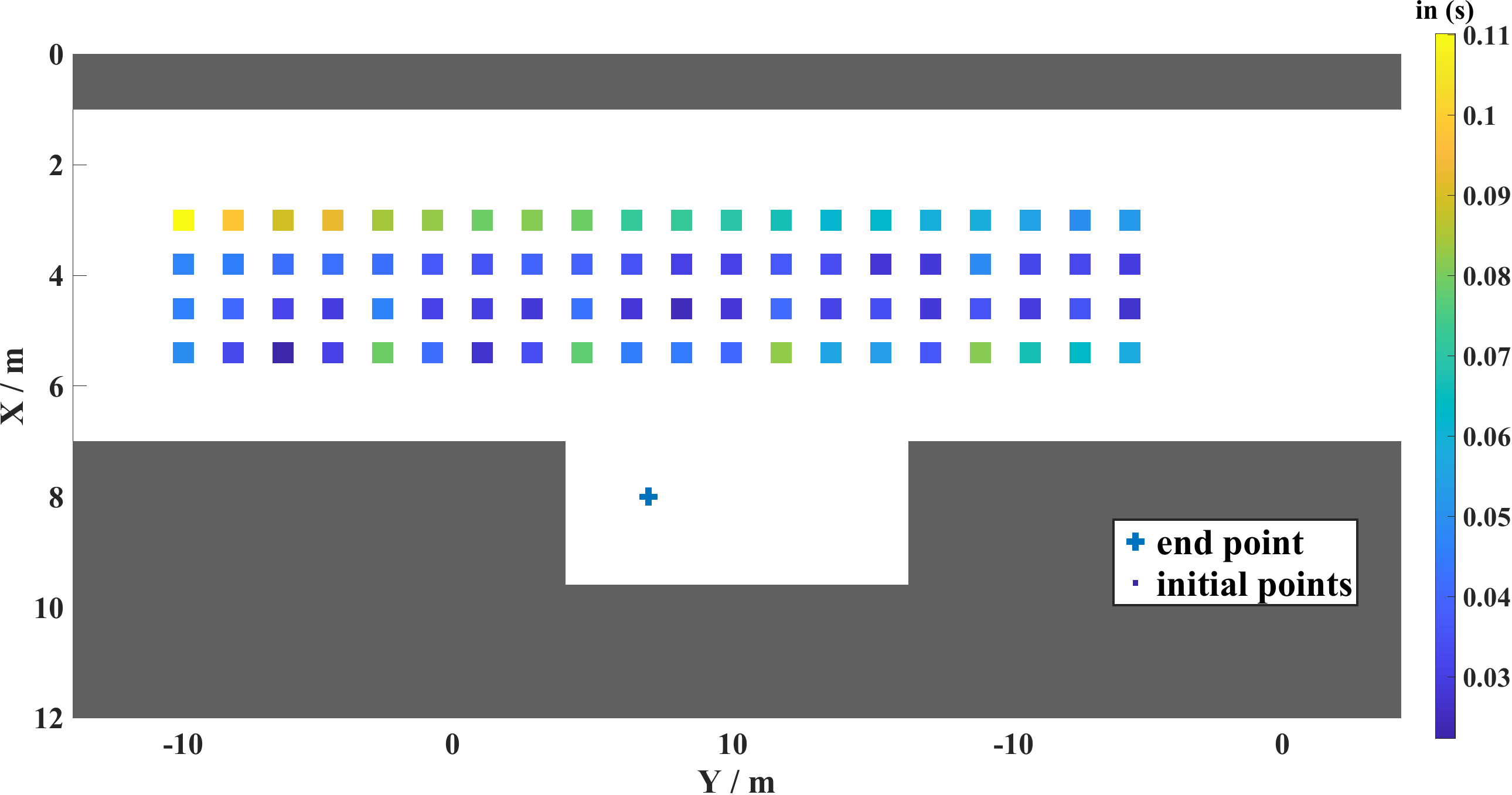}
			\label{fig_parallel_starts}}
		
		\hfil
		\subfloat[Computation time of reverse parking.]{\includegraphics[scale=0.18]{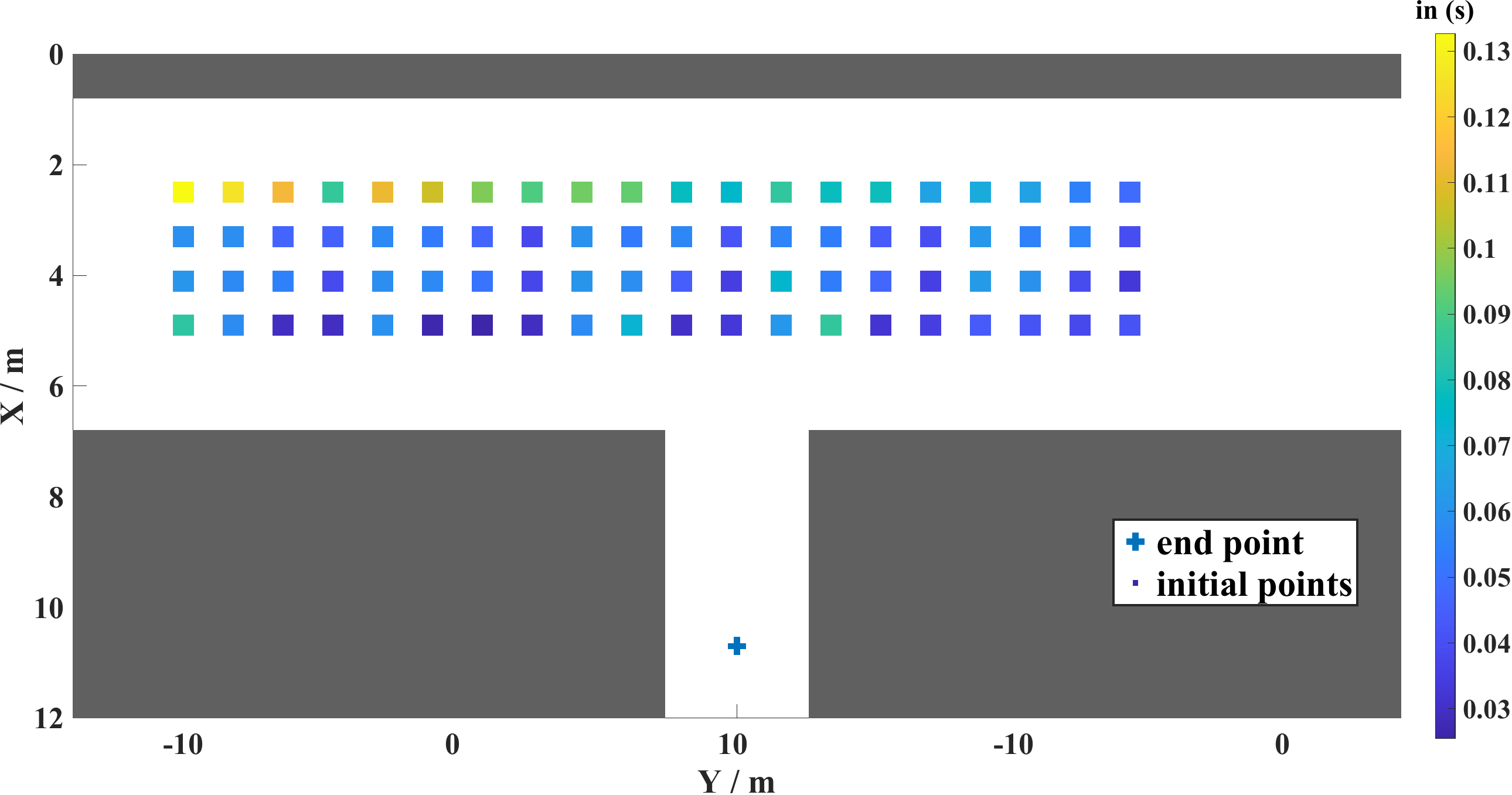}
			\label{fig_reverse_starts}}
		\hfil
		
		\subfloat[Computation time of diagonal parking.]{\includegraphics[scale=0.18]{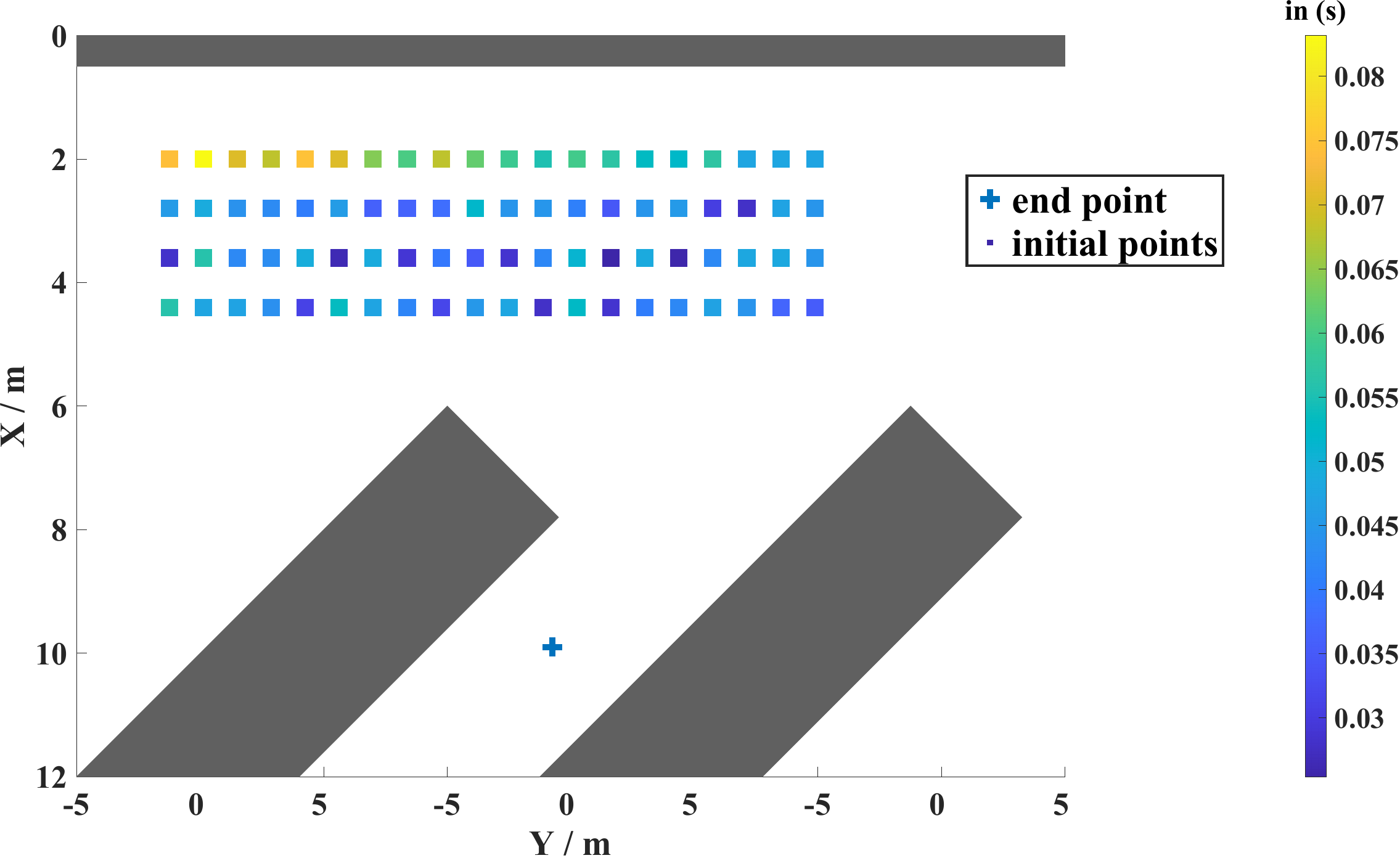}
			\label{fig_diagonal_starts}}
		\hfil
		
		\caption{Schematic diagram illustrating the distribution of initial, end poses, and computation time in various scenarios. The position represents the center of the vehicle's rear axle.}
		\label{fig_starts}
	\end{figure}
	
	To validate the time efficiency of the RITP method, we compared its computation time with that of the OBCA, VPF, and CFS methods. Additionally, we conducted ablation experiments for the RITP method to analyze the effects of different polynomials and TSC on the computation time of trajectory planning. In the comparison of ablation experiments, $\textrm{RITP}^{QIC}_{m=0}$ is chosen as the baseline (BL).
	
	We run ten tests with 80 different initial points within the maneuvering area, testing them in parallel, reverse, and diagonal parking scenarios (see Fig.~\ref{fig_starts}). We time each run and calculate the mean computation time from each initial pose to the end pose to determine the time efficiency.
	\newcommand{\ColoredBox}{\textcolor{blue}{\rule{0.5em}{0.5em}}}
	\newcommand{\CustomRing}[3]{%
		\raisebox{0.6ex}{%
			\tikz[baseline=(char.base)]{
				\node[draw, circle, minimum size=2*#3, line width=#2, text depth=-0.2ex, inner sep=0.5pt, color=#1] (char) {};
			}%
		}%
	}

	Referring to \cite{zhang2020optimization}, we model the parking candidate's initial points to the following settings:
	\begin{itemize}
		\item{For the parallel parking scenario, candidate initial points range: X direction distribution ${\begin{bmatrix} 3\,\textrm{m} & 5.4\,\textrm{m} \end{bmatrix}}$ with 0.8 m interval and Y direction distribution ${\begin{bmatrix} -10\,\textrm{m} & 7.1\,\textrm{m} \end{bmatrix}}$ with 0.9 m interval. The end pose is ${\begin{bmatrix} 8\,\textrm{m} & -1.6\,\textrm{m} & 0.5\,\pi \, \textrm{rad} \end{bmatrix}}^{ \intercal}$.}
		\item{For the reverse parking scenario, candidate initial points range: X direction distribution ${\begin{bmatrix} 2.5\,\textrm{m} & 4.9\,\textrm{m} \end{bmatrix}}$ with 0.8 m interval and Y direction distribution ${\begin{bmatrix} -10\,\textrm{m} & 7.1\,\textrm{m} \end{bmatrix}}$ with 0.9 m interval. The end pose is ${\begin{bmatrix} 10.7\,\textrm{m} & 0\,\textrm{m} & \pi \, \textrm{rad} \end{bmatrix}}^{ \intercal}$.}
		\item{For the diagonal parking scenario, candidate initial points range: X direction distribution ${\begin{bmatrix} 2\,\textrm{m} & 4.4\,\textrm{m} \end{bmatrix}}$ with 0.8 m interval and Y direction distribution ${\begin{bmatrix} -6.5\,\textrm{m} & 3.95\,\textrm{m} \end{bmatrix}}$ with 0.55 m interval. The end pose is ${\begin{bmatrix} 9.9\,\textrm{m} & -0.3\,\textrm{m} & 0.75 \pi \, \textrm{rad} \end{bmatrix}}^{ \intercal}$.}
	\end{itemize}

	The yaw angles of all initial poses are $0.5\,\pi$ rad. The schematic diagram of the initial and end poses are shown in Fig.~\ref{fig_starts}. The color of the \ColoredBox~represents the computation time of the RITP method. 
	
	As can be seen from Fig.~\ref{fig_starts}, the RITP method is very time efficient. The poses requiring more computation time for trajectory planning are concentrated in the first row near the left in all three scenarios. According to the results analyzed in Section  \ref{collision-free_planning-based}, since path planning based on these poses requires more sampling points for collision detection, it leads to additional computational resources. Consequently, this leads to a longer computation time for trajectory planning. Nevertheless, the RITP method remains very efficient, and the maximum time spent on trajectory planning is acceptable.
	
	The statistical data for the computation time of the ablation experiments in three common scenarios is presented in Table~\ref{tab:time_control}. The mean computation time (M.CT) from different initial poses to the end pose is used to evaluate the time efficiency of different methods. The mean reduction percentage (M.RP) is calculated as $\frac{M.CT_{BL}-M.CT_{Method}}{M.CT_{BL}}$.
	
	The M.CT for different methods in the comparison experiments is first analyzed. The time efficiency of the RITP method surpasses that of the OBCA and VPF methods with integrated models. In numerical simulations, the mean M.CT of the model-integrated OBCA and VPF methods ranges from 2.086s to 14.497s, and on the contrary, the mean M.CT of the RITP method ranges from 0.042s to 0.065s. This is because the RITP method simplifies path planning using differential flatness without including the vehicle model in the optimization problem. As a result, the path planning problem can be constructed as a QP problem, which can be solved quickly. In the three numerical simulation scenarios of autonomous parking, the trajectory planning time taken by the CFS method ranges from 0.123s to 0.205s. Compared to the CFS method, which also utilizes iterations for collision avoidance, the RITP method still demonstrates an advantage in time efficiency. This suggests that trajectory planning is more efficient when collision avoidance constraints are separated from the optimization problem and ensured in parallel iterations.

	The performance of the RITP method is then analyzed using statistical data from ablation experiments. The algorithm parameters are all set to be the same to ensure fairness. Analyzing the M.CT in the same TSC case, trajectory planning based on QAC and QIC polynomials requires a higher M.CT than that based on CBC polynomials. This is because the former two involve more optimization variables, thus requiring additional time to solve the optimization problem. However, trajectory planning based on CBC polynomials suffers failures in all three common parking scenarios. In contrast, trajectory planning based on QAC and QIC polynomials is able to guarantee a 0\% failure rate.  This is because of its inability to generate collision-free trajectories in tight areas due to the limited degrees of freedom. In terms of the standard deviation of M.CT, the dispersion of M.CT based on QAC and QIC polynomials for trajectory planning increases in parallel and diagonal parking scenarios. The minimum M.CT in these two scenarios is similar, while the maximum value shows significant differences. In association with Fig.~\ref{fig:ritp}, it can be observed that parallel and reverse parking usually involve long piecewise planned trajectories. As a result, the collision detection process is more time-consuming, leading to a higher maximum value of the M.CT, which increases the standard deviation of the statistical M.CT. However, this issue can be solved using fewer sampling points for collision detection in a relatively open space. Overall, using QAC and QIC polynomials for path planning ensures not only time efficiency but also a high trajectory planning success rate.

	In this subsection, we perform three categories of comparisons. First, we compare the RITP method with model-based trajectory planning methods, and the results highlight the time efficiency of the RITP method. Second, we compared it with the CFS method, which also utilizes iteration for collision avoidance. The results demonstrate that addressing collision avoidance through parallel iteration is more time-efficient than incorporating it directly into the optimization problem. Finally, we conducted ablation experiments using the RITP method. The results show that the TSC constraints do not impact the computation time, and the RITP method based on QIC and QAC polynomials can guarantee an excellent success rate and time efficiency.
	
	\subsection{On the Control Feasibility of Our RITP Method}\label{Energy_Efficiency}

	In this subsection, we compare the control feasibility of planned trajectories generated by the RITP method with those from the OBCA, VPF, and CFS methods. We use the controller employed in Section~\ref{Terminal_Smoothing} to track the trajectories planned by different methods. The mean tracking error (M.TE) for each trajectory is used to evaluate the control feasibility of that trajectory. As in Section~\ref{Time_Efficiency}, we conducted ablation experiments for the RITP method to analyze the effects of different polynomials and TSC on M.TE. In the comparison of ablation experiments, $\textrm{RITP}^{QIC}_{m=0}$ is chosen as the baseline (BL).

	The statistical values of the M.TE of the RITP, VPF, OBCA, and CFS methods in three common scenarios are shown in Table~\ref{tab:time_control}.  The mean improvement percentage (M.IP) is calculated as $\frac{M.TE_{Method}-M.TE_{BL}}{M.TE_{BL}}$.

	From the perspective of M.TE, the RITP method outperforms the VPF and OBCA methods with integrated models, which demonstrates the effectiveness of path planning using polynomials. This is ensured by the property of differential flatness. In contrast, the CFS method exhibits significant M.TE in all three scenarios, which is attributed to the fact that it does not consider either the vehicle model or the TSC in trajectory planning, which leads to the insufficient control feasibility of the planned trajectory.
	
	The ablation experiments conducted using the RITP method will be analyzed next. To ensure fairness, all algorithm parameters are set to the same to reflect fair competition. Firstly, trajectories planned by the RITP method achieve a 0\% collision rate (see $\textrm{RITP}^{QAC}_{m=2}$ and $\textrm{RITP}^{QIC}_{m=2}$) after being tracked by the controller when the TSC order is 2, demonstrating the strong control feasibility of the planned trajectory. Under the same TSC conditions, the trajectories planned based on QIC and QAC polynomials show similar control feasibility, suggesting that the QAC polynomial is sufficient for trajectory planning for automated parking. In the case of the same polynomials, using TSC significantly improves control feasibility. When using the QAC polynomials for path planning, the MT.E is reduced by 3.5\% using TSC over the MT.E without TSC in the parallel parking scenario, by 33.3\% using TSC over the MT.E without TSC in the reverse parking scenario, and by 11.1\% using TSC over the MT.E without TSC in the diagonal parking scenario.  The standard deviation analysis shows that the M.TE without TSC is generally more significant due to the discontinuity of the yaw angle on the planned trajectory near GSP, leading to less effective control feasibility. In general, the RITP method using QAC and QIC polynomials not only ensures the success rate of trajectory planning but also ensures the strong control feasibility of the planned trajectory. Moreover, the application of TSC has an excellent effect on the control feasibility and stability of the planned trajectory.

	Next, the difficulty of parking in different scenarios is analyzed. According to the control feasibility analysis, the parallel parking scenario has the highest M.TE, followed by the diagonal parking scenario, while the reverse parking has the lowest M.TE. This indicates that parallel parking is the most challenging, followed by diagonal parking, while reverse parking is a relatively easy scenario. This is consistent with the findings of the previous study\cite{zhang2020trajectory}.
	
	In this subsection, we verify the strong control feasibility of the trajectories planned by the RITP method. We confirm that although the CBC polynomial satisfies the requirements of the vehicle kinematic model through differential flatness, higher-order polynomials are more suitable for automated parking trajectory planning due to their increased degrees of freedom. We compare the QAC polynomial to the commonly used QIC polynomial and find that the QAC polynomial achieves rapid trajectory planning while ensuring similar control feasibility. Furthermore, we verify that the TSC significantly improves the control feasibility of the planned trajectory at GSP in common parking scenarios. This forms the foundation for the capability of the RITP method to perform parallel trajectory planning.

	\begin{figure}[!t]
		\centering
		\includegraphics[scale=0.18]{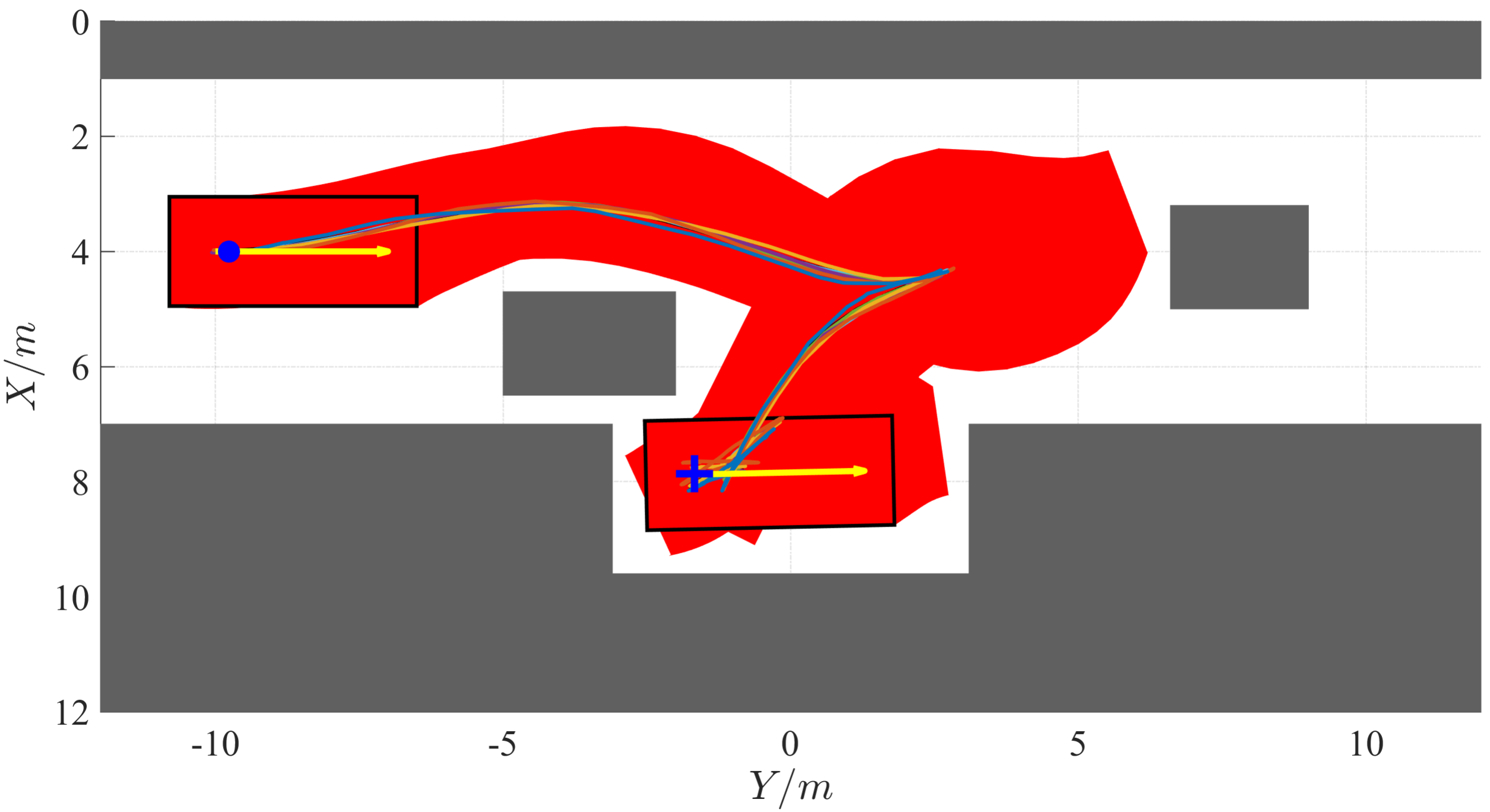}
		\caption{The tracked effect of the trajectories planned by the RITP method under the influence of the controller suffering the model uncertainty. \myrect denotes the space traversed by the tracked trajectories of the control inputs without disturbances. Different colored curves indicate the tracked trajectories when the control inputs suffer from different degrees of disturbance.}
		\label{fig:uncertain}
	\end{figure}

	\begin{table}[!t]
		\centering
		\caption{Tracking Error of the trajectory planned by the RITP method in parallel parking scenarios with additional obstacles.}
		\begin{adjustbox}{max width=8cm, max height=30cm}
			\begin{tabular}{cccccc}
				\toprule
				Level & Uncertainty & Max   & Mean  & TE.IP  & $\sigma$  \\
				\midrule
				\midrule
				BL    & 0     & 0.0396  & 0.0110  & -     & 0.0072  \\
				\midrule
				\multirow{2}[2]{*}{Low} & $[0.2 \textrm{m/s},\,0.05 \textrm{rad}]$ & 0.0429  & 0.0114  & 3.64\% & 0.0072  \\
				& $[0.3 \textrm{m/s},\, 0.08 \textrm{rad}]$ & 0.0526  & 0.0115  & 4.55\% & 0.0083  \\
				\midrule
				\multirow{2}[2]{*}{Medium} & $[0.4 \textrm{m/s},\, 0.12 \textrm{rad}]$ & 0.0547  & 0.0118  & 7.27\% & 0.0085  \\
				& $[0.6 \textrm{m/s},\, 0.17 \textrm{rad}]$ & 0.0632  & 0.0129  & 17.27\% & 0.0096  \\
				\midrule
				\multirow{2}[2]{*}{High} & $[0.8 \textrm{m/s},\, 0.26 \textrm{rad}]$ & 0.0840  & 0.0171  & 55.45\% & 0.0151  \\
				& $[1.0 \textrm{m/s},\, 0.34 \textrm{rad}]$ & 0.1034  & 0.0186  & 69.09\% & 0.0182  \\
				\bottomrule
			\end{tabular}%
		\end{adjustbox} 
		\label{tab:Uncertainty}%
	\end{table}%
	
	\subsection{Model Uncertainty}\label{Model_Uncertainty}

	In this subsection, the performance of the planned trajectory of the RITP method is verified when it is tracked by a disturbed controller, utilizing M.TE as the analytical metric. Because the parallel parking scenario is the most challenging, the planned trajectory for this scenario is chosen for analysis. The uniformly distributed noise is added to the localization data of the vehicle, and the disturbance amplitude is 0.01 m. Then, the different levels of uniformly distributed disturbance are added to control inputs, and the disturbance amplitude and the results are shown in Table~\ref{tab:Uncertainty}. The planned trajectories are shown in Fig.~\ref{fig:uncertain}. The tracking error improvement percentage (TE.IP) is calculated as $\frac{TE_{BL}-TE_{noise}}{TE_{BL}}$.
	
	As observed in Fig.~\ref{fig:uncertain}, in the absence of disturbance in control inputs, the planned trajectory ensures collision-free characteristics after being tracked by the controller. From Table~\ref{tab:Uncertainty}, it is evident that with relatively minor control input disturbances, the TE.IP of the planned trajectory can be maintained within 5\% compared to the case without disturbance. This demonstrates the strong control feasibility of the planned trajectories of the RITP method. Moreover, the standard deviation $\sigma$ can be maintained sufficiently, indicating stability during trajectory tracking. This is because the RITP method not only ensures the vehicle kinematics feasibility and applies the TSC in the path planning but also sets the 0 acceleration of the trajectory terminal in the velocity planning phase, which ensures robust control feasibility of the planned trajectory. When the disturbance of control inputs increases to a high value, it will produce a relatively large TE, and the standard deviation $\sigma$ of the TE will be relatively high. Naturally, considering the controller uncertainty in the trajectory planning phase can improve the control feasibility of the planned trajectory.
	
	\subsection{Experimental Setups and Results}\label{Experimental_Setups}

	\begin{figure}[!b]
		\centering
		\includegraphics[scale=0.23]{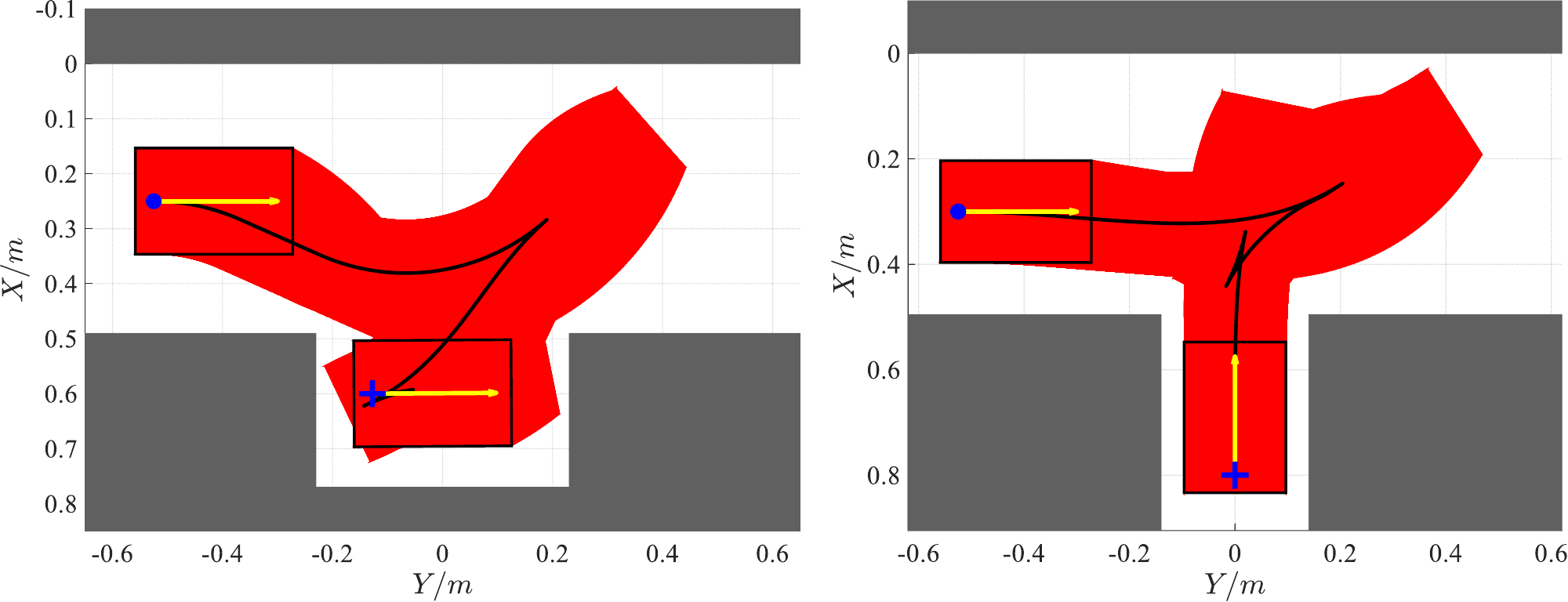}
		\caption{
			Schematic of the planned trajectory generated by the RITP method based on the loading parameters of the actual parking environment.}
		\label{fig:real_sim}
	\end{figure}
	
\begin{figure}[!t]
		\centering
		\subfloat[Block diagram of the ROS-based system for the NanoCar platform.]{\includegraphics[scale=0.28]{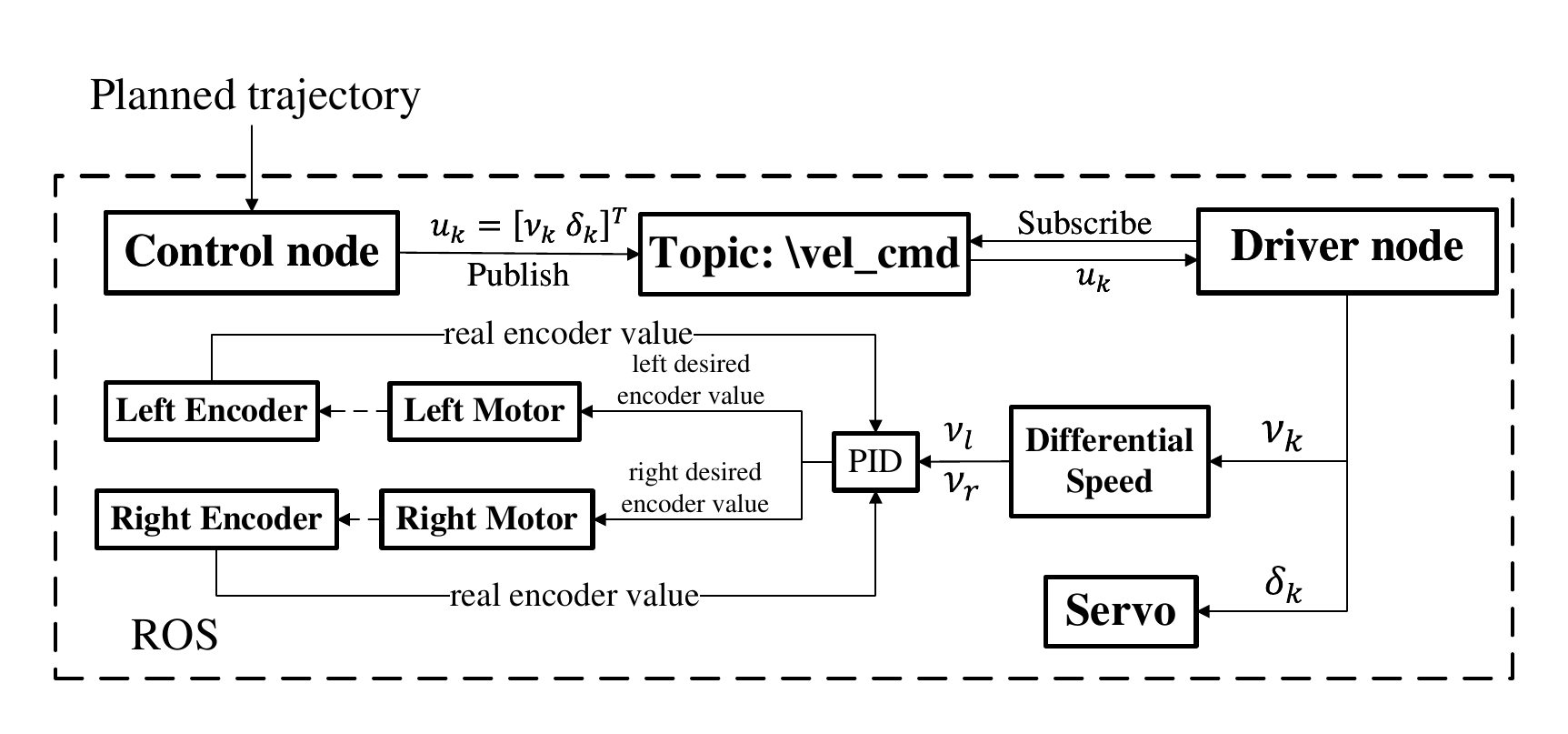}
			\label{fig_control_total}}
		
		\hfil
		\subfloat[ROS-based NanoCar platform.]{\includegraphics[scale=0.20]{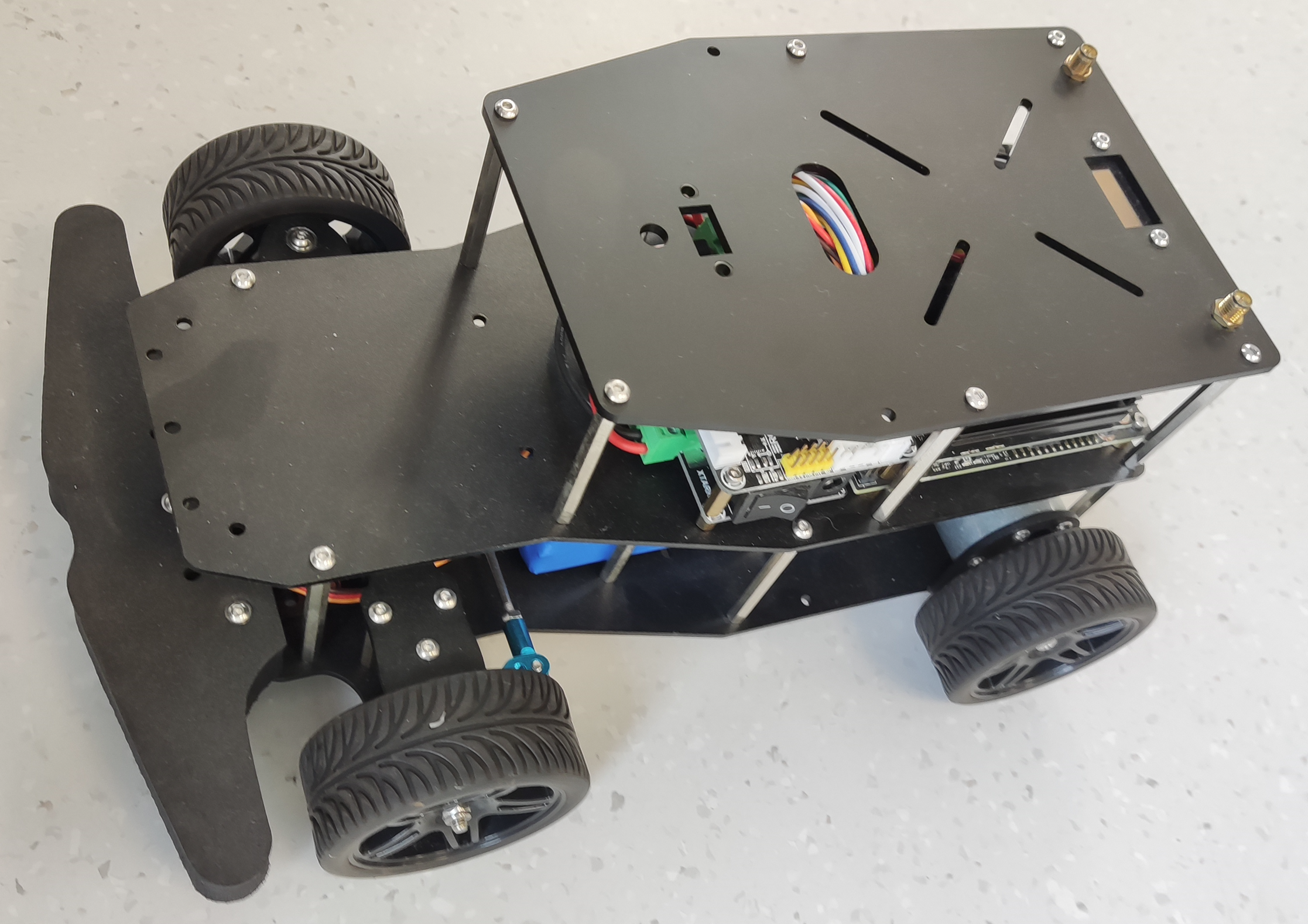}
			\label{fig:car}}
		\hfil
		
		\caption{Block diagram of the ROS system in NanoCar and a diagram of the vehicle structure.}
	\end{figure}

\begin{figure}[htbp]
		\centering
		\hspace*{0.1cm}
		\subfloat[Screenshots of parallel parking of NanoCar.]{\includegraphics[scale=0.21]{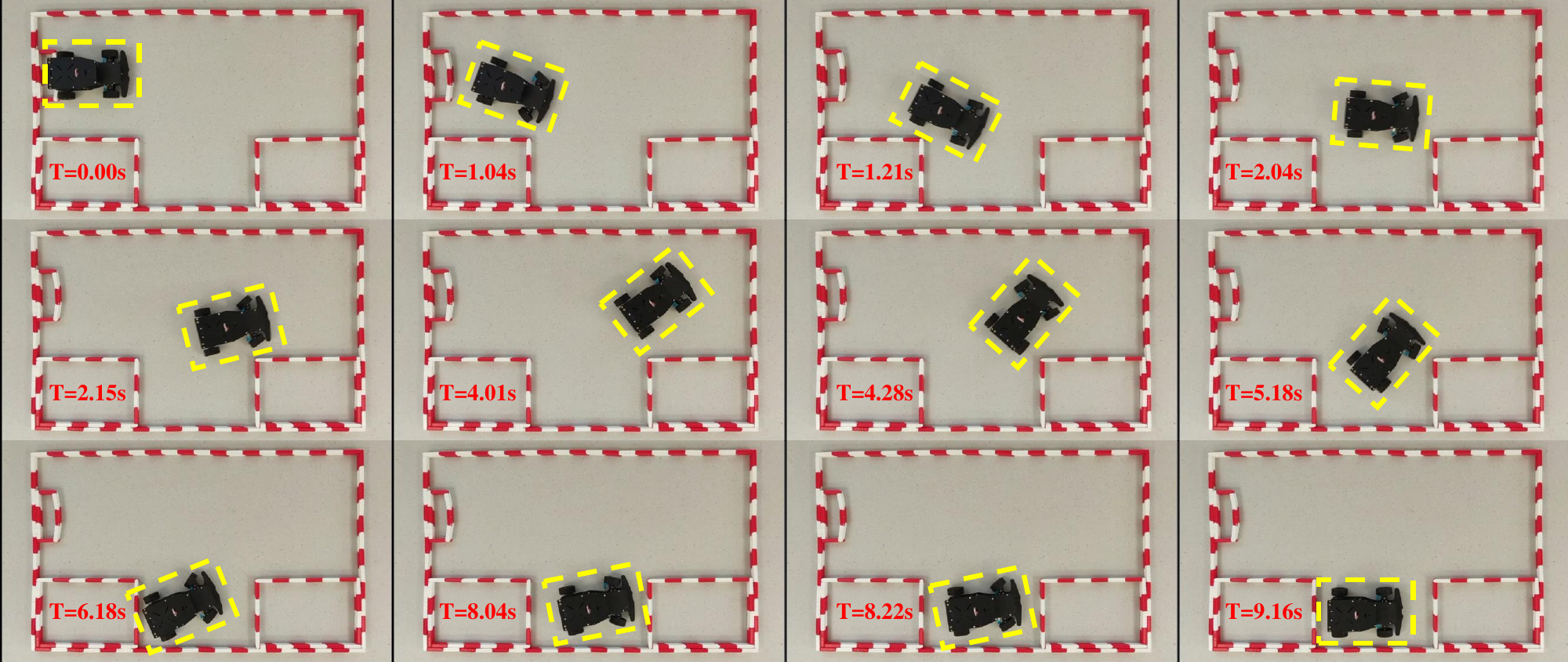}
			\label{fig_parallel_nanocar}}
		
		\hfil
		\subfloat[Screenshots of reverse parking of NanoCar.]{\includegraphics[scale=0.21]{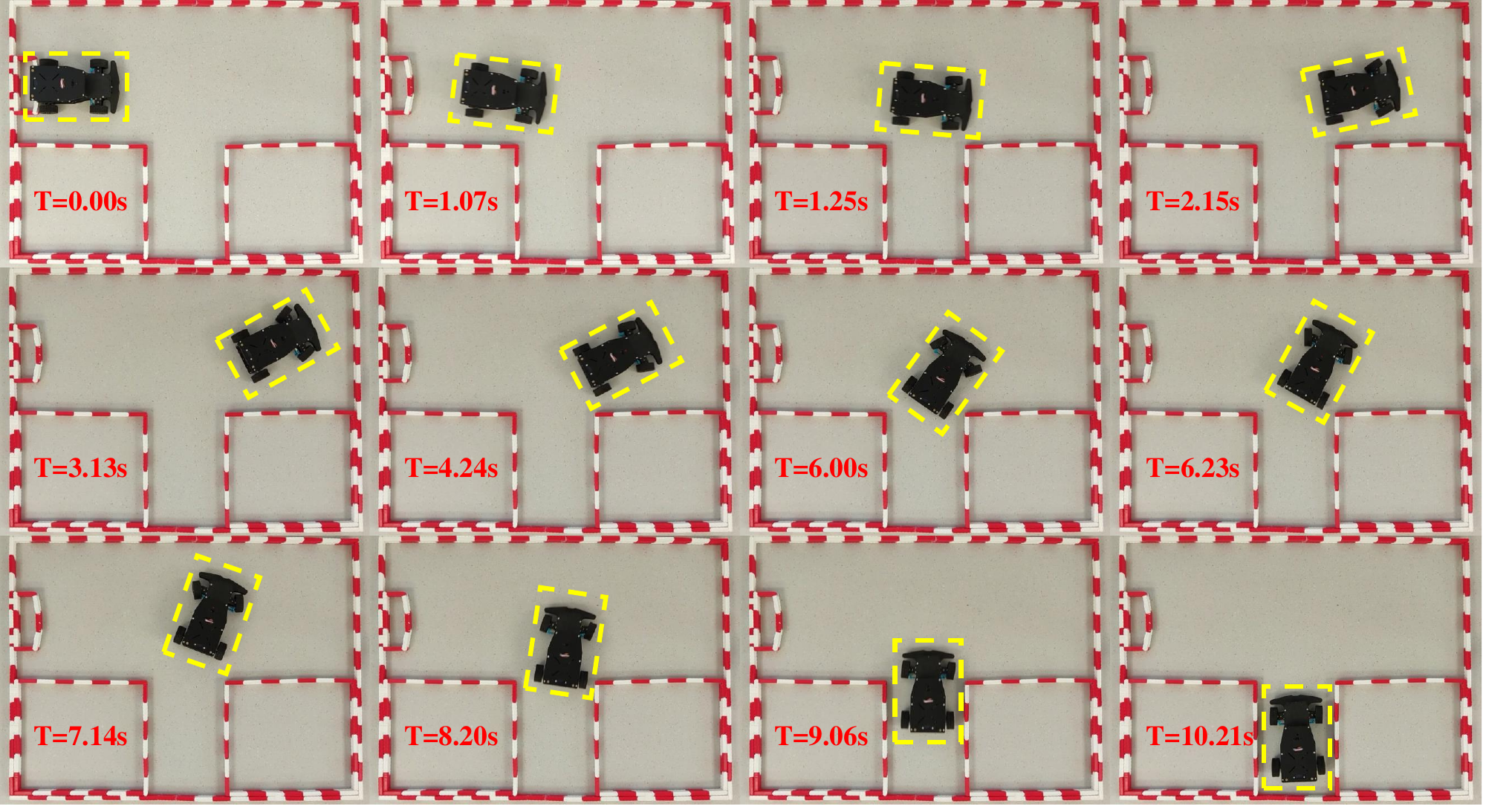}
			\label{fig_reverse_nanocar}}
		\hfil
		
		\caption{The trajectory planned by the RITP method controls the NanoCar to accomplish collision-free automated parking.}
		\label{fig:real_car}
	\end{figure}

	In this section, we utilized the RITP method on the ROS-based NanoCar platform (see Fig.~\ref{fig:car}). Then, we built parking scenarios for parallel and reverse parking (see Fig.~\ref{fig:real_car}) at a 1:10 scale and saved the parameters of these scenarios in Matlab. These parameters will serve as the environmental parameters for the RITP method. 
	
	The trajectory planned by the RITP method will be loaded into the ROS-based NanoCar to complete both parallel and reverse parking tasks. The simulation results of trajectory planning are shown in Fig~\ref{fig:real_sim}. The implementation block diagram of control inputs in the ROS-based NanoCar is shown in Fig.~\ref{fig_control_total}. The $\delta_k$ is calculated using~\eqref{eq:varphi_dot}.

	For the kinematic model, the NanoCar uses feed-forward control for automated parking. At the time $k$, the planned steering angle $\delta_k$ direct output without a controller and the planned velocity $v_k$ is used to calculate differential speed and then controlled by PID. The calculation of $v_ {\textrm{l}}$ and $v_ {\textrm{r}}$ is as follows:
	\begin{align*}
		v_ {\textrm{l}}=v_k\cdot(1-\frac{\mathbf{w} \cdot \tan(\delta_k)}{2\cdot L}),\\
		v_ {\textrm{r}}=v_k\cdot(1+\frac{\mathbf{w} \cdot \tan(\delta_k)}{2\cdot L}).
	\end{align*}
	where $\mathbf{w}=0.165 \, \textrm{m}$.
	
	Additionally, the following two issues need to be taken into account when applying the RITP method in actual vehicles. At first, because the feed-forward control method is used, it is necessary to ensure that the execution time of the control inputs is compatible with the time $\Delta \mathbf{t}$. Therefore, in practical applications, the impact of the delay of the program needs to be considered. Secondly, the control inputs obtained from the trajectory planned by the RITP method are the vehicle velocity and the steering angle, while actually, in the control of the chassis are the duty cycle of the motor and the servo. Therefore, in the practical application, it is necessary to consider the mapping relationship between the control inputs of the planned trajectory and the underlying control.
	
	The effectiveness of the RITP method applied to the NanoCar is shown in Fig.~\ref{fig:real_car}, which shows that NanoCar can also be parked in the parking space under a simple control strategy. When the RITP method is applied to different categories of vehicles, the relative relationship between vehicle size and parking space size needs to be considered. When the vehicle size is too large compared to the parking space, resulting in limited maneuvering space, the sizes of both the vehicle and obstacles can be inflated during collision detection in the ItCA process of the RITP method to ensure the safety of the planned trajectory. Enhancing the quality of the initial guess paths of the hybrid A* is also an excellent option.
	
\begin{figure}[!t]
		\centering
		\subfloat[Tracking effect of the left motor in  parallel parking.]{\includegraphics[scale=0.16]{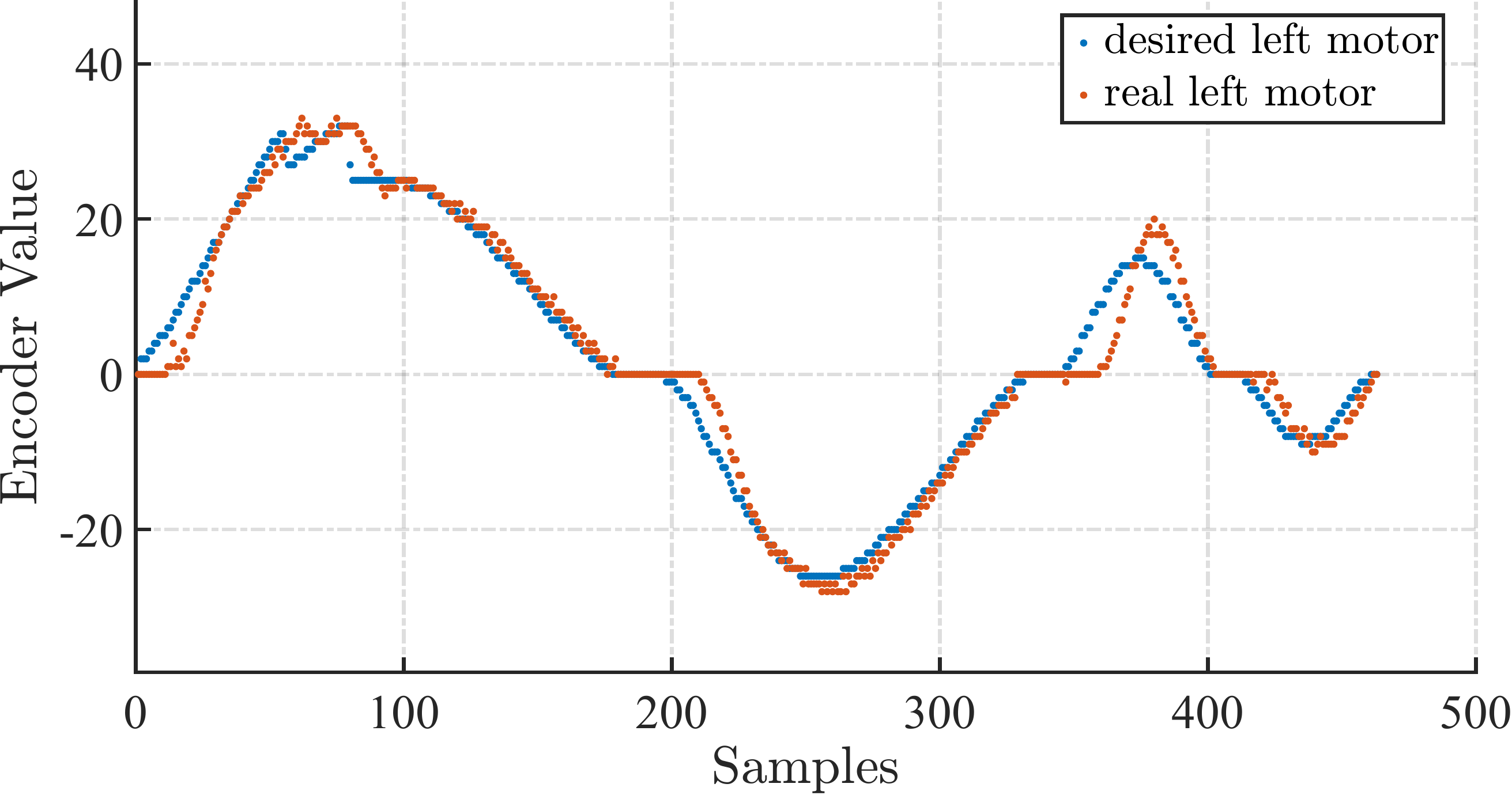}
			\label{fig_parallel_nanocar_data_1}}
		
		\hfil
		\subfloat[Tracking effect of the right motor in parallel parking.]{\includegraphics[scale=0.16]{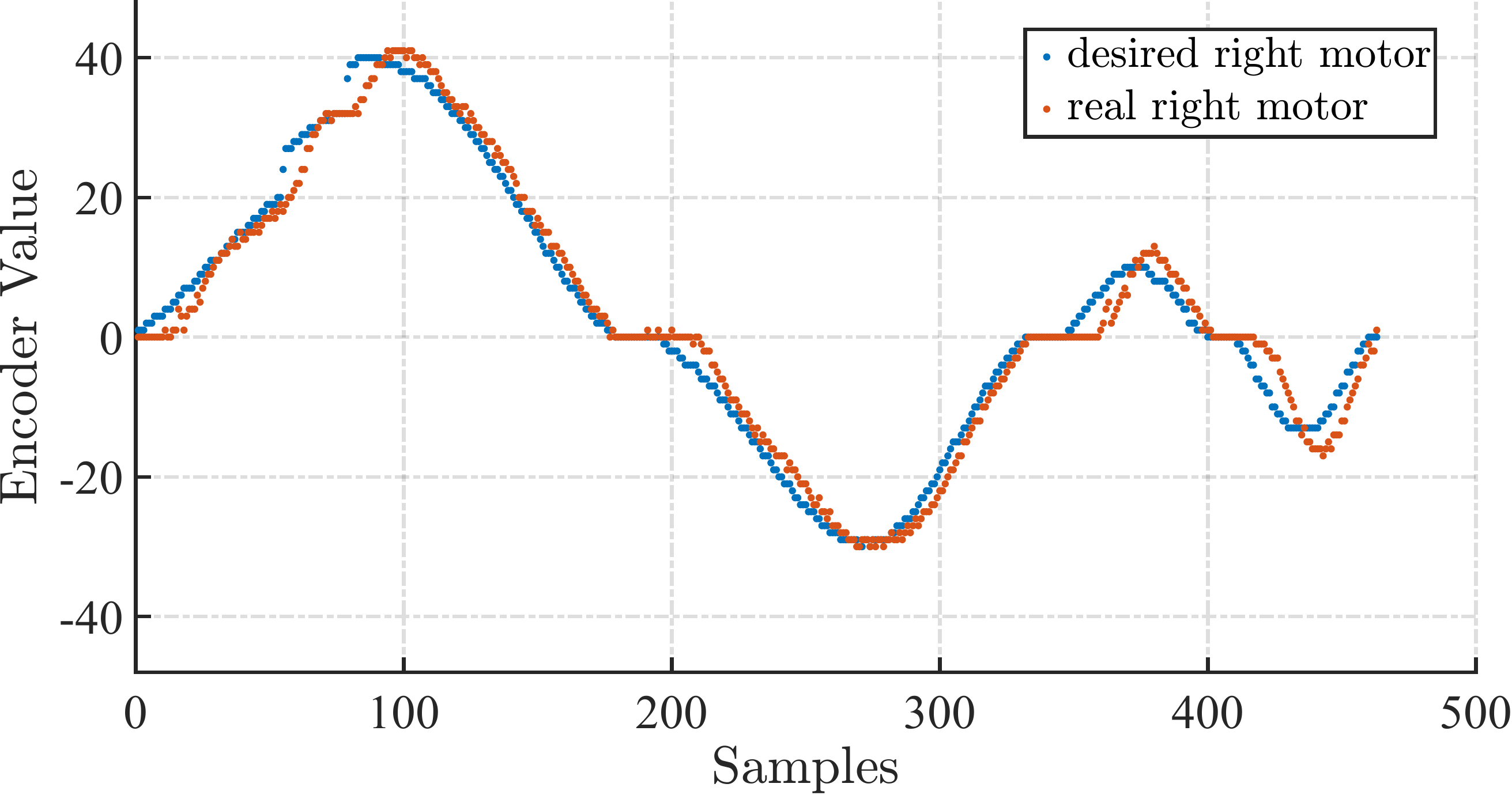}
			\label{fig_parallel_nanocar_data_2}}
		\hfil
		\subfloat[Tracking effect of the left motor in reverse parking.]{\includegraphics[scale=0.16]{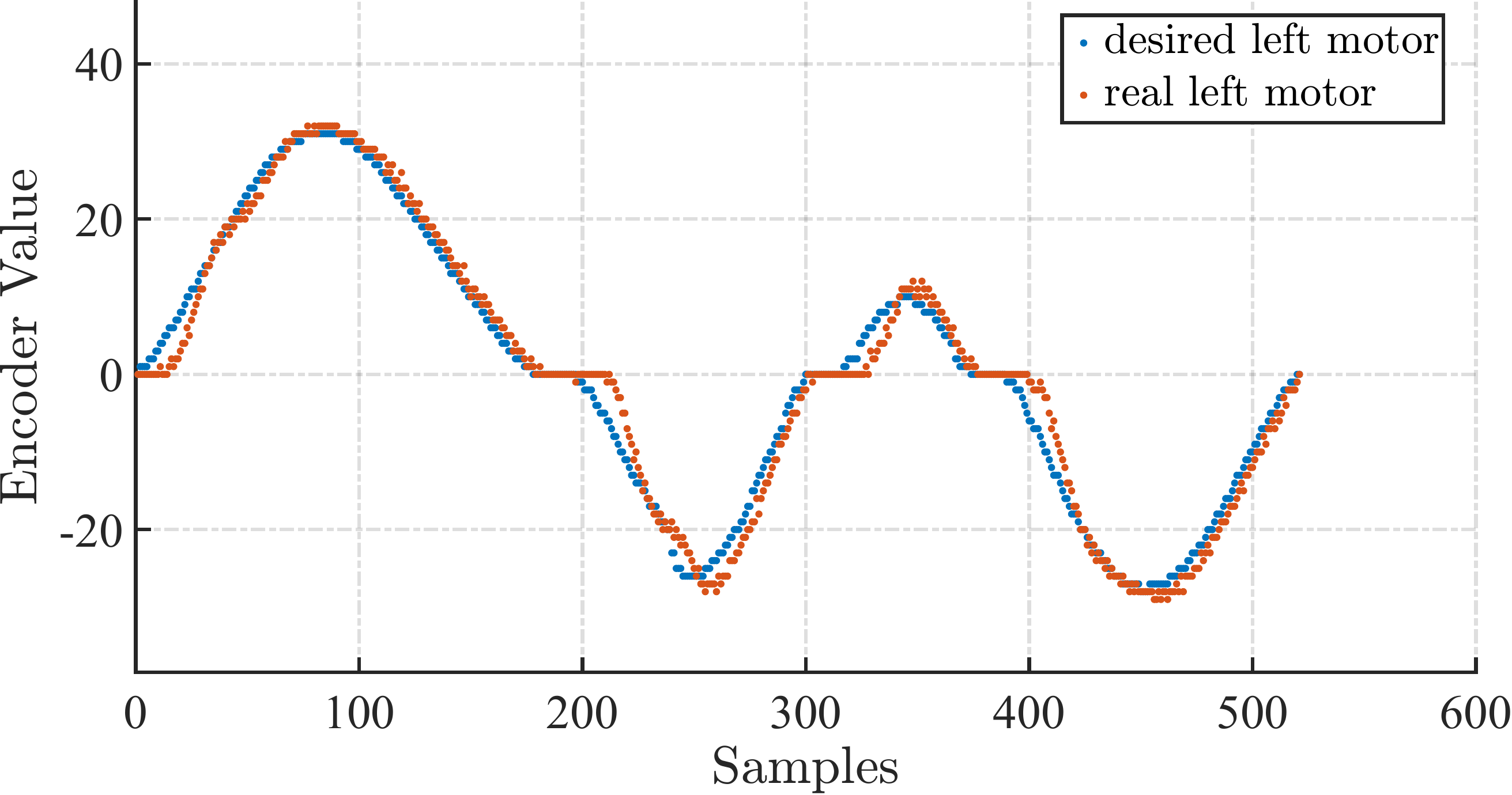}
			\label{fig_reverse_nanocar_data_1}}
		
		\hfil
		\subfloat[Tracking effect of the right motor in reverse parking.]{\includegraphics[scale=0.16]{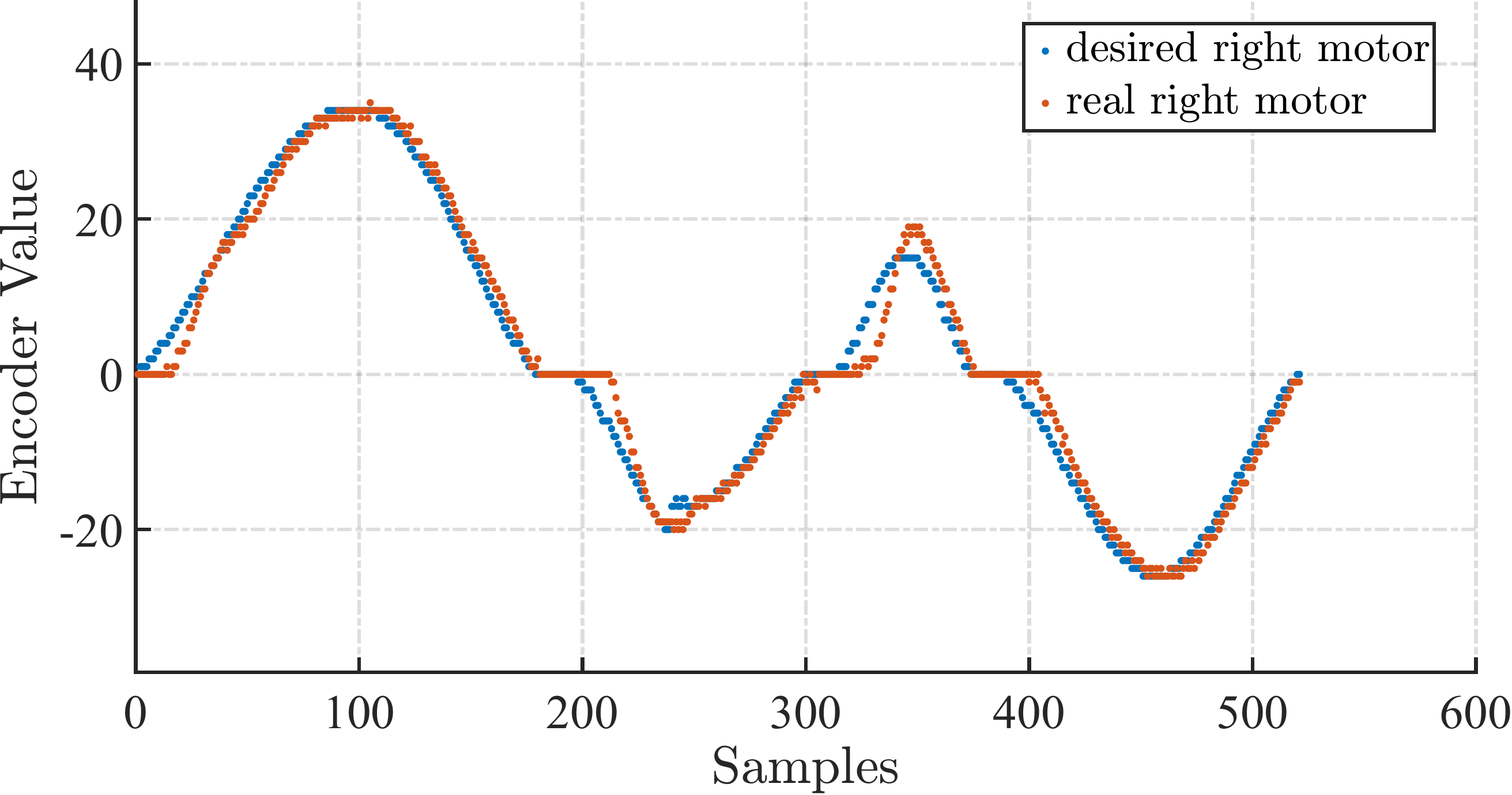}
			\label{fig_reverse_nanocar_data_2}}

		\caption{The desired and actual encoder values correspond to the left and right motors of NanoCar during parallel and reverse parking. The left and right motors are distinguished by the forward direction of the vehicle.}
		\label{fig_nanocar_data_parallel}
	\end{figure}

	The Fig.~\ref{fig_nanocar_data_parallel} shows the results of the actual and desired velocity of the NanoCar. The maximum velocity is $v_\textrm{max}=0.5 \,\textrm{m/s}$, which corresponds to the encoder value of 45. The maximum acceleration is $a_\textrm{max}=0.6 \, \textrm{m/s}^2$.  The wheelbase is $L=0.18$m, the width is $W=0.193$m, $a_f=0.725$m, $a_r=0.335$m. The trajectory planned using the RITP method can be accurately tracked by a simple controller. Experiments based on the NanoCar illustrate the strong control feasibility of trajectories planned by the RITP method as well as its application to actual vehicles.

	In this section, we deploy the trajectory planned by the RITP method in NanoCar and employ a simple feed-forward control strategy to track the planned trajectory, yielding encouraging outcomes. The experimental results validate that the RITP method can plan trajectories with strong control feasibility and is applicable in real-world vehicles.
	
	\section{Conclusion}\label{Conclusion}
	
	The proposed method incorporates several key features: Firstly, we introduce a path planning method that simultaneously achieves time-efficient and high-precision collision avoidance. This method effectively addresses precise and time-efficient collision avoidance by dynamically constructing the optimization problem during the iterative process and distributing the computational workload for collision detection across parallel computing units. Secondly, employing polynomials with orders higher than the cubic polynomial enables the planned path to satisfy the vehicle kinematics model through differential flatness. Finally, by incorporating terminal smoothing and kinematic constraints, the RITP method ensures a smoother trajectory with improved control feasibility, particularly at gear shifting points.

	The numerical simulation experiments verify the time efficiency of the proposed method as well as the strong control feasibility and collision-free performance of the planned trajectory. By incorporating parallel computing techniques, the proposed method provides significant time efficiency advantages over model-integrated and iteration-based trajectory planning methods. We find that separating collision avoidance from the optimization problem and leveraging parallel computing units to distribute the computational load of collision detection is effective. Collision detection with extremely dense sampling increases the computation time of the piecewise trajectory planning, but the total computation time is still low because of the presence of parallel computation, which distributes the computational load. By utilizing tracking error as the metric, we evaluate the control feasibility of the planned trajectory and confirm the effectiveness of using polynomials that guarantee the vehicle kinematics model through differential flatness for path planning. The ablation experiments reveal that while cubic polynomials can satisfy the vehicle kinematics model, their limited degrees of freedom hinders them from generating collision-free paths in tight scenarios. Compared to cubic polynomials, trajectory planning with quartic and quintic polynomials demonstrates a superior success rate. Furthermore, trajectories planned by quartic and quintic polynomials exhibit similar control feasibility, but trajectory planning using the quartic polynomial consumes less computation time. Moreover, the application of terminal smoothing constraints significantly enhances the control feasibility of planned trajectories. This is because terminal smoothing constraints guarantee the continuity of yaw angle changes of the planned trajectory at gear shifting points. Finally, we apply the proposed method to an actual ROS-based vehicle and confirm its applicability in practical environments.
	
	In our future work, we plan to implement the proposed method in more complex scenarios. Additionally, we will ensure that the parameters of the scenarios are normalized to avoid any impact on the effectiveness of the proposed method due to the varying sizes of the scenarios. At the same time, we plan to optimize the distribution of sampling points used to perform collision detection to improve the efficiency of collision avoidance further.

	\section*{Acknowledgments}
	The authors thank Ran Duo, Xiaoling Zhou, Zhiming Zhang, and Wangjia Weng for their help in accomplishing this research work. Thanks to Bei Zhou for her insightful suggestions on the writing of the paper. We thank all reviewers and editors for their very valuable and insightful comments.
	
	\bibliographystyle{elsarticle-num-names} 
	\bibliography{bibref.bib, speed_ref.bib}

\end{document}